\documentclass[twoside]{article}
\usepackage[accepted]{aistats2021}
\usepackage[round]{natbib}

\usepackage{titletoc}
\usepackage[utf8]{inputenc} 
\usepackage[T1]{fontenc}    
\usepackage{hyperref}       
\usepackage{url}            
\usepackage{booktabs}       
\usepackage{amsfonts}       
\usepackage{nicefrac}       
\usepackage{microtype}      

\usepackage{graphicx}
\usepackage{subfigure}
\usepackage{multirow}
\usepackage{adjustbox}
\usepackage{hyperref}

\usepackage{amsmath}
\usepackage{amssymb}
\usepackage{float}
\usepackage{caption}

\renewcommand{\vec}[1]{\ensuremath{\boldsymbol{#1}}}
\renewcommand{\v}[1]{\vec{#1}}

\newcommand{\x}{\ensuremath{\v{x}}}
\newcommand{\y}{\ensuremath{\v{y}}}
\newcommand{\z}{\ensuremath{\v{z}}}

\newcommand{\p}{\ensuremath{p_\theta}}
\newcommand{\q}{\ensuremath{q_\phi}}
\newcommand{\f}{\ensuremath{\phi}}

\usepackage{mathtools}
\usepackage{xspace}
\DeclarePairedDelimiterX{\KLx}[2]{(}{)}{%
  #1\:\delimsize\|\:#2%
}
\newcommand{\KL}{\ensuremath{\text{KL}\KLx}}
\newcommand{\E}{\ensuremath{\mathbb{E}}}

\newcommand{\ntrain}{\ensuremath{N_{\text{train}}}}

\newtheorem{thm}{Proposition}

\begin{document}

%

%
\runningauthor{A. Bozkurt, B. Esmaeili, J.-B. Tristan, D. H. Brooks, J. G. Dy, J.-W. van de Meent}

\twocolumn[

\aistatstitle{Rate-Regularization and Generalization in VAEs}
  
\aistatsauthor{Alican Bozkurt*\\
  Northeastern University\\
  \texttt{alican@ece.neu.edu}\\
  \And
  Babak Esmaeili*\\
  Northeastern University\\
  \texttt{esmaeili.b@northeastern.edu}\\
  \And
  Jean-Baptiste Tristan \\
  Boston College \\
  \texttt{tristanj@bc.edu}\\
  \AND
  Dana H. Brooks\\
  Northeastern University\\
  \texttt{brooks@ece.neu.edu}\\
  \And
  Jennifer G. Dy\\
  Northeastern University\\   
  \texttt{jdy@ece.neu.edu}\\
  \And
  Jan-Willem van de Meent\\
  Northeastern University\\
  \texttt{j.vandemeent@northeastern.edu}\\}

\aistatsaddress{ }]

\begin{abstract}
  Variational autoencoders optimize an objective that combines a reconstruction loss (the distortion) and a KL term (the rate). The rate is an upper bound on the mutual information, which is often interpreted as a regularizer that controls the degree of compression. We here examine whether inclusion of the rate also acts as an inductive bias that improves generalization. We perform rate-distortion analyses that control the strength of the rate term, the network capacity, and the difficulty of the generalization problem. Decreasing the strength of the rate paradoxically \emph{improves} generalization in most settings, and reducing the mutual information typically leads to underfitting. Moreover, we show that generalization continues to improve even after the mutual information saturates, indicating that the gap on the bound (i.e.\nobreak\ the KL divergence relative to the inference marginal) affects generalization. This suggests that the standard Gaussian prior is not an inductive bias that typically aids generalization, prompting work to understand what choices of priors improve generalization in VAEs.
\end{abstract}

\section{Introduction}
\label{sec:intro}

Variational autoencoders (VAEs) learn representations in an unsupervised manner by training an encoder, which maps high-dimensional data to a lower-dimensional latent code, along with a decoder, which parameterizes a manifold that is embedded in the data space~\citep{kingma2013auto,rezende2014stochastic}. Much of the work on VAEs has been predicated on the observation that distances on the learned manifold can reflect semantically meaningful factors of variation in the data. This is commonly illustrated by visualizing interpolations in the latent space, or more generally, interpolations along geodesics~\citep{chen2019fast}. 

The ability of VAEs to interpolate is often attributed to the variational objective~\citep{ghosh2019variational}. VAEs maximize a lower bound on the log-marginal likelihood, which comprises a reconstruction loss and a Kullback-Leibler (KL) divergence between the encoder and the prior (called the rate). Minimizing the reconstruction loss in isolation is equivalent to training a deterministic autoencoder. For this reason, the rate is often interpreted as a regularizer that induces a smoother representation~\citep{chen2016variational,berthelot2018understanding}. 

In this paper, we ask the question of whether the inclusion of the rate term also improves generalization. 
That is, does this penalty reduce the reconstruction loss for inputs that were unseen during training? A known property of VAEs is that the optimal decoder will memorize training data in the limit of infinite capacity~\citep{alemi2018fixing,shu2018amortized}, as will a deterministic autoencoder~\citep{uhler}. At the same time, there is empirical evidence that VAEs can underfit the training data, and that reducing the strength of the rate term can mitigate underfitting~\citep{hoffman2017v-vaes}. Therefore, we might hypothesize that VAEs behave like any other model in machine learning; high-capacity VAEs will overfit the training data, but we can improve generalization by adjusting the strength of the KL term to balance overfitting and underfitting.

To test this hypothesis, we performed experiments that systematically vary the strength of the rate term and the network capacity. In these experiments, we deliberately focus on comparatively simple network architectures in the form of linear and convolutional layers with standard spherical Gaussian priors. These architectures remain widely used in work on VAEs, particularly work that focuses on disentangled representations, and systematically investigating these cases provides us with results that can form a basis for understanding the wide variety of more sophisticated architectures that exist in the literature. 

The primary aim of our experiments is to carefully control the difficulty of the generalization problem. Our goal in doing so is to disambiguate between apparent generalization that can be achieved by simply reconstructing the most similar memorized training examples and generalization that requires reconstruction of examples that differ substantially from those seen in the training set. To achieve this goal, we have created a dataset of J-shaped tetrominoes that vary in color, size, position, and orientation. This dataset gave us a sufficient variation of both the amount of training data and the density of data in the latent space, as well as sufficient sensitivity of reconstruction loss to variation in these factors, in order to evaluate out-of-domain generalization to unseen combinations of factors.

The surprising outcome of our experiments is that the rate term does not, in general, improve generalization in terms of the reconstruction loss. We find that VAEs memorize training data in practice, even for simple 3-layer fully-connected architectures. However, contrary to intuition, \emph{reducing} the strength of the rate term \emph{improves} generalization under most conditions, including in out-of-domain generalization tasks. The only case where an optimum level of rate-regularization emerges is when low-capacity VAEs are trained on data that are sparse in the latent space. We show that these results hold for both MLP and CNN-based architectures, as well as a variety of datasets. 

These results suggest that we need to more carefully quantify the effect of each term in the VAE objective on the generalization properties of the learned representation. To this end, we decompose the KL divergence between the encoder and the prior into its constituent terms: the mutual information (MI) between data and the latent code and the KL divergence between the inference marginal and the prior. We find that the MI term saturates as we reduce the strength of the rate term, which indicates that it is in fact the KL between the inference marginal and prior that drives improvements in generalization in high-capacity models. This suggests that the standard spherical Gaussian prior in VAEs is not an inductive bias that aids generalization in most cases, and that more flexible learned priors may be beneficial in this context.


\section{Variational Autoencoders}
\label{sec:vae}
VAEs jointly train a generative model $\p(\x,\z)$ and an inference model $\q(\x, \z)$. The generative model comprises a prior $p(\z)$, typically a spherical Gaussian, and a likelihood $\p(\x \,|\, \z)$ that is parameterized by a neural network known as the decoder. The inference model is defined in terms of
a variational distribution $\q(\z \,|\, \x)$, parameterized by an encoder network, and a data distribution $q(\x)$, which is typically an empirical distribution  $q(\x) \!=\! \frac{1}{N} \sum_{n} \delta_{\x_n}(\x)$ over training data $\{\x_1, \dots, \x_N\}$. The two models are optimized by maximizing a variational objective~\citep{higgins2017beta}
\begin{align}
    \label{eq:beta-vae}
    \mathcal{L}_{\beta}(\theta, \phi) 
    =&
    \E_{\q(\z,\x)}
    \big[
    \log \p(\x \,|\, \z)
    \big]\\
    &-
    \beta \:
    \E_{q(\x)}
    \big[
    \KL{\q(\z \,|\, \x)}{p(\z)}
    \big]
    . \nonumber
\end{align}
The multiplier $\beta$, which in a standard VAE is set to 1, controls the relative strength of the reconstruction loss and the KL loss. We will throughout this paper refer to these two terms $\mathcal{L}_\beta = -D - \beta R$ as the distortion $D$ and the rate $R$. The distortion defines a reconstruction loss, whereas the rate constrains the encoder distribution $\q(\z \,|\, \x)$ to be similar to the prior $p(\z)$. As $\beta$ approaches 0, the VAE objective becomes similar to that of a deterministic autoencoder; in absence of the rate term, the distortion is minimized when the encoding is a delta-peak at the maximum-likelihood value $\text{argmax}_{\z} \log p(\x \mid \z)$. For this reason, a standard interpretation is that the rate serves to induce a smoother representation and ensures that samples from the generative model are representative of the data.

While there is evidence that the rate term indeed induces a smoother representation~\citep{shamir2010learning}, it is not clear whether this smoothness mitigates overfitting, or indeed to what extent VAEs are prone to overfitting in the first place. Several researchers~\citep{bousquet2017optimal,rezende2018taming,alemi2018fixing, shu2018amortized} have pointed out that an infinite-capacity optimal decoder will memorize training data, which suggests that high-capacity VAEs will overfit. On the other hand, there is also evidence of underfitting; setting $\beta < 1$ can improve the quality of reconstructions in VAEs for images~\citep{hoffman2017v-vaes,engel2017latent}, natural language~\citep{wen2017latent}, and recommender systems~\citep{liang2018variational}. 

More broadly, precisely what constitutes generalization and overfitting in this model class is open to interpretation. If we view the VAE objective primarily as a means of training a generative model, then it makes sense to evaluate model performance in terms of the log marginal likelihood $\log \p(\x)$. This view is coherent for the standard VAE objective ($\beta=1$), which defines a lower bound 
\begin{align*}
    \mathcal{L}(\theta, \phi)
    &=
    \mathbb{E}_{q(\x)} 
    \big[
      \log \p(\x)
      -
      \KL{\q(\z \,|\, \x)}{\p(\z \,|\, \x)}
    \big]\\ 
    &\le
    \mathbb{E}_{q(\x)} 
    \big[
      \log \p(\x)
    \big]
    .
\end{align*}
The KL term 
indirectly regularizes the generative model when the encoder capacity is constrained~\citep{shu2018amortized}. Note however that $\mathcal{L}_{\beta}$ is not a lower bound on $\log \p(\x)$ when $\beta < 1$. This means that it does not make sense to evaluate generalization in terms of $\log \p(\x)$ when $\beta \to 0$, or in deterministic autoencoders that do not define a generative model to begin with.

In this paper, we view the VAE primarily as a model for learning representations in an unsupervised manner. In this view, generation is more ancillary; The encoder and decoder serve to define a lossy compressor and decompressor, or equivalently to define a low-dimensional manifold that is embedded in the data space. Our hope is that the learned latent representation reflects semantically meaningful factors of variation in the data, whilst discarding nuisance variables.

The view of VAEs as lossy compressors can be formalized by interpreting the objective $\mathcal{L}_{\beta}$ as a special case of information-bottleneck (IB) objectives~\citep{tishby2000information,alemi2017deep,alemi2018fixing}. This interpretation relies on the the observation that the decoder $\p(\x \,|\, \z)$ defines a lower bound on the MI in the inference model $\q(\z, \x)$
in terms of a distortion $D$ and entropy $H$
\begin{align}
    &H-D 
    \le 
    I_q[\x ; \z],\\
    &D = 
    -\E_{\q(\x, \z)}
    \big[
      \log \p(\x \,|\, \z)
    \big], \\
    &H = 
    -\E_{q(\x)}
    \big[
      \log q(\x)
    \big].
\end{align}
Similarly, the rate $R$ is an upper bound on this same mutual information $R \ge I_q[\x ; \z]$,
\begin{align}
    R 
    &= 
    \E_{\q(\x,\z)}
    \big[
     \KL{\q(\z \,|\, \x)}{p(\z)}
    \big] \\ 
    &= 
    I_q[\x ; \z] 
    + 
    \KL{\q(\z )}{p(\z)}
    . \nonumber
    \label{eq:r-decomposition}
\end{align}
Here the term $\KL{\q(\z )}{p(\z)}$ is sometimes called ``the marginal KL'' in the literature~\citep{rezende2018taming}. The naming of the rate and distortion terms originates from rate-distortion theory~\citep{cover2012elements}, which seeks to minimize $I_q[\x;\z]$ subject to the constraint $D \le D^*$. The connection to VAEs now arises from the observation that $\mathcal{L}_\beta$ is a Lagrangian relaxation of the rate-distortion objective 
\begin{align}
    \mathcal{L}_{\beta} = -D - \beta \: R.
\end{align}
The appeal of this view is that it suggests an interpretation of the distortion $D$ as an empirical risk and of $I_q[\x;\z]$ as a regularizer~\citep{shamir2010learning}. This leads to the hypothesis that VAEs may exhibit a classic bias-variance trade-off: In the limit $\beta \to 0$, we may expect low distortion on the training set but poor generalization to the test set, whereas increasing $\beta$ may mitigate this form of overfitting.

At the same time, the rate-distortion view of VAEs gives rise to some peculiarities. Standard IB methods use a regressor or classifier $\p(\y \,|\, \x)$ to define a lower bound $H-D \le I_q[\y;\z]$ on the MI between the code $\z$ and a target variable $\y$~\citep{tishby2000information}. The objective is to maximize $I_q[\y;\z]$, which serves to learn a representation $\z$ which is predictive of $\y$, whilst minimizing $I_q[\x ;\z]$, which serves to compress $\x$ by discarding information irrelevant to $\y$. However, this interpretation does not translate to the special case of VAEs, where $\x = \y$. Here any compression will necessarily increase the distortion since $D \ge H - I_q[\x;\z]$.

In our experiments, we will explicitly investigate to what extent $\beta$ controls a trade-off between overfitting and underfitting. To do so, we will compute $RD$ curves that track the rate and distortion under varying $\beta$. While $RD$ curves have been used to evaluate model performance on the training set~\citep{alemi2018fixing, rezende2018taming}, we are not aware of work that explicitly probes generalization to a test set. 

To see how overfitting and underfitting may manifest in this analysis, we can consider the hypothetical case of infinite-capacity encoders and decoders. For such networks, both bounds will be tight at the optimum and $\mathcal{L}_\beta = (1-\beta) I_q[\x;\z] - H$. Maximizing $\mathcal{L}_\beta$ with respect to $\f$ will lead to an \emph{autodecoding} limit when $\beta > 1$, which minimizes $I_q[\x;\z]$, and an \emph{autoencoding} limit when $\beta < 1$, which maximizes $I_q[\x;\z]$~\citep{alemi2018fixing}. One hypothesis is that we will observe poor generalization to the test set in either limit, since maximizing $I_q[\x;\z] $ could lead to overfitting whereas minimizing $I_q[\x;\z]$ could lead to underfitting. Moreover, an infinite-capacity generator will fully memorize the training data, which could lead to poor generalization performance in terms of the log marginal likelihood.

\begin{figure*}[!t]
    \centering
    \includegraphics[width=\linewidth]{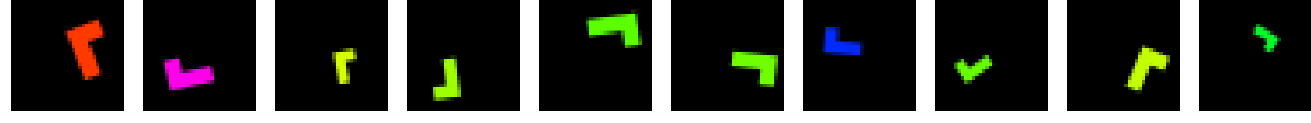}
    \vspace*{-3ex}
    \caption{We simulate 164k tetrominoes that vary in position, orientation, size, and color. 
    }
    \label{fig:tetris}
\end{figure*}

In practice, it may well be that the decoder $\p(\x \,|\, \z)$ can be approximated as an infinite-capacity model. We present empirical evidence of this phenomenon in Appendix~\ref{app:sec:mnist} that is consistent with recent analyses~\citep{bousquet2017optimal,rezende2018taming,alemi2018fixing, shu2018amortized}. However, it is typically not the case that the prior $p(\z)$ has a high capacity. In fact, a standard spherical Gaussian prior effectively has 0 capacity, since its mean and variance define an affine transformation that can be trivially absorbed into the first linear layer of any encoder and decoder. This means that the upper bound will be loose and that the rate $R$ may in practice represent a trade-off between $I_q[\x; \z]$ and $\KL{\q(\z)}{p(\z)}$, at least when the encoder capacity is limited. We present evidence of this trade-off in Section~\ref{sec:mi-vs-kl}.

\vspace{-0.25\baselineskip}
\section{Related Work}
\label{sec:related-work}
\vspace{-0.25\baselineskip}

\paragraph{Generalization in VAEs.} Recent work that evaluates generalization in VAEs has primarily considered this problem from the perspective of VAEs as generative models. \citet{shu2018amortized} consider whether constraining encoder capacity can serve to mitigate data memorization, whereas \citet{zhao2018bias} ask whether VAEs can generate examples that deviate from training data. \citet{kumar2020implicit} derive a deterministic approximation to the $\beta$-VAE objective and show that $\beta$-VAE regularizes the generative by imposing a constraint on the Jacobian of the encoder. Whereas \citet{kumar2020implicit} evaluate generalization in terms of FID scores~\citep{fid}, we here focus on $RD$ curves. \citet{huangevaluating} also discuss evaluating deep generative models based on $RD$ curves. They show that this type of analysis can be used to uncover some of the known properties of VAEs such as the ``holes problem''~\citep{rezende2018taming} by tracking the change in the curve for different sizes of latent space. In our work, we focus on the change of the $RD$ curve as the generalization problem becomes more difficult.  

\paragraph{Generalization and regularization in deterministic autoencoders.}  \citet{samy} and \citet{uhler} study generalization in deterministic autoencoders, showing that these models can memorize training data if they are over-parameterized. We overall observed a similar behaviour in our experiments. However, for our experiments, we did not consider architectures as deep as the ones in \citet{samy} and \citet{uhler}. \citet{ghosh2019variational} show that combining deterministic autoencoders with regularizers other than the rate can lead to competitive generative performance.

\paragraph{Generalization of disentangled representations.} Our work is indirectly related to research on disentangled representations, in the sense that some of this work is motivated by the desire to learn representations that can generalize to unseen combinations of factors \citep{narayanaswamy2017learning,kim2018disentangling,pmlr-v89-esmaeili19a,chen2018isolating,francesco}. There has been some work to quantify the effect of disentangling on generalization~\citep{eastwood, pmlr-v89-esmaeili19a, francesco}, but the extent of this effect remains poorly understood. In this paper, we explicitly design our experiments to test generalization to data with unseen combinations of factors, but we are not interested in disentanglement per se.



\section{Experiments}
\label{sec:experiments}

To quantify the effect of rate-regularization on generalization, we designed a series of experiments that systematically control three factors in addition to the $\beta$-coefficient: the amount of training data, the density of training data relative to the true factors of variation, and the depth of the encoder and decoder networks. To establish baseline results, we begin with experiments that vary all three factors in fully-connected architectures on a simulated dataset of Tetrominoes. We additionally consider convolutional architectures, as well as other simulated and non-simulated datasets.

\vspace*{-1.5ex}
\subsection{Tetrominoes Dataset}
\vspace*{-1.0ex}
\label{sec:tetris}

When evaluating generalization we have two primary requirements for a dataset. The first is that failures in generalization should be easy to detect. A good way to ensure this is to employ data for which we can achieve high-quality reconstructions for training examples, which makes it easier to identify degradations for test examples. The second requirement is that we need to be able to disambiguate effects that arise from a lack of data from those that arise from the difficulty of the generalization problem. When a dataset comprises a small number of  examples, this may not suffice to train an encoder and decoder network. Conversely, even when employing a large training set, a network may not generalize when there are a large number of generative factors.

To satisfy both requirements, we begin with experiments on simulated data. This ensures that we can explicitly control the density of data in the space of generative factors, and that we can easily detect degradations in reconstruction quality. We initially considered the dSprites dataset~\citep{dsprites17}, which contains 3 shapes at 6 scales, 40 orientations, and $32^2$ positions. Unfortunately, shapes in this dataset are close to convex. Varying either the shape or the rotation results in small deviations in pixel space, which in practice makes it difficult to evaluate whether a model memorizes the training data.

To overcome this limitation, we created the Tetrominoes dataset. This dataset comprises 163,840 procedurally generated 32$\times$32 color images of a J-shaped tetromino, which is concave and lacks rotational symmetry. We generate images based on five i.i.d.\nobreak\ continuous generative factors, which are sampled uniformly at random: rotation (sampled from the [0.0, 360.0] range), color (hue, sampled from [0.0, 0.875] range), scale (sampled from [2.0, 5.0] range), and horizontal and vertical position (sampled from an adaptive range to ensure no shape is placed out of bounds). To ensure uniformity of the data in the latent space, we generate a stratified sample; we divide each feature range into bins and sample uniformly within bins. Examples from the dataset are shown in Figure~\ref{fig:tetris}.

\begingroup
\begin{figure}[!t]
    \centering
    \resizebox{\columnwidth}{!}{%
    \renewcommand{\arraystretch}{0.1}
        \begin{tabular}{lcc}
          \toprule\\[4pt]
          & \textsf{\large \textbf{``Easy''}} & \textsf{\large \textbf{``Hard''}}\\[4pt]
          & {\small \textsf{\textbf{(test data close to training data)}}}&
          {\small \textsf{\textbf{(test data far from training data)}}}\\[6pt]
          \midrule\\[6pt]
          & \textsf{\large Default (300/300)} & \textsf{\large Checkerboard (300/300)} \\ 
         \begin{tabular}{r}
         \rotatebox[origin=c]{90}{\textbf{\large \textsf{Large training data}}}
         \end{tabular} &
         \adjustbox{valign=c}{ \includegraphics[height=2.1in]{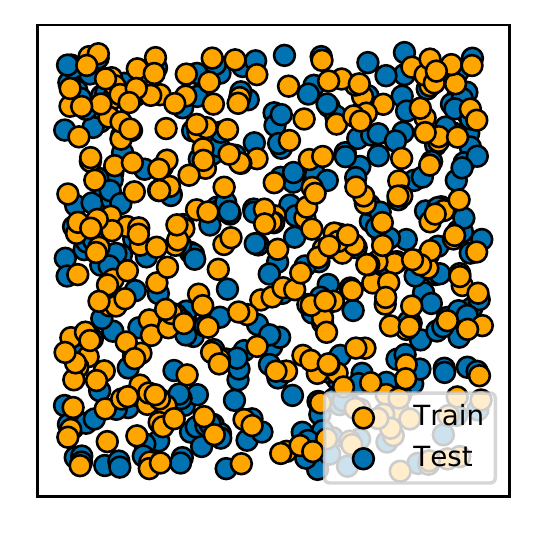}} &
         \adjustbox{valign=c}{
         \includegraphics[height=2.1in]{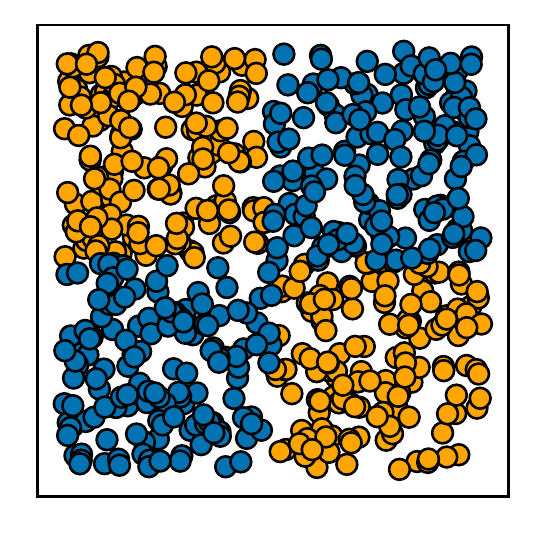}} \\
          & \textsf{\large Constant Density (60/60)} & \textsf{\large Constant Volume (60/540)} \\ 
         \begin{tabular}{r}
         \rotatebox[origin=c]{90}{\textbf{\large \textsf{Small training data}}}
         \end{tabular} & \adjustbox{valign=c}{
         \includegraphics[height=2.1in]{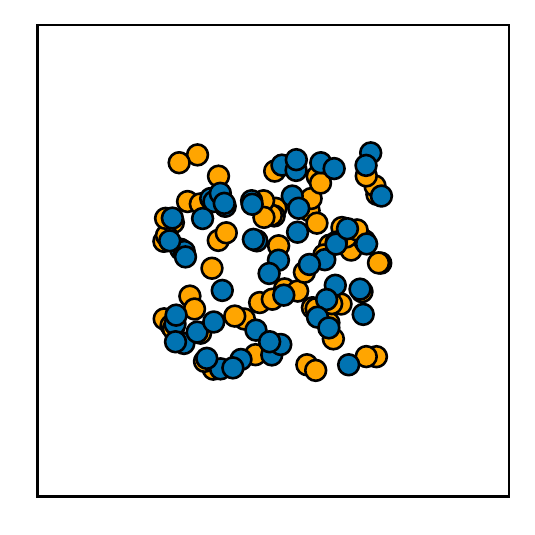}} &
         \adjustbox{valign=c}{
         \includegraphics[height=2.1in]{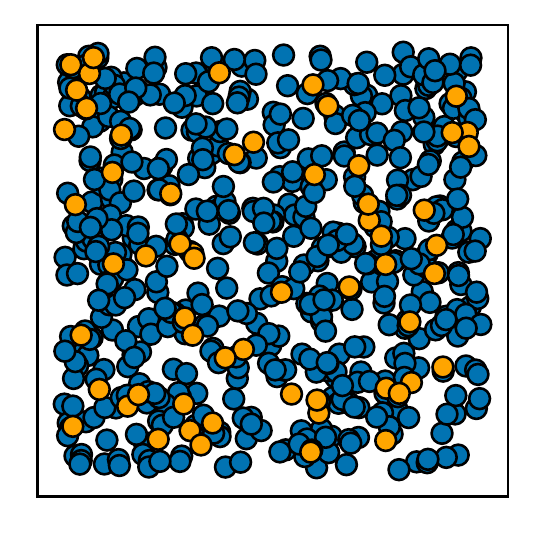}} \\
         \bottomrule
        \end{tabular}
    }
    \vspace*{-1ex}
    \caption{We define 4 train/test splits, which vary in the amount of training data and the typical distance between test data and their nearest neighbors in the training set. Here we show 600 samples with 2 generative factors for visualization.
    }
    \label{fig:generalization}
\end{figure}
\endgroup

\subsection{Train/Test Splits}

In our experiments, we compare 4 different train/test splits that are designed to vary two components: (1) the amount of training data, (2) the typical distance between training and test examples.

\emph{1. 50/50 random split (Default).} The base case in our analysis (Figure~\ref{fig:generalization}, 1\textsuperscript{st} from left) is a 82k/82k random train/test split of the full dataset. This case is designed to define an ``easy'' generalization problem, where similar training examples will exist for most examples in the test set.

\emph{2. Large data, (Checkerboard) split.} We create a 82k/82k split in which a 5-dimensional ``checkerboard'' mask partitions the training and test set (Figure~\ref{fig:generalization}, 2\textsuperscript{nd} from left). This split has the same amount of training data as the base case, as well as the same (uniform) marginal distribution for each of the feature values. This design ensures that for any given test example, there are 5 training examples that differ in one feature (e.g.\nobreak\ color) but are similar in all other features (e.g.\nobreak\ position, size, and rotation). This defines an out-of-domain generalization task, whilst at the same time ensuring that the model does not need to extrapolate to unseen feature values.

\emph{3. Small data, constant density (CD).} We create train/test splits for datasets of \{8k, 16k, 25k, 33k, 41k, 49k, 57k, 65k\} examples by constraining the range of feature values (Figure~\ref{fig:generalization}, 2\textsuperscript{nd} from right), ensuring that the density in the feature space remains constant as we reduce the amount of data. 

\emph{4. Small data, constant volume (CV).} Finally, we create train/test splits by selecting \{8k, 16k, 25k, 33k, 41k, 49k, 57k, 65k\} training examples at random without replacement (Figure~\ref{fig:generalization}, 1\textsuperscript{st} from right). This reduces the amount of training data whilst keeping the volume fixed, which increases the typical distance between training and test examples.

\subsection{Network Architectures and Training}

We use ReLU activations for both fully-connected and convolutional networks with a Bernoulli likelihood in the decoder\footnote{The Bernoulli likelihood is a very common choice in the VAE literature even for input domain of $[0,1]$. For a more detailed discussion, see Appendix~\ref{app:experiment-settings}}. We use a 10-dimensional latent space and assume a spherical Gaussian prior. All models are trained for 257k iterations with Adam using a batch size of 128, with 5 random restarts. 
For MLP architectures, we keep the number of hidden units fixed to 512 across layers. For the CNN architectures, we use 64 channels with kernel size 4 and stride 2 across layers. 
See Appendix~\ref{app:experiment-settings} for further details. 
\begin{figure*}[!t]
    \centering
    \includegraphics[width=0.9\linewidth]{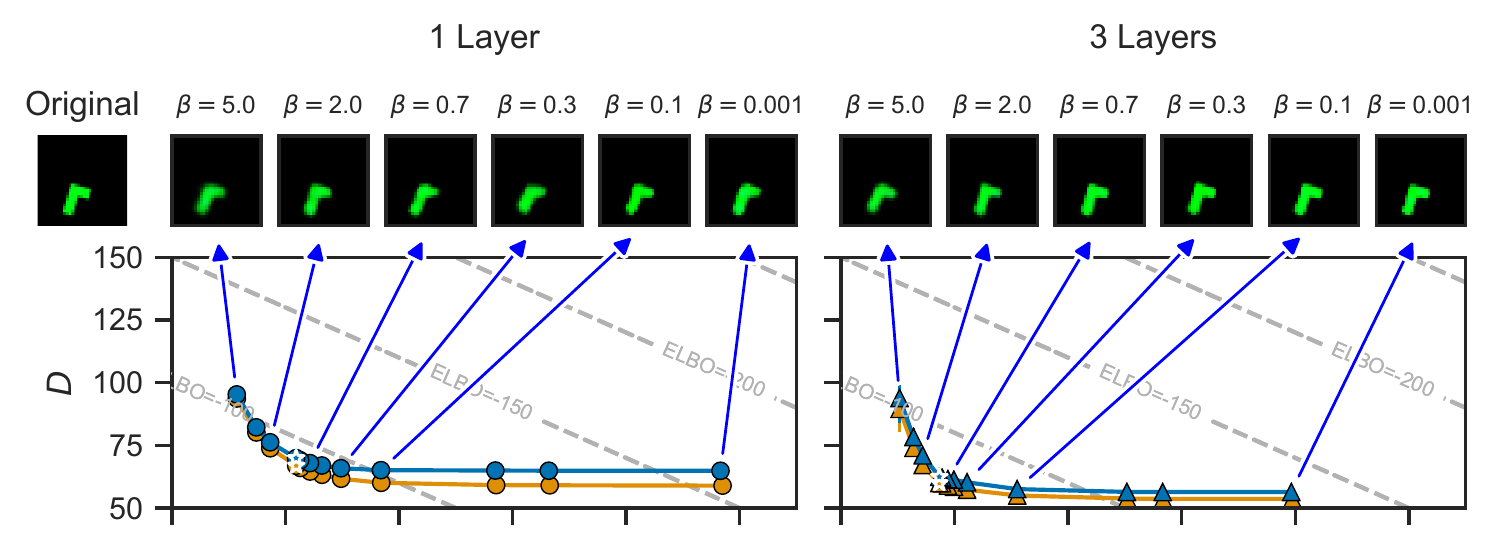}
    \includegraphics[width=0.9\linewidth]{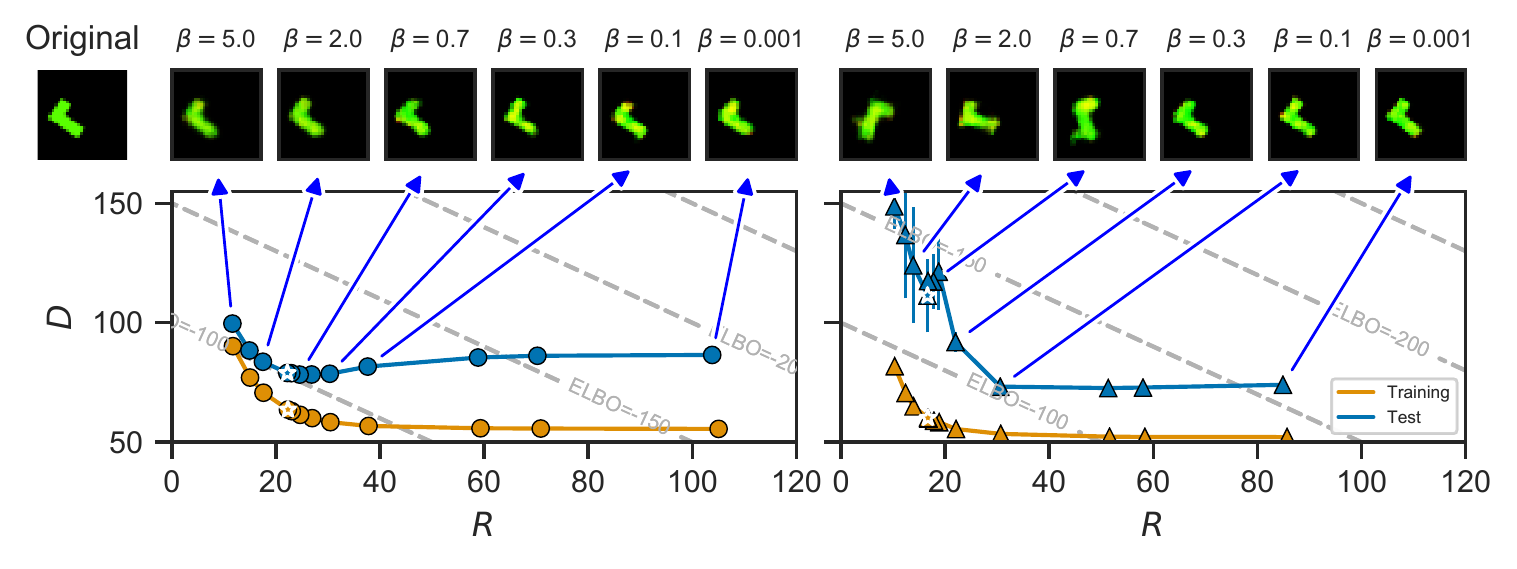}
    \vspace*{-2ex}
    \caption{Training and test $RD$ curves evaluated on the CV(82k/82k) split (\emph{Top}) and CV(16k/147k) split (\emph{bottom}), for a 1-layer and a 3-layer architecture. Each dot constitutes a $\beta$ value (white stars indicate the $\beta$=1), averaged over 5 restarts. Images show reconstructions of a test example.}
    \label{fig:rd-recons-curves}
\end{figure*}


\subsection{Results}
\label{sec:exp1}

\paragraph{Fully-Connected Architectures on Dense and Sparse Data.} We begin with a comparison between 1-layer and 3-layer fully-connected architectures on a dense CV (82k/82k) split and a sparser CV (16k/147k) split. Based on existing work~\citep{uhler}, our hypothesis in this experiment is that the 3-layer architecture will be more prone to overfitting the training data (particularly in the sparser case), and our goal is to establish to what extent rate-regularization affects the degree of overfitting.   


Figure~\ref{fig:rd-recons-curves} shows $RD$ curves on the training and test set. We report the mean across 5 restarts, with bars indicating the standard deviation, for 12 $\beta$ values\footnote{$\beta$ $\in$ \{0.001, 0.005, 0.01, 0.1, 0.3, 0.5, 0.7, 0.9, 1., 2., 3., 5.\}}. White stars mark the position of the standard VAE ($\beta$=1) on the $RD$ plane. Diagonal lines show iso-contours of the evidence lower bound $\mathcal{L}_{\beta=1}=-D-R$.
Above each panel, we show reconstructions for a test-set example that is difficult to reconstruct, in the sense that it falls into the 90\textsuperscript{th} percentile in terms of the $\ell_2$-distance between its nearest neighbor in the training set.

For the dense CV (82k/82k) split (\emph{top}), we observe no evidence of memorization. Moreover, increasing model capacity uniformly improves generalization, in the sense that it decreases both the rate and the distortion, shifting the curve to the bottom left.

For the sparse CV (16k/147k) split (\emph{bottom}), we see a different pattern. In the 1-layer model, we observe a trend that appears consistent with a classic bias-variance trade-off. The distortion on the training set decreases monotonically as we reduce $\beta$, whereas the distortion on the test set initially decreases, achieves a minimum, and somewhat increases afterwards. This suggests that $\beta$ may control a trade-off between overfitting and underfitting, although there is no indication of data memorization. When we perform early stopping (see Appendix~\ref{app:sec:early-stopping}), the RD curve once again becomes monotonic, which is consistent with this interpretation in terms of overfitting.

\begin{figure*}[!t]
    \centering
    \includegraphics[width=\linewidth]{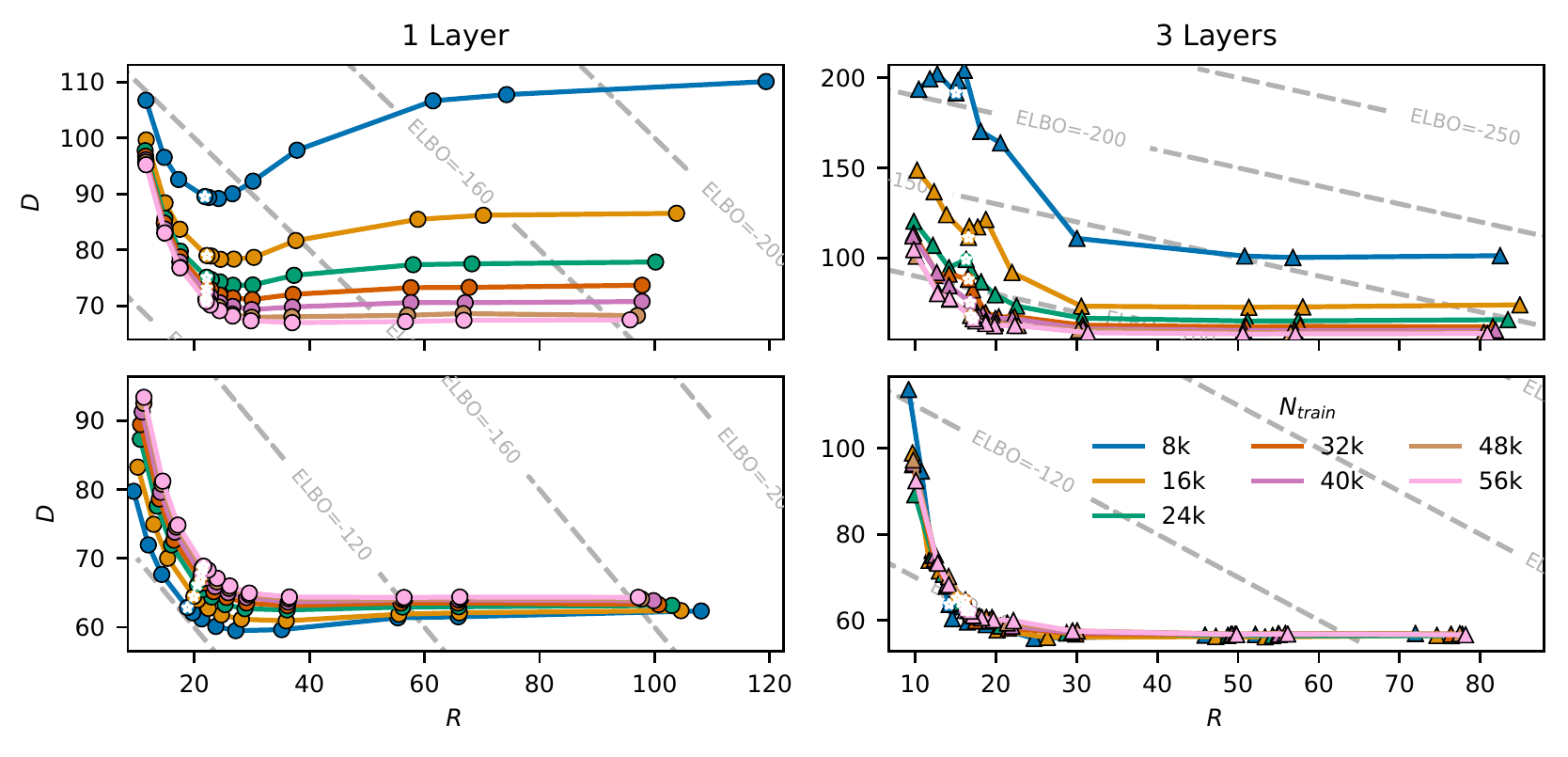}
    \vspace*{-4.5ex}
    \caption{Test-set $RD$ curves for constant volume (\emph{top}) and constant density splits (\emph{bottom}) with varying training set sizes. White stars indicate the $RD$ value for a standard VAE ($\beta$=1).}
    \vspace*{-1.5ex}
    \label{fig:datasize-rd}
\end{figure*}

In the 3-layer architecture, we observe a qualitatively different trend. Here we see evidence of data memorization; some reconstructions resemble memorized neighbors in the training set. However, counterintuitively, no memorization is apparent at smaller $\beta$ values. When looking at the iso-contours, we observe that the test-set lower bound $\mathcal{L}_{\beta=1} \le \log \p(\x)$ achieves a maximum at $\beta=0.1$. Additional analysis (see Appendix~\ref{app:sec:elbo-lm}) shows that this maximum also corresponds to the maximum of the log marginal likelihood $\log \p(\x)$. In short, high-capacity networks are capable of memorizing the training data, as expected. However, paradoxically, this memorization occurs when $\beta$ is large, where we would expect underfitting based on the 1-layer results, and the generalization gap, in terms of both $D$ and $\log \p(\x)$, is smallest at $\beta=0.1$.

\begin{figure}[!b]
\includegraphics[width=\linewidth]{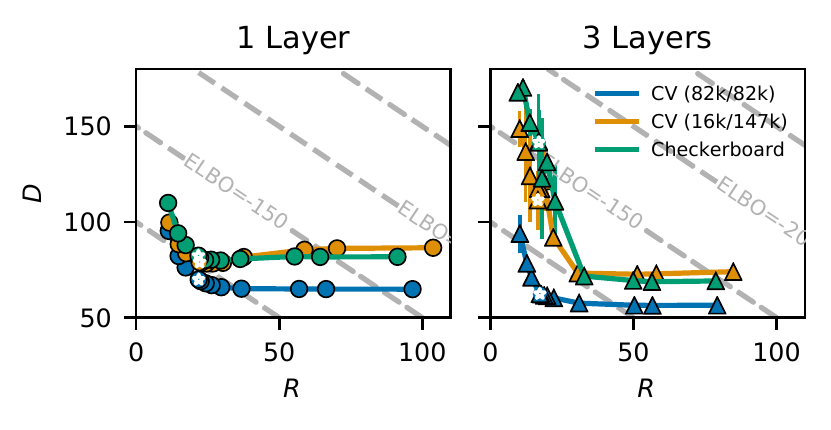}
\vspace*{-4ex}
\caption{$RD$ Curves for the CV(82k/82k), CV(16k/147k), and Checkerboard Splits.}
\label{fig:split-comparison}
\end{figure}

\paragraph{Role of the Training Set Size.}
The qualitative discrepancy between training and test set $RD$ curves in Figure~\ref{fig:rd-recons-curves} has to our knowledge not previously been reported. One possible reason for this is that this behavior would not have been apparent in other experiments; there is virtually no generalization gap in the dense CV (82k/82k) split. The differences between 1-layer and 3-layer architectures become visible in the sparse CV (16k/147k) split. Whereas the dense CV (82k/82k) split is representative of typically simulated datasets in terms of the number of examples and density in the latent space, the CV\nobreak\ (16k/147k) split has a training set that is tiny by deep learning standards. Therefore, we need to verify that the observed effects are not simply attributable to the size of the training set.

To disambiguate between effects that arise from the size of the data and effects that arise due to the density of the data, we compare CV and CD splits with training set sizes $N_\text{train} = \{8\text{k}, 16\text{k}, 32\text{k}, 56\text{k}\}$. Since CD splits have a fixed density rather than a fixed volume, the examples in the test set will be closer to their nearest neighbors in the training set, resulting in an easier generalization problem.

\begin{figure}[!b]
\includegraphics[width=\linewidth]{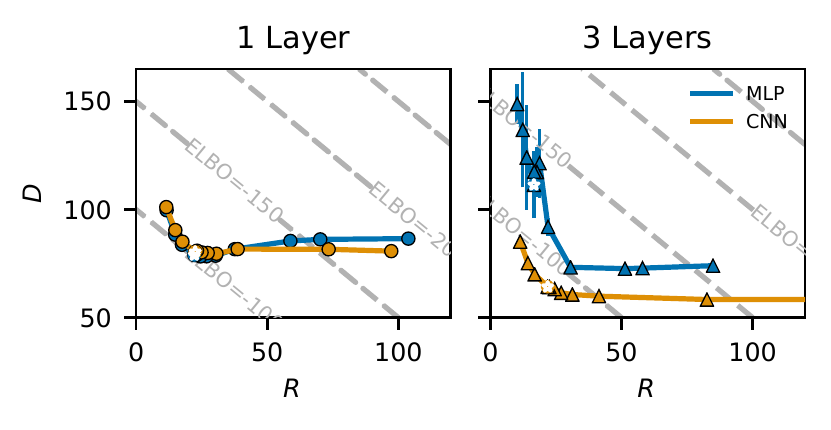}
\vspace*{-4ex}
\caption{$RD$ Curves for the CV(16k/147k) for MLP and CNN Architectures.}
\label{fig:mlp-vs-cnn}
\end{figure}

\begin{figure*}[!t]
\centering
\includegraphics[width=\linewidth]{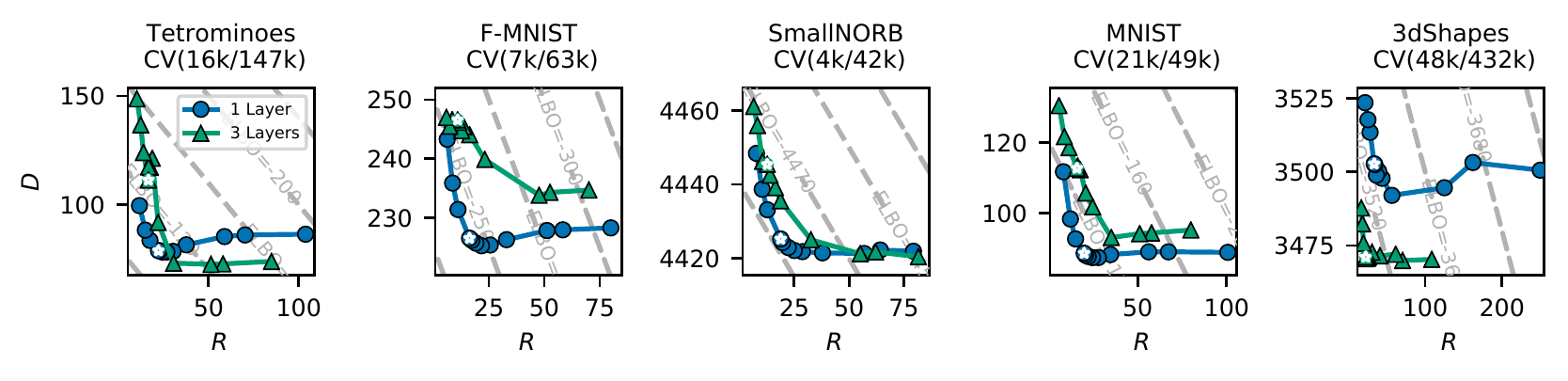}
\vspace*{-6ex}
\caption{$RD$ curves shown on various datasets trained with 1 and 3 layers.}
\vspace*{-1ex}
\label{fig:other-datastes}
\end{figure*}
Figure~\ref{fig:datasize-rd} shows the test-set $RD$ curves for this experiment. In the CV splits, the qualitative discrepancy between 1-layer and 3-layer networks becomes more pronounced as we decrease the size of the training set. However, in the CD splits, discrepancies are much less pronounced. $RD$ curves for 3-layer networks are virtually indistinguishable. $RD$ curves for 1-layer networks still exhibit a minimum, but there is a much weaker dependence on the training set size. Moreover, generalization performance marginally improves as we decrease the size of the training set. This may be attributable to the manner in which we construct the splits. Because we simulate data using a 5-dimensional hypercube of generative factors, limiting the volume has the effect of decreasing the surface to volume ratio, which would mildly reduce the typical distance between training and test set examples.



\paragraph{In-Sample and Out-of-Sample Generalization.} A possible takeaway from the results in Figure~\ref{fig:datasize-rd} is that the amount of training data itself does not strongly affect generalization performance, but that the similarity between test and training set examples does. To further test this hypothesis, we compare the CV (82k/82k) and CV (16k/147k) splits to the Checkerboard (82k/82k) split, which allows us to evaluate out-of-sample generalization to unseen combinations of factors. $RD$ curves in Figure~\ref{fig:split-comparison} show similar generalization performance for the Checkerboard and CV (16k/147k) splits. This is consistent with the fact that these splits have a similar distribution over pixel-distances between test set and nearest training set examples (Figure~\ref{app:fig:tetris-l2-hists}).

\paragraph{Convolutional architectures.} A deliberate limitation of our experiments is that we have considered fully-connected networks, which are an extremely simple architecture. There are of course many other encoder and decoder architectures for VAEs~\citep{kingma2016improved,gulrajani2016pixelvae,pixelcnn}. In Figure~\ref{fig:mlp-vs-cnn}, we compare $RD$ curves for MLPs with those for 1-layer and 3-layer CNNs (see Table~\ref{app:tab:experiment-settings} for details). We observe a monotonic curve for 3-layer CNNs and only a small degree of non-monotonicity in the 1-layer CNN. Since most architectures will have a higher capacity than a 3-layer MLP or CNN, we can interpret the results for 3-layer networks as the most representative of other architectures.

\paragraph{Additional Datasets.} Our analysis thus far
shows that the generalization gap grows when we increase the difficulty of a generalization problem, which is expected. The unexpected result is that, depending on model capacity, we either observe U-shaped $RD$ curves that are consistent with a bias-variance trade-off, or L-shaped curves in which generalization improves as we reduce $\beta$. To test whether both phenomena also occur in other datasets, we perform experiments on the Fashion-MNIST~\citep{fmnist}, SmallNORB~\citep{smallnorb}, MNIST~\citep{mnist}, and 3dShapes~\citep{3dshapes} datasets. 

We show the full results of this analysis for a range of CV splits in Appendix~\ref{app:sec:extra-datasets}. In Figure~\ref{fig:other-datastes} we compare 1-layer and 3-layer networks for a single split with a small training set for each dataset. We see that the $RD$ curves for the 1-layer network exhibits a local minimum in most datasets. Curves for the 3-layer network are generally closer to monotonic, although a more subtle local minimum is visible in certain cases. The one exception is the 3dShapes dataset, where the 3-layer network exhibits a more pronounced local minimum than the 1-layer network.

\begin{figure}[!b]
\centering
\includegraphics[width=0.9\linewidth]{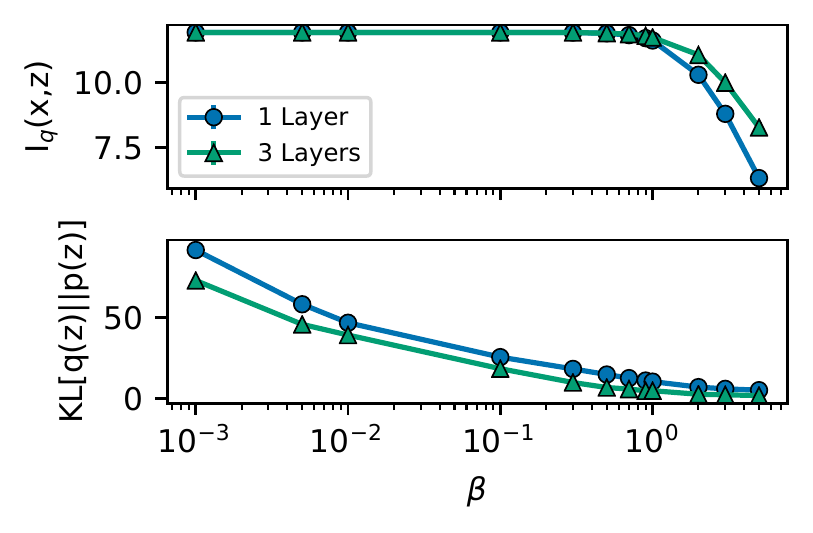}
\vspace*{-2.5ex}
\caption{$I_{q}(\x, \z)$ and $\KL{\q(\z)}{p(\z)}$ vs $\beta$ for $\beta$-VAE Trained on CV(16k/147k). 
}
\label{fig:DvsKLandIxz}
\end{figure}

\subsection{Is the Rate a Regularizer?}
\label{sec:mi-vs-kl}
Our experiments suggest that the rate is not an inductive bias that typically reduces the reconstruction loss in high-capacity models. One possible explanation for these findings is that we should consider both terms in the rate $R=I_q[\x; \z] + \KL{\q(\z)}{p(\z)}$ when evaluating the effect of rate-regularization. The term $I_q[\x ; \z]$ admits a clear interpretation as a regularizer~\citep{shamir2010learning}. However, $\KL{\q(\z)}{p(\z)}$ is not so much a regularizer as a constraint that the aggregate posterior $\q(\z)$ should resemble the prior $p(\z)$, which may require a less smooth encoder and decoder when learning a mapping from a multimodal data distribution to a unimodal prior. While we have primarily concerned ourselves with continuous factors for a single Tetramino shape, it is of course common to fit VAEs to multimodal data, particularly when the data contains distinct classes. A unimodal prior forces the VAE to learn a decoder that ``partitions'' the contiguous latent space into regions associated with each class, which will give rise to sharp gradients near class boundaries.



To understand how each of these two terms contributes to the rate, we compute estimates of $I_q(\x ;\z)$ and $\KL{\q(\z)}{p(\z)}$ by approximating $\q(\z)$ with a Monte Carlo estimate over batches of size 512 (see \citet{pmlr-v89-esmaeili19a}). Figure~\ref{fig:DvsKLandIxz} shows both estimates as a function of $\beta$ for the CV (16k/147k) split. As expected, $I_q(\x ; \z)$ decreases when $\beta > 1$ but saturates to its maximum $\log N_\text{train}$ when $\beta < 1$. Conversely, the term $\KL{\q(\z)}{p(\z)}$ is small when $\beta > 1$ but increases when $\beta < 1$. Based on the fact that the generalization gap in terms of both the reconstruction loss and $\log\p{\x}$ is minimum at $\beta=0.1$, it appears that the $\KL{\q(\z)}{p(\z)}$ term can have a significant effect on generalization performance. 
Additional experiments where we train VAEs with either the marginal KL or the MI term removed from the loss function confirm this effect of the marginal KL term on the generalization performance of VAEs (see Appendix~\ref{app:sec:mi_kl}).


Our reading of these results is that it is reasonable to interpret the rate as an approximation of the MI when $\beta$ is large. However, our experiments suggest that VAEs typically underfit in this regime, and therefore do not benefit from this form of regularization. When $\beta$ is small, the MI saturates and we can approximate the rate as $R = \log \ntrain{} + \KL{\q(\z)}{p(\z)}$. In this regime, we should not interpret the rate as a regularizer, but as a constraint on the learned representation, and there can be a trade-off between this constraint and the reconstruction accuracy.

\section{Discussion}
In this empirical study, we trained over 6000 VAE instances to evaluate how rate-regularization in the VAE objective affects generalization to unseen examples. Our results demonstrate that high-capacity VAEs can and do overfit the training data. However, paradoxically, memorization effects can be mitigated by decreasing $\beta$. These effects are more pronounced when test-set examples differ substantially from their nearest neighbors in the training set. For real-world datasets, this is likely to be the norm rather than the exception; few datasets have a small number of generative factors. 



Based on these results, we argue that we should give the role of priors as inductive biases in VAEs more serious consideration. The KL relative to a standard Gaussian prior does not improve generalization performance in the majority of cases. With the benefit of hindsight, this is unsurprising; When we use a VAE to model a fundamentally multimodal data distribution, then mapping this data onto a contiguous unimodal Gaussian prior may not yield a smooth encoder, semantically meaningful distances in the latent space, or indeed a representation that generalizes to unseen data. This motivates future work to determine to what extent other priors, including priors that attempt to induce structured or disentangled representations, can aid generalization performance.

While these experiments are comprehensive, we have explicitly constrained ourselves to comparatively simple architectures and datasets. These architectures are not representative of the state of the art~\citep{vahdat_nvae_nodate,maaloe2019biva,razavi_generating_2019,gulrajani2016pixelvae,pixelcnn}, particularly when we are primarily interested in generation. It remains an open question to what extent rate-regularization affects generalization in much higher-capacity architectures that are trained on larger datasets of natural images. Moreover, there are other factors that could potentially impact our results which we do not study here, including but not limited to: dimensionality of the latent space, the choice of prior, and the choice of training method. We leave the investigation of these factors in $RD$ analysis to future work. 

\subsubsection*{Acknowledgements}

We would like to thank reviewers of a previous version of this manuscript for their detailed comments, as well as Sarthak Jain and Heiko Zimmermann for helpful discussions. This project was supported by the Intel Corporation, the 3M Corporation, the Air Force Research Laboratory (AFRL) and DARPA, NSF grant 1901117, NIH grant R01CA199673 from NCI, and startup funds from Northeastern University.

\bibliographystyle{abbrvnat}
\bibliography{aistats_2021}

\begin{thebibliography}{42}
\providecommand{\natexlab}[1]{#1}
\providecommand{\url}[1]{\texttt{#1}}
\expandafter\ifx\csname urlstyle\endcsname\relax
  \providecommand{\doi}[1]{doi: #1}\else
  \providecommand{\doi}{doi: \begingroup \urlstyle{rm}\Url}\fi

\bibitem[Alemi et~al.(2018)Alemi, Poole, Fischer, Dillon, Saurous, and
  Murphy]{alemi2018fixing}
A.~Alemi, B.~Poole, I.~Fischer, J.~Dillon, R.~A. Saurous, and K.~Murphy.
\newblock Fixing a broken {ELBO}.
\newblock In \emph{International Conference on Machine Learning}, pages
  159--168, 2018.

\bibitem[Alemi et~al.(2017)Alemi, Fischer, Dillon, and Murphy]{alemi2017deep}
A.~A. Alemi, I.~Fischer, J.~V. Dillon, and K.~Murphy.
\newblock Deep {{Variational Information Bottleneck}}.
\newblock \emph{International Conference on Learning Representations}, 2017.

\bibitem[Berthelot et~al.(2018)Berthelot, Raffel, Roy, and
  Goodfellow]{berthelot2018understanding}
D.~Berthelot, C.~Raffel, A.~Roy, and I.~Goodfellow.
\newblock Understanding and improving interpolation in autoencoders via an
  adversarial regularizer.
\newblock \emph{arXiv preprint arXiv:1807.07543}, 2018.

\bibitem[Bousquet et~al.(2017)Bousquet, Gelly, Tolstikhin, Simon-Gabriel, and
  Schoelkopf]{bousquet2017optimal}
O.~Bousquet, S.~Gelly, I.~Tolstikhin, C.-J. Simon-Gabriel, and B.~Schoelkopf.
\newblock From optimal transport to generative modeling: the {VEGAN} cookbook.
\newblock \emph{arXiv preprint arXiv:1705.07642}, 2017.

\bibitem[Burgess and Kim(2018)]{3dshapes}
C.~Burgess and H.~Kim.
\newblock {3D} shapes dataset.
\newblock https://github.com/deepmind/3dshapes-dataset/, 2018.

\bibitem[Chen et~al.(2019)Chen, Ferroni, Klushyn, Paraschos, Bayer, and van~der
  Smagt]{chen2019fast}
N.~Chen, F.~Ferroni, A.~Klushyn, A.~Paraschos, J.~Bayer, and P.~van~der Smagt.
\newblock Fast approximate geodesics for deep generative models.
\newblock In \emph{International Conference on Artificial Neural Networks},
  pages 554--566. Springer, 2019.

\bibitem[Chen et~al.(2018)Chen, Li, Grosse, and Duvenaud]{chen2018isolating}
T.~Q. Chen, X.~Li, R.~B. Grosse, and D.~K. Duvenaud.
\newblock Isolating sources of disentanglement in variational autoencoders.
\newblock In \emph{Advances in Neural Information Processing Systems}, pages
  2610--2620, 2018.

\bibitem[Chen et~al.(2016)Chen, Kingma, Salimans, Duan, Dhariwal, Schulman,
  Sutskever, and Abbeel]{chen2016variational}
X.~Chen, D.~P. Kingma, T.~Salimans, Y.~Duan, P.~Dhariwal, J.~Schulman,
  I.~Sutskever, and P.~Abbeel.
\newblock Variational lossy autoencoder.
\newblock \emph{arXiv preprint arXiv:1611.02731}, 2016.

\bibitem[Cover and Thomas(2012)]{cover2012elements}
T.~M. Cover and J.~A. Thomas.
\newblock \emph{Elements of information theory}.
\newblock John Wiley \& Sons, 2012.

\bibitem[Eastwood and Williams(2018)]{eastwood}
C.~Eastwood and C.~K.~I. Williams.
\newblock A {{Framework}} for the {{Quantitative Evaluation}} of {{Disentangled
  Representations}}.
\newblock In \emph{International Conference on Learning Representations}, Feb.
  2018.

\bibitem[Engel et~al.(2017)Engel, Hoffman, and Roberts]{engel2017latent}
J.~Engel, M.~Hoffman, and A.~Roberts.
\newblock Latent {{Constraints}}: {{Learning}} to {{Generate Conditionally}}
  from {{Unconditional Generative Models}}.
\newblock \emph{arXiv:1711.05772 [cs, stat]}, Nov. 2017.

\bibitem[Esmaeili et~al.(2019)Esmaeili, Wu, Jain, Bozkurt, Siddharth, Paige,
  Brooks, Dy, and van~de Meent]{pmlr-v89-esmaeili19a}
B.~Esmaeili, H.~Wu, S.~Jain, A.~Bozkurt, N.~Siddharth, B.~Paige, D.~H. Brooks,
  J.~Dy, and J.-W. van~de Meent.
\newblock Structured disentangled representations.
\newblock In K.~Chaudhuri and M.~Sugiyama, editors, \emph{Proceedings of
  Machine Learning Research}, volume~89 of \emph{Proceedings of Machine
  Learning Research}, pages 2525--2534. PMLR, 16--18 Apr 2019.

\bibitem[Ghosh et~al.(2019)Ghosh, Sajjadi, Vergari, Black, and
  Sch{\"o}lkopf]{ghosh2019variational}
P.~Ghosh, M.~S. Sajjadi, A.~Vergari, M.~Black, and B.~Sch{\"o}lkopf.
\newblock From variational to deterministic autoencoders.
\newblock \emph{arXiv preprint arXiv:1903.12436}, 2019.

\bibitem[Gulrajani et~al.(2017)Gulrajani, Kumar, Ahmed, Taiga, Visin, Vazquez,
  and Courville]{gulrajani2016pixelvae}
I.~Gulrajani, K.~Kumar, F.~Ahmed, A.~A. Taiga, F.~Visin, D.~Vazquez, and
  A.~Courville.
\newblock Pixelvae: A latent variable model for natural images.
\newblock In \emph{International {Conference} on {Representations}}, 2017.

\bibitem[Heusel et~al.(2017)Heusel, Ramsauer, Unterthiner, Nessler, and
  Hochreiter]{fid}
M.~Heusel, H.~Ramsauer, T.~Unterthiner, B.~Nessler, and S.~Hochreiter.
\newblock {GANs} trained by a two time-scale update rule converge to a local
  {Nash} equilibrium.
\newblock In \emph{Advances in neural information processing systems}, pages
  6626--6637, 2017.

\bibitem[Higgins et~al.(2017)Higgins, Matthey, Pal, Burgess, Glorot, Botvinick,
  Mohamed, and Lerchner]{higgins2017beta}
I.~Higgins, L.~Matthey, A.~Pal, C.~Burgess, X.~Glorot, M.~Botvinick,
  S.~Mohamed, and A.~Lerchner.
\newblock beta-{VAE}: Learning basic visual concepts with a constrained
  variational framework.
\newblock In \emph{International Conference on Learning Representations}, 2017.

\bibitem[Hoffman et~al.(2017)Hoffman, Riquelme, and Johnson]{hoffman2017v-vaes}
M.~D. Hoffman, C.~Riquelme, and M.~J. Johnson.
\newblock The {$\beta$}-{{VAE}}'s {{Implicit Prior}}.
\newblock In \emph{Workshop on {{Bayesian Deep Learning}}, {{NIPS}}}, pages
  1--5, 2017.

\bibitem[Huang et~al.(2020)Huang, Makhzani, Cao, and Grosse]{huangevaluating}
S.~Huang, A.~Makhzani, Y.~Cao, and R.~Grosse.
\newblock Evaluating lossy compression rates of deep generative models.
\newblock \emph{arXiv preprint arXiv:2008.06653}, 2020.

\bibitem[Kim and Mnih(2018)]{kim2018disentangling}
H.~Kim and A.~Mnih.
\newblock Disentangling by factorising.
\newblock In \emph{International Conference on Machine Learning}, pages
  2654--2663, 2018.

\bibitem[Kingma and Welling(2013)]{kingma2013auto}
D.~P. Kingma and M.~Welling.
\newblock Auto-encoding variational bayes.
\newblock \emph{International Conference on Learning Representations}, 2013.

\bibitem[Kingma et~al.(2016)Kingma, Salimans, Jozefowicz, Chen, Sutskever, and
  Welling]{kingma2016improved}
D.~P. Kingma, T.~Salimans, R.~Jozefowicz, X.~Chen, I.~Sutskever, and
  M.~Welling.
\newblock Improved variational inference with inverse autoregressive flow.
\newblock In \emph{Advances in neural information processing systems}, pages
  4743--4751, 2016.

\bibitem[Kumar and Poole(2020)]{kumar2020implicit}
A.~Kumar and B.~Poole.
\newblock On implicit regularization in $\beta$-vaes.
\newblock \emph{arXiv preprint arXiv:2002.00041}, 2020.

\bibitem[LeCun et~al.(1998)LeCun, Bottou, Bengio, Haffner, et~al.]{mnist}
Y.~LeCun, L.~Bottou, Y.~Bengio, P.~Haffner, et~al.
\newblock Gradient-based learning applied to document recognition.
\newblock \emph{Proceedings of the IEEE}, 86\penalty0 (11):\penalty0
  2278--2324, 1998.

\bibitem[LeCun et~al.(2004)LeCun, Huang, and Bottou]{smallnorb}
Y.~LeCun, F.~J. Huang, and L.~Bottou.
\newblock Learning methods for generic object recognition with invariance to
  pose and lighting.
\newblock In \emph{Proceedings of the 2004 IEEE Computer Society Conference on
  Computer Vision and Pattern Recognition, 2004. CVPR 2004.}, volume~2, pages
  II--104. IEEE, 2004.

\bibitem[Liang et~al.(2018)Liang, Krishnan, Hoffman, and
  Jebara]{liang2018variational}
D.~Liang, R.~G. Krishnan, M.~D. Hoffman, and T.~Jebara.
\newblock Variational {{Autoencoders}} for {{Collaborative Filtering}}.
\newblock In \emph{Proceedings of the 2018 {{World Wide Web Conference}}},
  {{WWW}} '18, pages 689--698, {Lyon, France}, Apr. 2018. {International World
  Wide Web Conferences Steering Committee}.
\newblock ISBN 978-1-4503-5639-8.
\newblock \doi{10.1145/3178876.3186150}.

\bibitem[Locatello et~al.(2019)Locatello, Bauer, Lucic, Raetsch, Gelly,
  Sch{\"o}lkopf, and Bachem]{francesco}
F.~Locatello, S.~Bauer, M.~Lucic, G.~Raetsch, S.~Gelly, B.~Sch{\"o}lkopf, and
  O.~Bachem.
\newblock Challenging common assumptions in the unsupervised learning of
  disentangled representations.
\newblock In \emph{International Conference on Machine Learning}, pages
  4114--4124, 2019.

\bibitem[Maal{\o}e et~al.(2019)Maal{\o}e, Fraccaro, Lievin, and
  Winther]{maaloe2019biva}
L.~Maal{\o}e, M.~Fraccaro, V.~Lievin, and O.~Winther.
\newblock Biva: A very deep hierarchy of latent variables for generative
  modeling.
\newblock In \emph{33rd Conference on Neural Information Processing Systems},
  page 8882. Neural Information Processing Systems Foundation, 2019.

\bibitem[Matthey et~al.(2017)Matthey, Higgins, Hassabis, and
  Lerchner]{dsprites17}
L.~Matthey, I.~Higgins, D.~Hassabis, and A.~Lerchner.
\newblock dsprites: Disentanglement testing sprites dataset.
\newblock https://github.com/deepmind/dsprites-dataset/, 2017.

\bibitem[Narayanaswamy et~al.(2017)Narayanaswamy, Paige, Van~de Meent,
  Desmaison, Goodman, Kohli, Wood, and Torr]{narayanaswamy2017learning}
S.~Narayanaswamy, T.~B. Paige, J.-W. Van~de Meent, A.~Desmaison, N.~Goodman,
  P.~Kohli, F.~Wood, and P.~Torr.
\newblock Learning disentangled representations with semi-supervised deep
  generative models.
\newblock In \emph{Advances in Neural Information Processing Systems}, pages
  5925--5935, 2017.

\bibitem[Radhakrishnan et~al.(2019)Radhakrishnan, Yang, Belkin, and
  Uhler]{uhler}
A.~Radhakrishnan, K.~Yang, M.~Belkin, and C.~Uhler.
\newblock Memorization in overparameterized autoencoders.
\newblock \emph{arXiv preprint arXiv:1810.10333v3}, 2019.

\bibitem[Razavi et~al.(2019)Razavi, van~den Oord, and
  Vinyals]{razavi_generating_2019}
A.~Razavi, A.~van~den Oord, and O.~Vinyals.
\newblock Generating {Diverse} {High}-{Fidelity} {Images} with {VQ}-{VAE}-2.
\newblock In H.~Wallach, H.~Larochelle, A.~Beygelzimer, F.~d. Alché-Buc,
  E.~Fox, and R.~Garnett, editors, \emph{Advances in {Neural} {Information}
  {Processing} {Systems}}, volume~32. Curran Associates, Inc., 2019.
\newblock URL
  \url{https://proceedings.neurips.cc/paper/2019/file/5f8e2fa1718d1bbcadf1cd9c7a54fb8c-Paper.pdf}.

\bibitem[Rezende and Viola(2018)]{rezende2018taming}
D.~J. Rezende and F.~Viola.
\newblock Taming {VAE}s.
\newblock \emph{arXiv preprint arXiv:1810.00597}, 2018.

\bibitem[Rezende et~al.(2014)Rezende, Mohamed, and
  Wierstra]{rezende2014stochastic}
D.~J. Rezende, S.~Mohamed, and D.~Wierstra.
\newblock Stochastic backpropagation and approximate inference in deep
  generative models.
\newblock In \emph{Proceedings of The 31st International Conference on Machine
  Learning}, pages 1278--1286, 2014.

\bibitem[Shamir et~al.(2010)Shamir, Sabato, and Tishby]{shamir2010learning}
O.~Shamir, S.~Sabato, and N.~Tishby.
\newblock Learning and generalization with the information bottleneck.
\newblock \emph{Theoretical Computer Science}, 411\penalty0 (29):\penalty0
  2696--2711, June 2010.
\newblock ISSN 0304-3975.
\newblock \doi{10.1016/j.tcs.2010.04.006}.

\bibitem[Shu et~al.(2018)Shu, Bui, Zhao, Kochenderfer, and
  Ermon]{shu2018amortized}
R.~Shu, H.~H. Bui, S.~Zhao, M.~J. Kochenderfer, and S.~Ermon.
\newblock Amortized inference regularization.
\newblock In \emph{Advances in Neural Information Processing Systems}, pages
  4393--4402, 2018.

\bibitem[Tishby et~al.(2000)Tishby, Pereira, and Bialek]{tishby2000information}
N.~Tishby, F.~C. Pereira, and W.~Bialek.
\newblock The information bottleneck method.
\newblock \emph{arXiv:physics/0004057}, Apr. 2000.

\bibitem[Vahdat and Kautz()]{vahdat_nvae_nodate}
A.~Vahdat and J.~Kautz.
\newblock {NVAE}: {A} {Deep} {Hierarchical} {Variational} {Autoencoder}.
\newblock page~13.

\bibitem[Van~den Oord et~al.(2016)Van~den Oord, Kalchbrenner, Espeholt,
  Vinyals, Graves, et~al.]{pixelcnn}
A.~Van~den Oord, N.~Kalchbrenner, L.~Espeholt, O.~Vinyals, A.~Graves, et~al.
\newblock Conditional image generation with pixelcnn decoders.
\newblock In \emph{Advances in neural information processing systems}, pages
  4790--4798, 2016.

\bibitem[Wen et~al.(2017)Wen, Miao, Blunsom, and Young]{wen2017latent}
T.-H. Wen, Y.~Miao, P.~Blunsom, and S.~Young.
\newblock Latent {{Intention Dialogue Models}}.
\newblock In \emph{International {{Conference}} on {{Machine Learning}}}, pages
  3732--3741, July 2017.

\bibitem[Xiao et~al.(2017)Xiao, Rasul, and Vollgraf]{fmnist}
H.~Xiao, K.~Rasul, and R.~Vollgraf.
\newblock Fashion-mnist: a novel image dataset for benchmarking machine
  learning algorithms.
\newblock \emph{arXiv preprint arXiv:1708.07747}, 2017.

\bibitem[Zhang et~al.(2019)Zhang, Bengio, Hardt, and Singer]{samy}
C.~Zhang, S.~Bengio, M.~Hardt, and Y.~Singer.
\newblock Identity crisis: Memorization and generalization under extreme
  overparameterization.
\newblock \emph{arXiv preprint arXiv:1902.04698}, 2019.

\bibitem[Zhao et~al.(2018)Zhao, Ren, Yuan, Song, Goodman, and
  Ermon]{zhao2018bias}
S.~Zhao, H.~Ren, A.~Yuan, J.~Song, N.~Goodman, and S.~Ermon.
\newblock Bias and generalization in deep generative models: An empirical
  study.
\newblock In \emph{Advances in Neural Information Processing Systems}, pages
  10815--10824, 2018.

\end{thebibliography}

\newpage
\onecolumn
\aistatstitle{Appendix for ``Rate-Regularization and Generalization in Variational Autoencoders''}
\setcounter{page}{1}
\captionsetup[figure]{list=yes}
\renewcommand\thesection{A\arabic{section}}
\setcounter{section}{0}
\renewcommand\thesubsection{\thesection.\arabic{subsection}}

\renewcommand\thetable{A\arabic{table}}
\setcounter{table}{0}
\renewcommand\thefigure{A\arabic{figure}}
\setcounter{figure}{0}

\startcontents[sections]
\printcontents[sections]{l}{1}{\setcounter{tocdepth}{2}}

\section{Experiment Setup}
\label{app:experiment-settings}

Various settings for all our experiments are displayed in Table~\ref{app:tab:experiment-settings}. We decided to keep all the hyperparameters (other than $\beta$, depth, and training set size) fixed. We use ReLU activations for both fully-connected and CNN architectures. Based on our initial experiments, where the effect of network width on generalization proved to be minor (see Figure~\ref{app:fig:vae-netcap-tetris}), we use 512 hidden units in all experiments in the main text. There are a variety of ways for changing the capacity of convolutional neural networks. In our experiments, we decided to focus on changing the number of layers and keep the other hyperparameters such number of channels, kernel size, stride, and padding fixed. In all models, we use a 10-dimensional latent space and assume a spherical Gaussian prior. All models are trained for 257k iterations with Adam (default parameters, amsgrad enabled) using a batch size of 128, with 5 random restarts. We have also performed experiments with early-stopping which we will discuss in Section~\ref{app:sec:early-stopping}.

\subsection{Implementation Details}
All experiments were ran on NVIDIA 1080Ti and Tesla V100 GPUS (depending on availability), using Pytorch 1.3.0 and ProbTorch commit f9f5c9. Most models are trained with 32-bit precision. A few models (3-layer $\beta$-VAEs with $\beta<0.1$) that didn't train were retrained using 64-bit precision.   

\subsection{Likelihood}

\begin{figure}[!h]
\centering
\includegraphics[width=\linewidth]{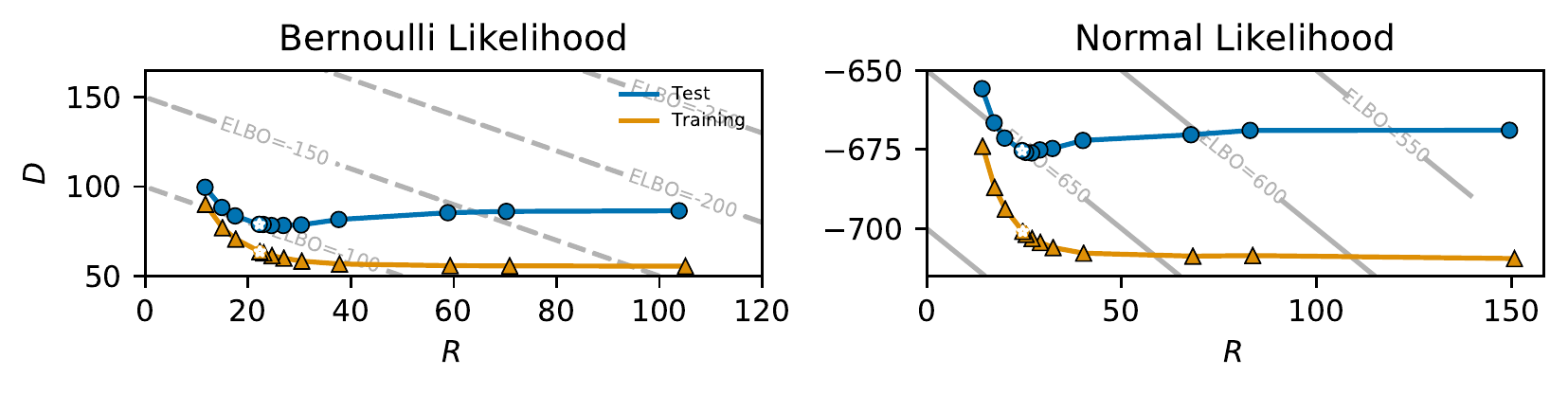}
\vspace*{-5ex}
\caption[$RD$ curves shown for Bernoulli and Gaussian likelihood.]{$RD$ curves shown for CV (16k/147k) split for VAE with MLP architecture trained with a Bernoulli likelihood and a Gaussian likelihood ($\sigma^2$ fixed to 0.1) .}
\label{app:fig:normal-likelihood}
\end{figure}

We use a standard Bernoulli likelihood in the decoder. This eliminates the extra tunable parameter $\sigma$ that is present in Gaussian decoders, which is redundant since it controls the strength of the reconstruction loss in the same manner as the $\beta$ coefficient. The Bernoulli likelihood is in fact a very common choice in the VAE literature and appears in the original VAE paper, tutorials, and in reference implementations in deep learning frameworks. Although not reported, we have verified that Gaussian decoders with fixed $\sigma$ show the same $RD$ trends as the Bernoulli decoder in the 1-layer case (see Figure \ref{app:fig:normal-likelihood}). In the 3-layer case, we observed that using a normal likelihood significantly suffers from a mode-collapse problem.

\begin{table}[!h]
\centering
\caption{Hyperparameters common to each of the considered datasets}
\label{app:tab:experiment-settings}
\resizebox{\linewidth}{!}{
\begin{tabular}{rccccc}
    \toprule
                                & Tetrominoes   & MNIST &   F-MNIST & 3dShapes  & SmallNORB     \\
    \midrule
    Batch-size                  & 128           & 128    & 128        & 128       & 128 \\
    Number of iterations        & 256k          & 256k  & 256k      & 256k      & 256k \\
    Latent space dimension      & 10            & 10    & 10        & 10        & 10 \\
    Number of hidden units (MLP) & 512           & 512   & 512       & 512       & 512 \\
    Number of channels (CNN)   & 64            & 64    & 64        & 64        & 64 \\
    Kernel size (CNN)          & 4             & 4     & 4         & 4         & 4\\
    Stride (CNN)               & 2             & 2     & 2         & 2         & 2\\
    Padding (CNN)               & 1             & 3     & 3         & 1         & 1\\
    \bottomrule
\end{tabular}}
\end{table}

\section{Tetrominoes dataset}
\label{app:sec:tetris-details}

\begin{table}[!h]
  \centering
  \caption{Names, and training and test set sizes of Tetrominoes datasets used in the paper.}
    \begin{tabular}{ccrr}
    \toprule
          \multicolumn{2}{c}{Dataset} & \multicolumn{1}{c}{Training} & \multicolumn{1}{c}{Test} \\
          \midrule
    \multicolumn{1}{c}{\multirow{7}[0]{*}{CV}}
    & 8k/157k & 8193 & 155647 \\
          & 16k/147k & 16384 & 147456 \\
          & 25k/139k & 24577 & 139263 \\
          & 33k/131k & 32768 & 131072 \\
          & 41k/123k & 40960 & 122880 \\
          & 49k/115k & 49153 & 114687 \\
          & 57k/106k & 57344 & 106496 \\
          \midrule
    Default & & 81920 & 81920 \\
    \midrule
    \multicolumn{1}{c}{\multirow{7}[0]{*}{CD}} & 8k/8k & 8159 & 8275 \\
          & 16k/16k & 16405 & 16286 \\
          & 25k/25k & 24642 & 24592 \\
          & 33k/33k & 32998 & 32754 \\
          & 41k/41k & 41138 & 40954 \\
          & 49k/49k & 49216 & 49285 \\
          & 57k/57k & 57416 & 57403 \\
          \midrule
    Checkerboard & & 82021 & 81819\\
    \bottomrule
    \end{tabular}%
  \label{app:tab:dataset_sizes}%
\end{table}%

In this section, we take a closer look at the ``difficulty'' of generalization problem in the Tetrominoes dataset. One can argue that generalization ``difficulty'' in any dataset is essentially linked to the closeness of training and test set in pixel space. This of-course depends not only on the nature of the dataset, but on size of both training and test sets. Moreover, we need to define the notion of \emph{closeness} between training and test set in advance. One approach to quantify this concept is the following: for every example in the test, what is the distance to the closest example in training set for a given distance metric? 



\begin{figure}[H]
    \centering
    \includegraphics[width=\textwidth]{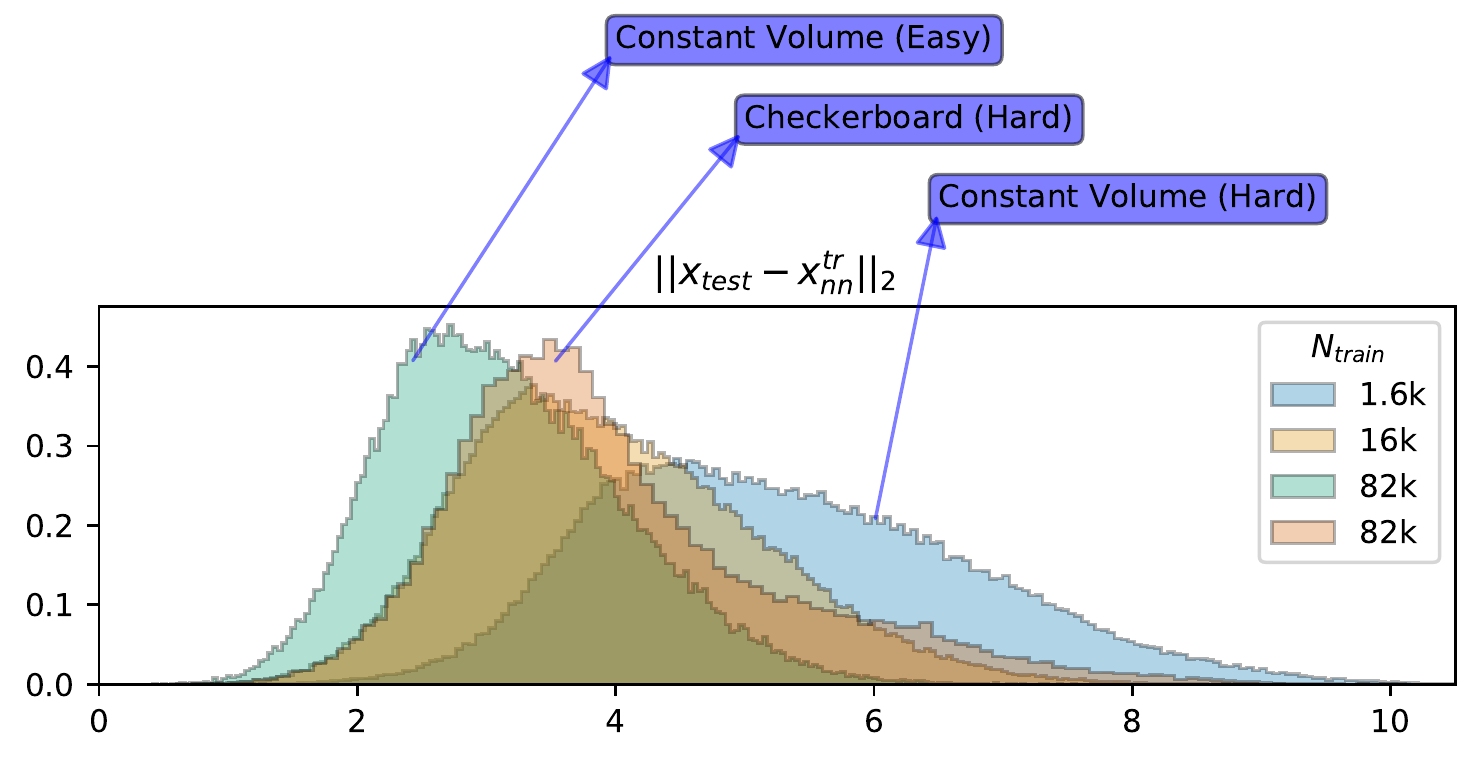}
    \includegraphics[width=\textwidth]{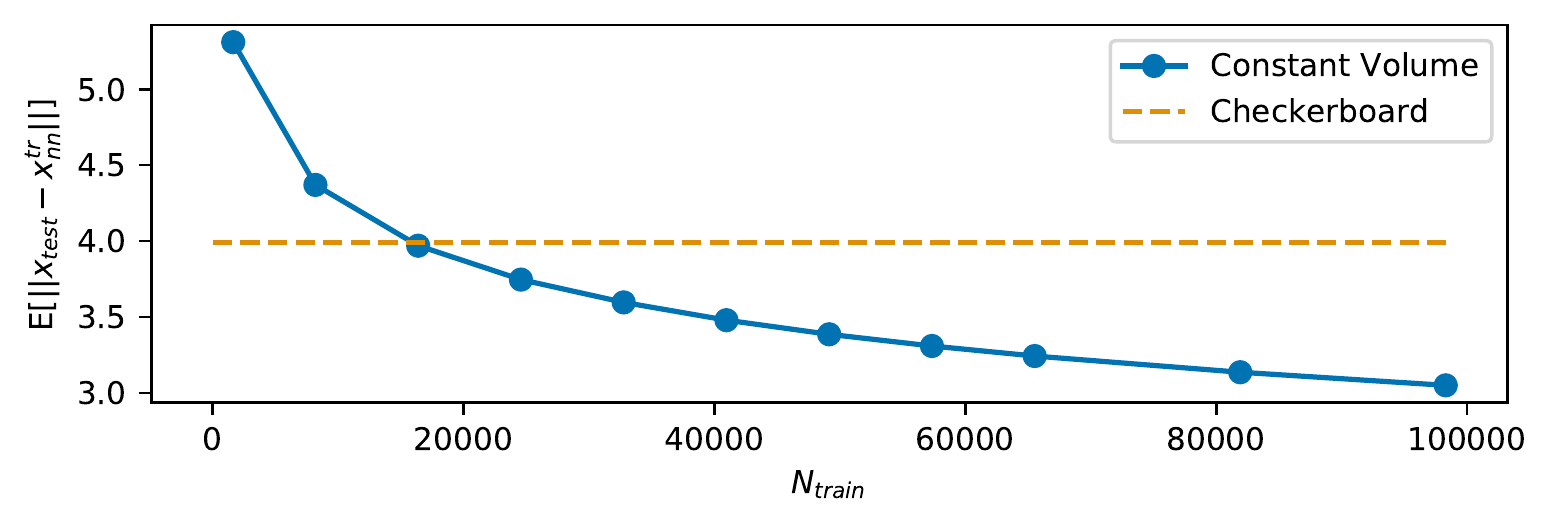}
    \vspace*{-5ex}
    \caption[Analysis of $\ell_2$ distance between training and test examples in Tetrominoes dataset]{Analysis of $\ell_{2}$ distance in pixel space between training and test set in Tetrominoes dataset based on $N_{train}$. (\emph{Top}) Normalized histograms of $\ell_{2}$ distance between test examples $\x$ and the nearest neighbour in the training set $\x^{tr}_{nn}$. Unsurprisingly, As the amount of training data increases, the distribution of $\ell_{2}$ norm between test examples and their nearest neighbour in the training set moves towards 0. (\emph{Bottom}) Mean of $\|\x-\x^{tr}_{nn}\|_{2}$ distribution as a function of train of ratio.} 
    \label{app:fig:tetris-l2-hists}
\end{figure}

\begin{figure}[H]
    \centering
    \vspace*{-3ex}
    \includegraphics[width=0.89\textwidth]{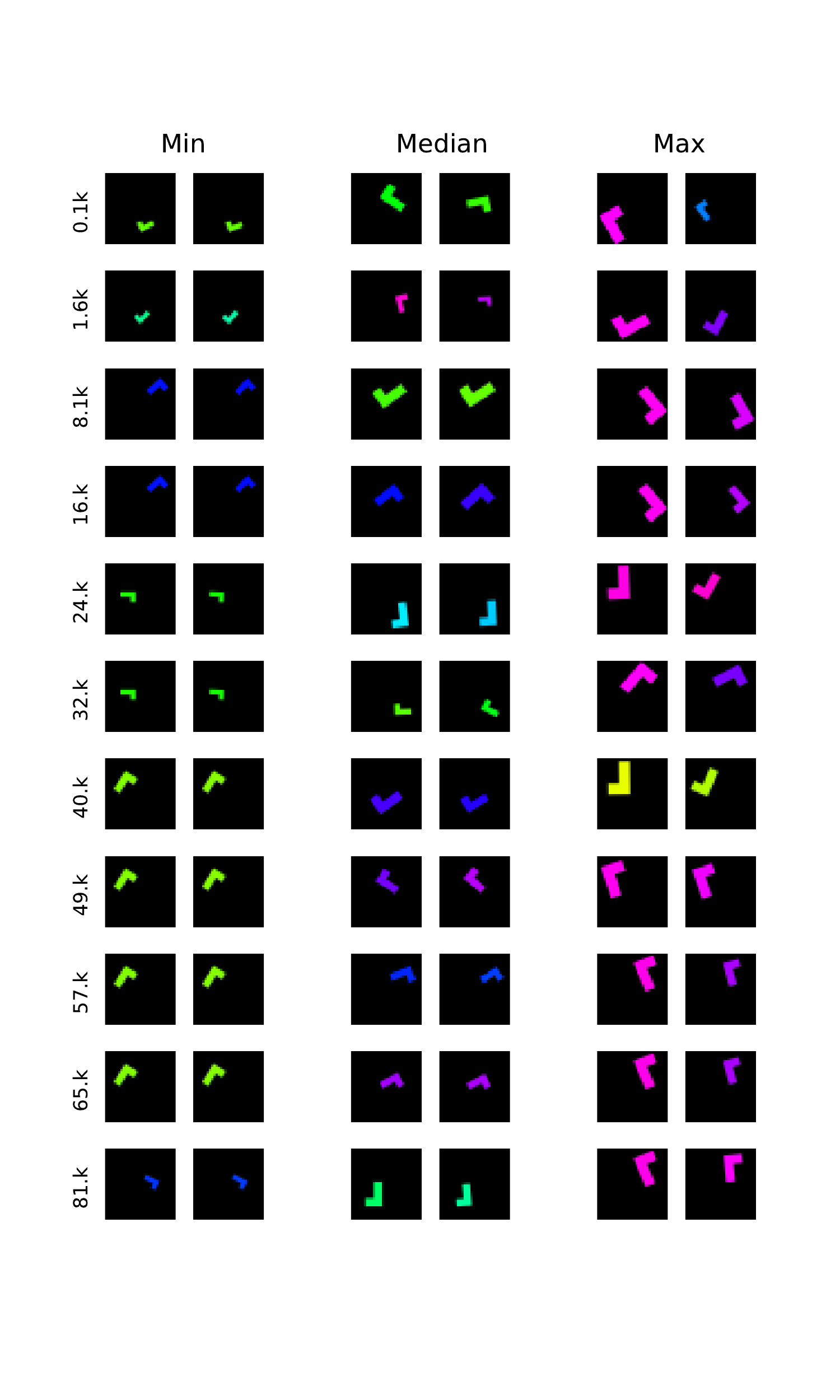}
    \vspace*{-19ex}
    \caption[Test samples and their nearest neighbours in training set for Tetrominoes]{Test samples with minimum (\emph{left}), median (\emph{middle}), and maximum (\emph{right}) $\ell_{2}$ norm between their nearest neighbour in training set, for splits with various \ntrain{}. In each column, the test sample is displayed on the left, and the nearest neighbour is displayed on the right.}
    \label{app:fig:tetris-neareset-neighbours}
\end{figure}

In Figure~\ref{app:fig:tetris-l2-hists}, we show the normalized $\ell_{2}$ histograms of test examples to their nearest neighbour in training set for different amount of training data. In Figure~\ref{app:fig:tetris-neareset-neighbours}, we show test samples and their nearest $\ell_{2}$ neighbour in training set for different splits.

\subsection{Which Features Are the Most Different in Pixel Space?}

\begin{figure}[!h]
    \centering
    \includegraphics{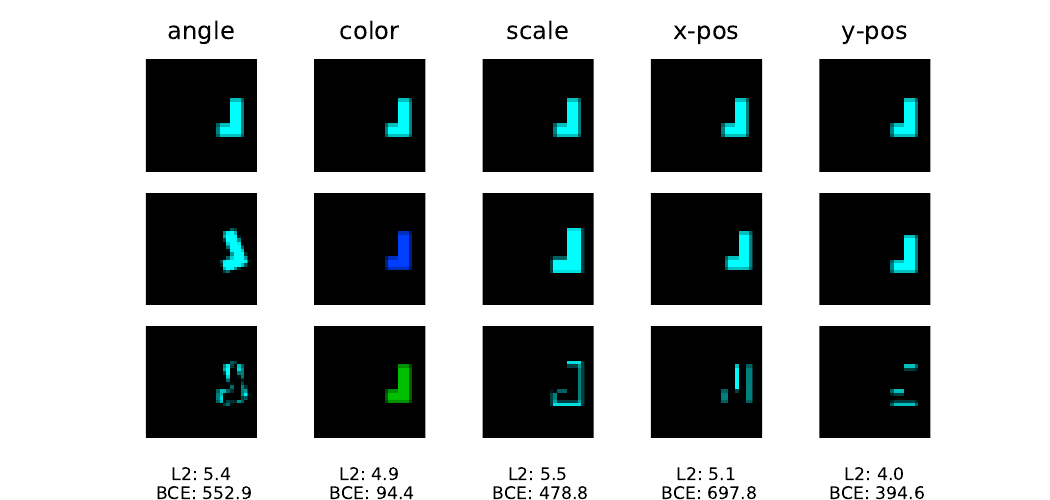}
    \caption[The deviation in pixel space caused by each feature in Tetrominoes dataset]{Effect of each feature in pixel space for Tetrominoes dataset. (\emph{Top}) Original image. (\emph{Middle}) A single feature in the original image modified by one unit. (\emph{Bottom}) $\ell_{2}$ distance between images in the top and middle.}
    \label{app:fig:tetris-features}
\end{figure}

One crucial factor in the difficulty of generalization in a dataset is the change caused in image space that is caused by moving in feature space. Not only this property can be different for different features, but it may also depend on the location in features space that change is happening (see Figure~\ref{app:fig:tetris-features}). In order to have a better understating of which features are more difficult to generalize to, we performed the following experiments. For all 163,840 tetromino images, we changed a single feature by a single unit. Figure~\ref{app:fig:tetris-features} shows the $\ell_{2}$ distance between the corresponding images.

\begin{figure}[!h]
    \centering
    \includegraphics[width=\textwidth]{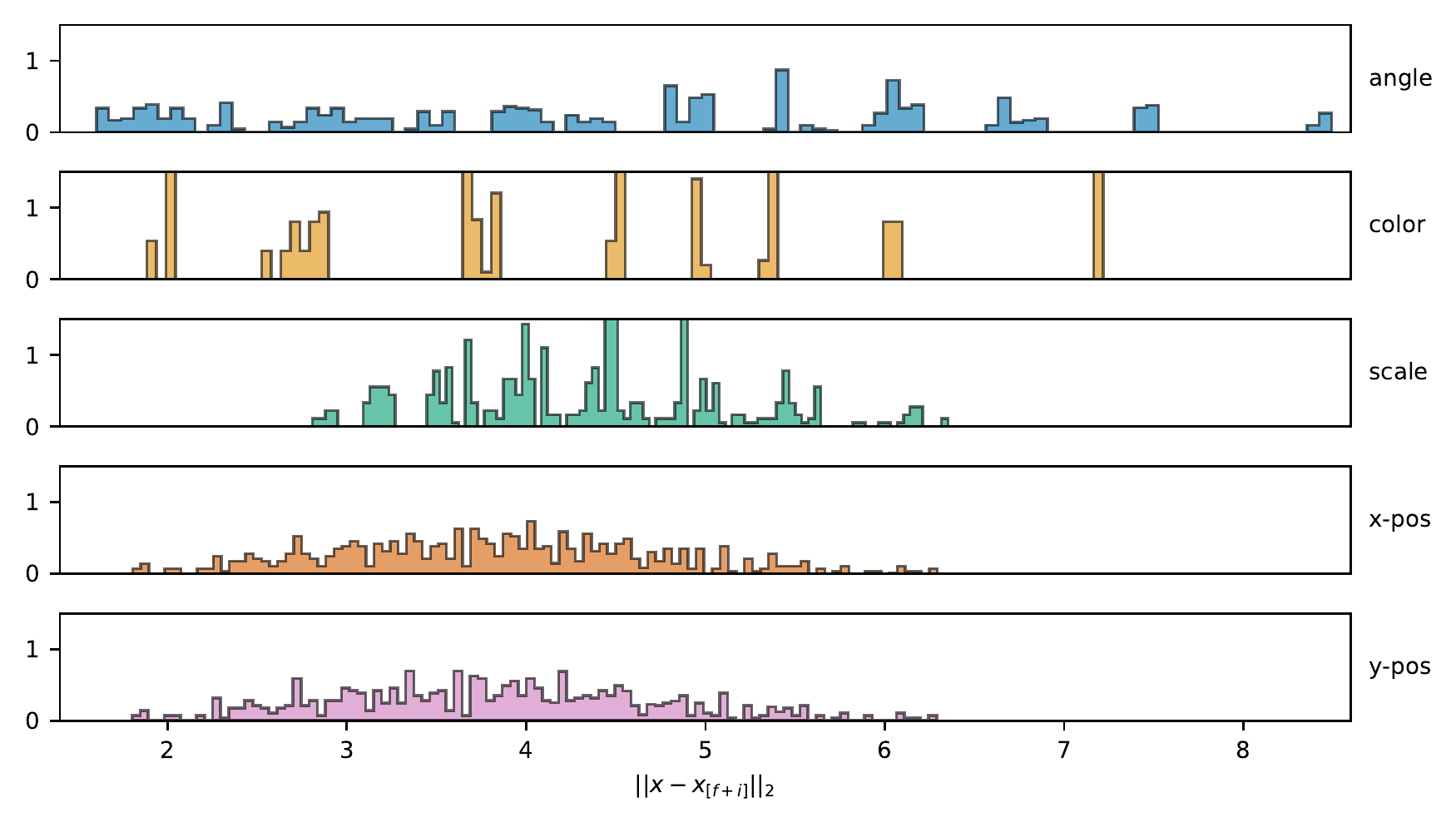}
    \vspace*{-4ex}
    \caption[Histograms of $\ell_2$ distance caused by changing each feature in Tetrominoes]{Histograms of $\ell_{2}$ distance between each tetromino image and the same tetromino modified in a single feature by 1 unit. We can observe that size causes the least difference in pixel space. x and y-positions seem to be the second and third most influential factors. Angle causes the most difference in pixel space.}
    \label{app:fig:tetris-features-l2diff}
\end{figure}

\newpage
\section{Additional Experimental Results}

\vspace*{-1ex}
\subsection{MNIST 9-Removal Experiment}
\label{app:sec:mnist}
\vspace*{-1ex}

As a means of gaining intuition on the VAE's ability to generalize to unseen example, we first carried out the following experiment. We trained a VAE with a 50 dimensional latent space on all MNIST digits with the exception of the 9s, and then tested out-of-domain generalization by attempting to reconstruct 9s. Figure~\ref{app:fig:9-removal} shows the decoder output, the weighted average $\mu$ from Equation~\ref{eq:shu-theorem}, and the 3 training examples with the largest weights. We compare a shallow (1 hidden layer, 400 neurons) and a deep (3 hidden layers, 400 neurons each) network.

As Proposition~\ref{shu-theorem} predicts, the deep VAE reconstructions closely resemble the nearest neighbors in the
training data in latent space. In most cases, a single sample dominates the weighted average. This is evident from the histogram of weight perplexities, which is strongly peaked at 1. This, combined with decoder outputs and neighbours with largest weights, suggests that VAEs can memorize training data even for simple encoder/decoder architectures with moderate capacity, and reconstructions are well-approximated by nearest neighbors in the training set when they do so.

\begin{figure}[!bt]
    \centering
    \includegraphics[width=0.5\textwidth]{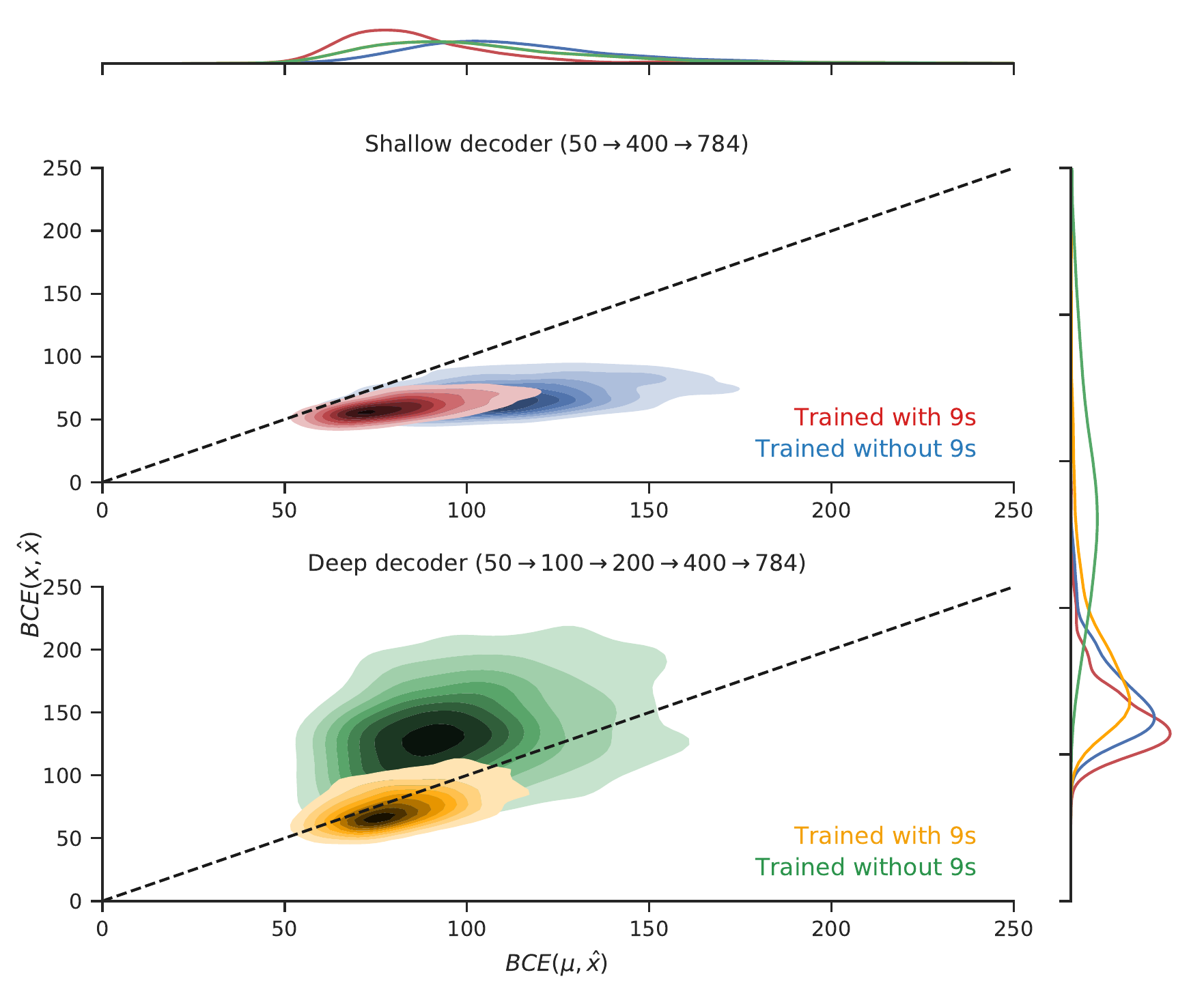}
    \vspace*{-1.5ex}
    \caption[Comparison of the BCE loss for the 9-removal experiment]{Distribution of the BCE loss between input image $\x$ and the decoder output $\hat{x}$, vs the loss between weighted average image $\mu$ and $\hat{x}$; calculated over test set consisting of images of 9s. The reconstructions are closer to the input than the weighted average for the 1-layer VAE (most of the mass of blue distribution lies below the line). On the other hand, the reconstructions are closer closer to the weighted average than the input for the 3-layer VAE (most of the mass of green distribution lies above the line).}
    \label{app:fig:bce-mnist}
\end{figure}

However, it is not the case that VAEs always memorize training data. A surprising finding is that shallow VAEs show comparatively good generalization to out-of-domain samples; reconstructions of 9s are passable, even though this digit class was not seen during training. The same trend is also visible when we compare the binary cross entropy (BCE) between reconstructions and samples from the withheld class, to the BCE between the reconstructions and the weighted averages (see Figure~\ref{app:fig:bce-mnist}). This suggests that the assumption of infinite capacity in Proposition 1 clearly matters, and that layer depth significantly affects the effective capacity of the network. 

\begin{figure}[!t]
    \centering
    \includegraphics[width=\textwidth]{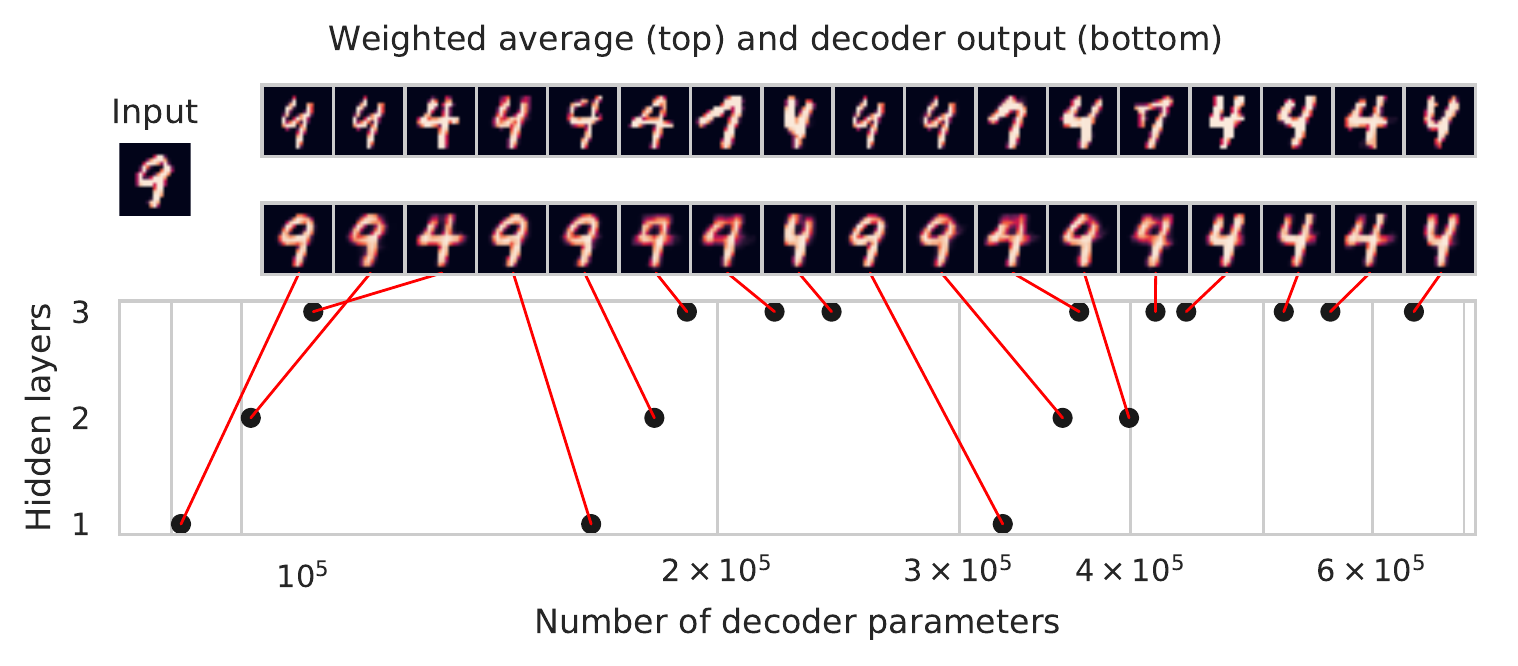}
    \vspace*{-5ex}
    \caption[Reconstructions and weighted averages $\mu$ for MNIST]{Decoder outputs and weighted average images for VAEs with different architectures.}
    \vspace*{-2ex}
    \label{app:fig:arch-capcaity-mnist}
\end{figure}

In the previous experiment, we provided a comparison between a shallow and a deep VAE. We now do a finer grained complexity analysis, and identify regions where Theorem~\ref{shu-theorem} holds. As with ``generalization'', ``capacity'' of a neural network is a hard aspect to characterize. Here, we will use number of parameters and layers as simple proxies for capacity. In Figure~\ref{app:fig:arch-capcaity-mnist}, we show reconstructed and weighted average images for 17 VAEs with different network architectures given an input sample from the withheld class. As VAEs get more complex, they overfit the training data therefore fail to reconstruct the unseen digit. Moreover, we observe that the reconstruction is closer to the weighted average for higher capacity networks. Another -intuitive- observation here the number of layers plays a more crucial role in complexity than number of parameters, since for VAEs with 3 hidden layers, reconstructions are more similar to weighted average, while in single hidden layer VAEs, the reconstructions match the input sample regardless of the number of parameters.

\vspace*{-1ex}
\subsection{Effect of Network Capacity When \texorpdfstring{$\beta$}{Beta}=1}
\vspace*{-1ex}

To probe the role of the model capacity in generalization, we compare 9 architectures that are trained using a standard VAE objective on the Default, CV\,(16k/147k), and Checkerboard splits. The CV(16k/147k) split is designed have similar typical pixel distance between nearest neighbors in the training and test set to the Checkerboard split (see Appendix~\ref{app:sec:tetris-details}), which enables a fair comparison by controlling for the difficulty of the generalization problem. We vary model capacity by using architectures with \{1, 2, 3\} layers that each have \{256, 512, 1024\} neurons.

\begin{figure}[!bht]
\centering
\includegraphics[width=\textwidth]{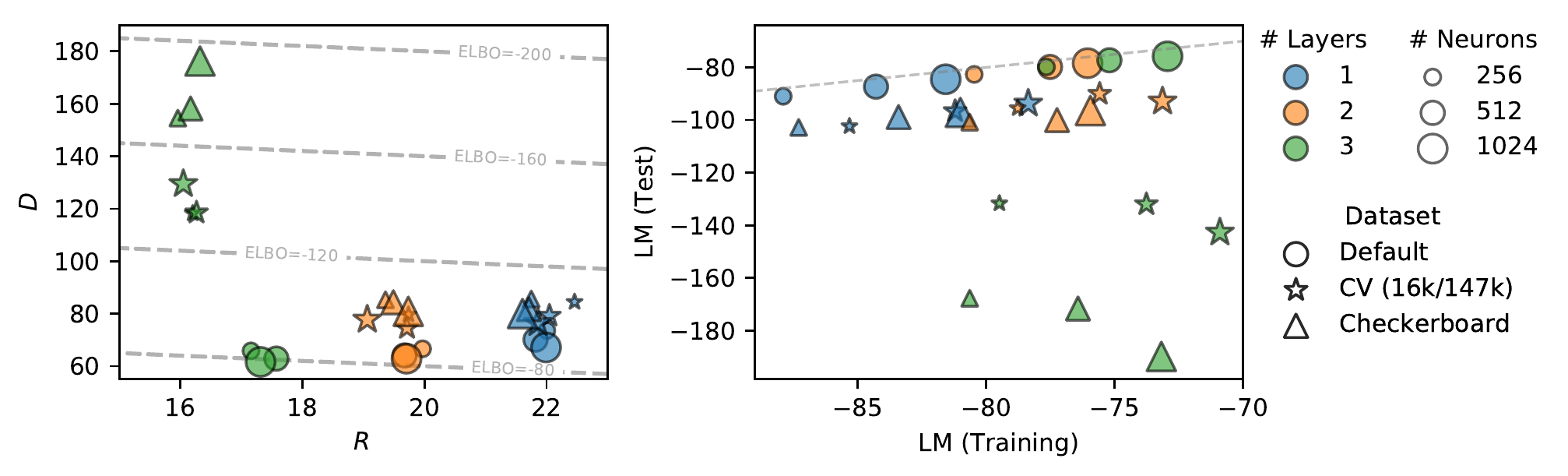}
\vspace*{-3ex}
\caption[Analysis of network capacity]{Analysis of network capacity: Each point is a standard VAE model with a different architecture, evaluated on a different dataset. \emph{(Left)} Test-set rate and distortion. The dashed lines denote contours of constant ELBO. 
\emph{(Right)} Training LM versus test LM. The dashed line denotes the contour where the LM for training and test sets are equal.}
\vspace*{-1ex}
\label{app:fig:vae-netcap-tetris}
\end{figure}

Figure~\ref{app:fig:vae-netcap-tetris} (\emph{left}) shows the test-set rate and distortion for all 27 models. Dashed lines indicate contours of equal $\text{ELBO} = D + R$. In Figure~\ref{app:fig:vae-netcap-tetris} (\emph{right}) we compare the LM for the training and test set. Here the dashed line marks the boundary where the test LM equals the training LM. For the Default split, increasing the model capacity uniformly improves generalization. Conversely, for the CV\,(16k/147k) and Checkerboard splits, we observe a strong deterioration in generalization performance in all 3-layer architectures. 

Figure~\ref{app:fig:vae-netcap-tetris} suggests that increasing model capacity can either improve or hurt generalization, depending on the difficulty of the generalization problem. In the Default split, which poses a comparatively easy generalization task, we observe that increasing model capacity improves generalization, whereas for more difficult tasks (as classically predicted in terms of a bias-variance trade-off), increasing model capacity deteriorates generalization.
We note here that the discrepancy between generalization performance on the CV and Checkerboard splits is relatively small, which suggests that these in-domain and out-of-domain tasks are indeed comparable in terms of their difficulty. 
\clearpage
\vspace*{-1ex}
\subsection{Memorization and Generalization When \texorpdfstring{$\beta$}{Beta}=1}

\begin{figure}[!h]
    \centering
    \includegraphics[width=\textwidth]{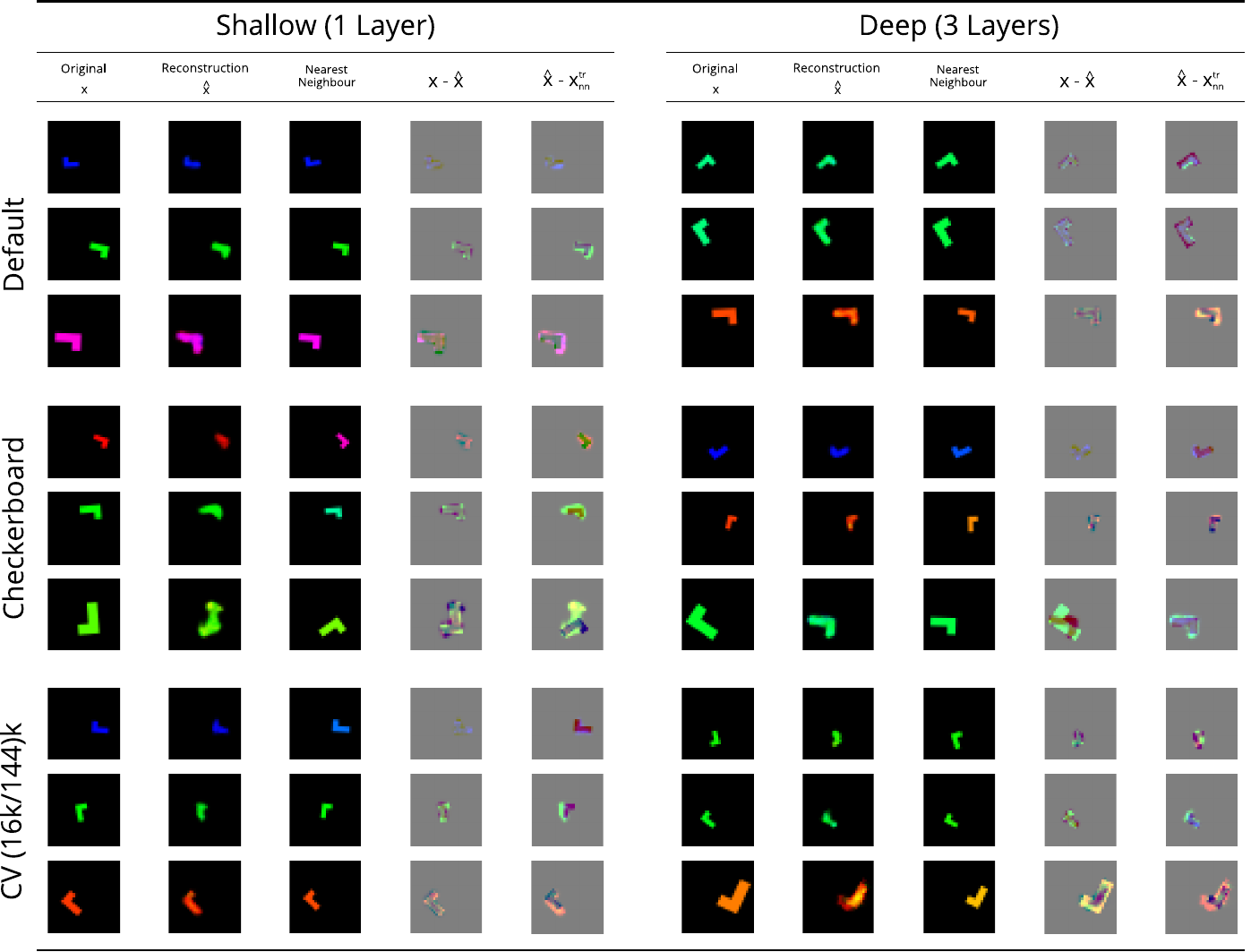}
    \vspace*{-3ex}
    \caption[Reconstructions and nearest neighbours in training set for Tetrominoes datasets]{Reconstruction of test samples from all datasets for 1-layer and 3-layer VAEs. Rows show examples with increasing reconstruction loss, randomly selected from the $10^{th}$ (\emph{top}), $40^{th}$ to $60^{th}$ (\emph{middle}), and $90^{th}$ (\emph{bottom}) percentiles.}
    \vspace*{-1ex}
    \label{app:fig:vae-recons-tetris}
\end{figure}

Regardless of our metric for generalization performance, there is evidence that VAEs can both underfit and overfit the training data. Several researchers~\citep{bousquet2017optimal,rezende2018taming,alemi2018fixing, shu2018amortized} have pointed out that an infinite-capacity optimal decoder will memorize the training data. Concretely, the following proposition holds for an optimal decoder:

\begin{thm}[\citet{shu2018amortized}]
Assume a likelihood $p(\x \,|\, \z)$ in an exponential family with mean parameters $\mu$ and sufficient statistics $T(\cdot)$, a fixed encoder $q(\z \,|\, \x)$, and training data $\{\x_1, \ldots, \x_{\ntrain{}}\}$. In the limit of infinite capacity, the optimal decoder $\mu(\z)$ is
\begin{equation}
\mu(\z)=
\sum_{n=1}^{\ntrain{}}
w_n(\z) \:
T(\x_n), \quad w_n(\z)
=
\frac{\q(\z\mid\x_n)}
     {\sum_m \q(\z\mid\x_m)}.
\label{eq:shu-theorem}
\end{equation}
\label{shu-theorem}
\end{thm}
We qualitatively evaluate the extent to which VAEs memorize the training data (as predicted by Proposition~\ref{shu-theorem}). Figure~\ref{app:fig:vae-recons-tetris} compares the reconstructions of test examples in the Default, CV\,(16k/147k), and Checkerboard splits, by 1-layer and 3-layer architectures. For each architecture, we show 3 examples from the test set along with reconstructions and nearest training neighbours (with respect to $w_n$) for both models. The 3 examples are representatives of easy ($<$10\textsuperscript{th} percentile), typical (45\textsuperscript{th}-55\textsuperscript{th} percentile), or difficult ($>$90\textsuperscript{th} percentile) samples in terms of pixel-wise nearest-neighbor distance to the training data. In the case of the CV\,(16k/147k) and Checkerboard splits, we see that the 1-layer VAE can reconstruct unseen examples even when the nearest neighbour in the latent space is quite different, while reconstructions for the 3-layer VAE are consistent with the memorization behavior described by Proposition~\ref{shu-theorem}. In the Default split, we observe that reconstructions are similar for 1-layer and 3-layer architectures, and are often well-approximated by their nearest-neighbors in the training data.

\begin{figure}[!h]
    \centering
    \includegraphics[width=0.32\linewidth]{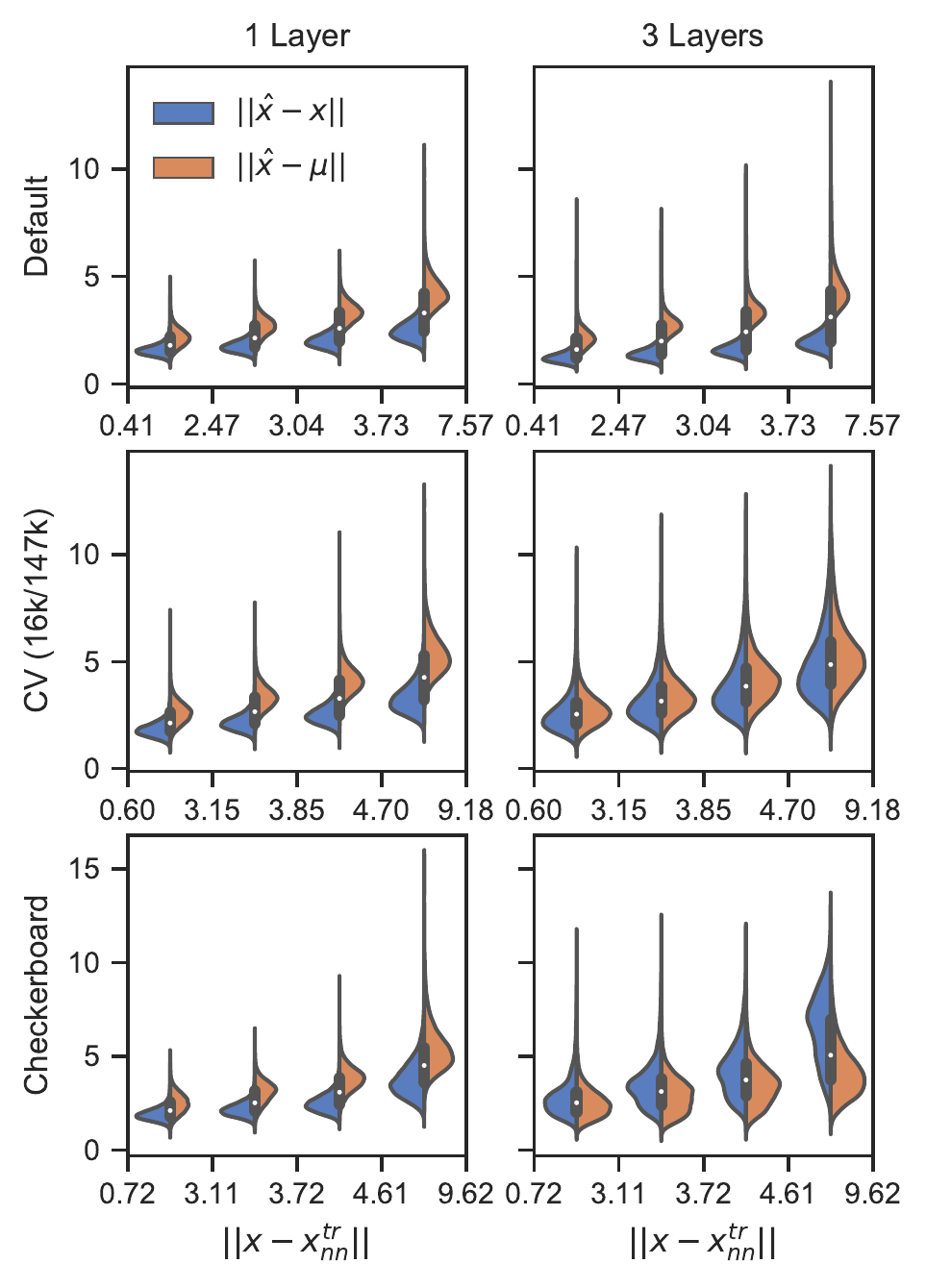}
    \includegraphics[width=0.32\linewidth]{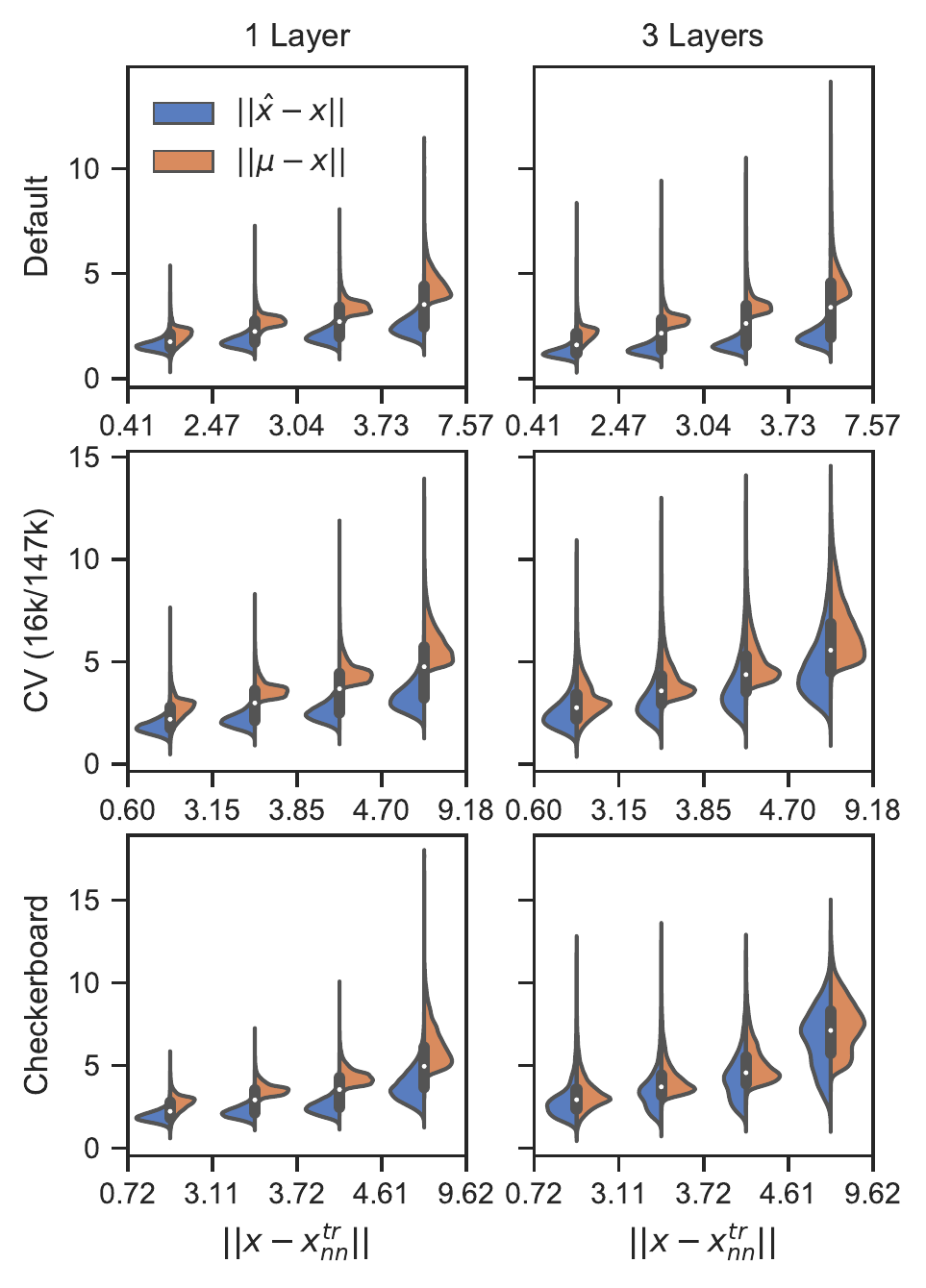}
    \includegraphics[width=0.32\linewidth]{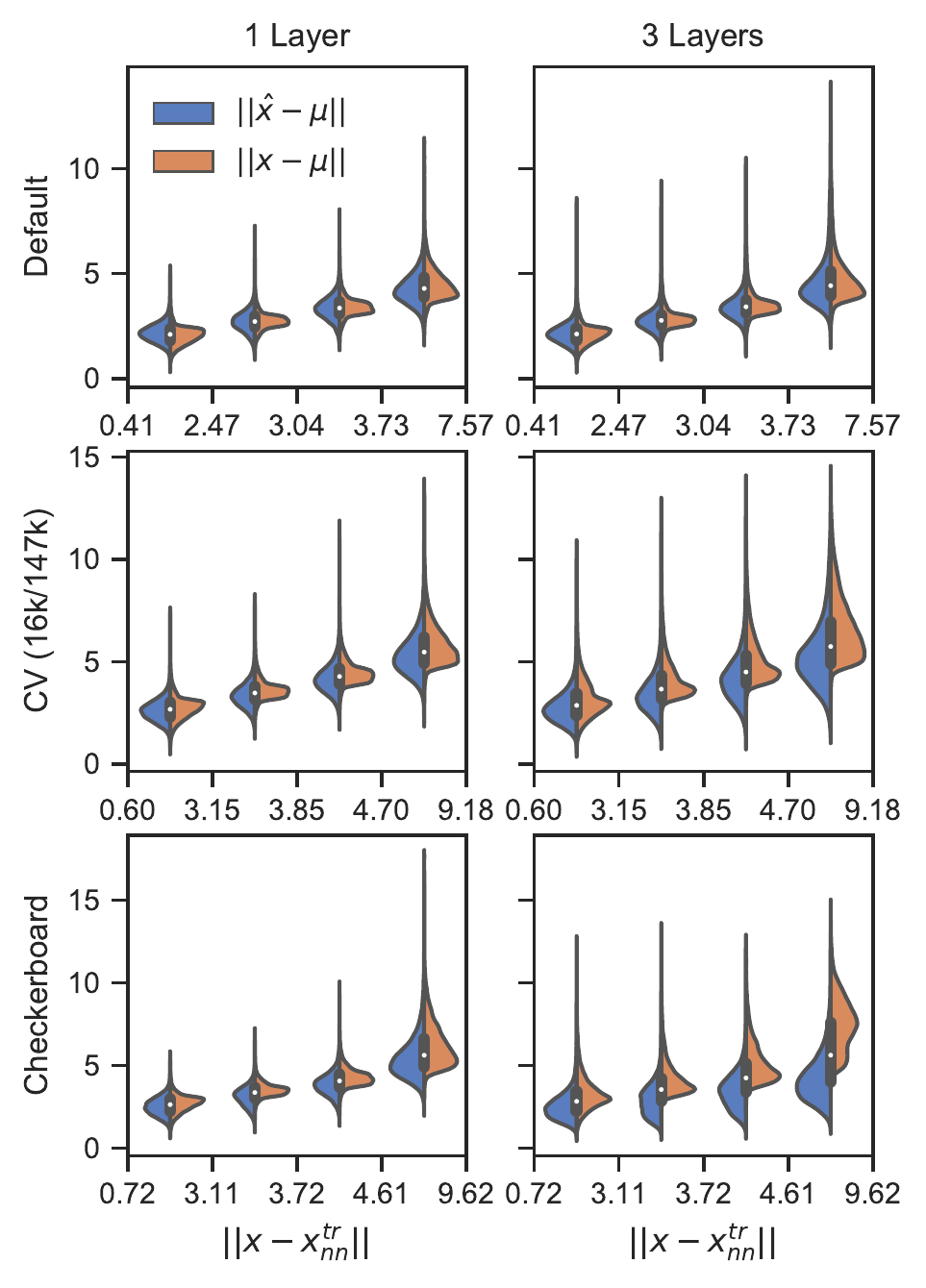}
    \caption[Additional violin plots for analyzing the the reconstructions]{Distributions of distance between test data and output of the decoder ($\|\hat{x}-x\|$), distance between test data and output of the infinite capacity decoder ($\|\mu-x\|$, $\mu$ from Equation~\eqref{eq:shu-theorem}), and distance between test data and and output of the infinite capacity decoder ($\|\hat{x}-\mu\|$), partitioned according distance to nearest training sample ($\|x-x_{nn}^{tr}\|$) into 4 bins of equal sizes. Values in x-axis are the limits of the bins.}
    \vspace*{-4ex}
    \label{app:fig:violin-plots}
\end{figure}

To provide a more quantitative evaluation, we report distances between $\x$, $\v{\mu}$ and $\hat{\x}$. Figure~\ref{app:fig:violin-plots} shows violin plots of the pixel-wise distances between test images and their reconstructions ($\|\hat{\x}-\x\|$) and the infinite-capacity decoder outputs ($\|\hat{\x}-\v{\mu}\|$) as defined in Equation~\eqref{eq:shu-theorem}. We split the test set into 4 bins of equal sizes according to the distance of examples to their nearest training example ($\|\x-\x^{tr}_{nn}\|$) and show histogram pairs for each bin. Values in x-axis indicate the limits of each bin.

For splits that are not trivial, we see different behavior across different bins. Looking at the rightmost histogram-pairs (the most difficult 25\% of test samples) in each panel, we once again observe qualitatively different behaviors for 1-layer and 3-layer networks. For the 1-layer networks $\|\hat{\x}-\x\|$ is smaller than $\|\hat{\x}-\v{\mu}\|$, which shows that reconstructions cannot be explained by memorization alone. This result holds across all 3 train-test splits. For 3-layer networks, we see that $\|\hat{\x}-\x\|$ slightly decreases relative to the 1-layer model in the Default split, once again indicating that overparameterization aids generalization (here in terms of reconstruction loss) in this regime. Conversely for the CV and Checkerboard splits, we see that $\|\hat{\x}-\x\|$ increases relative to the 1-layer model. Moreover, in the Checkerboard split, we observe that $\|\hat{\x}-\v{\mu}\|$ is smaller than $\|\hat{\x}-\x\|$, which shows that reconstructions are closer to memorized data than to the actual test examples.


\begingroup
\begin{figure*}[!tb]
    \vspace{-0.5em}
    \centering
    \resizebox{\linewidth}{!}{%
    \renewcommand{\arraystretch}{0.1}
        \begin{tabular}{ccccccccccccccccc}
          \toprule\\
          \multirow{2}{*}[-0.5\dimexpr \aboverulesep + \belowrulesep + \cmidrulewidth]{\textsf{\large Original}} & \multicolumn{8}{c}{\textsf{\large \textbf{1 Layer}}} & \multicolumn{8}{c}{\textsf{\large \textbf{3 Layers}}}\\
          \cmidrule(lr){2-9} \cmidrule(lr){10-17}
          & \textsf{\large Recons.} & \begin{tabular}{c}\textsf{\large Weighted}\\\textsf{\large Average} \end{tabular} 
          & \multicolumn{3}{c}{\textsf{\large Neighbours with Largest Weight}} 
          & \multicolumn{3}{c}{\textsf{\large Weight Perplexity}}
         & \textsf{\large Recons.} & \begin{tabular}{c} \textsf{\large Weighted}\\\textsf{\large Average} \end{tabular} 
         & \multicolumn{3}{c}{\textsf{\large Neighbours with Largest Weight}} & \multicolumn{3}{c}{\textsf{\large Weight Perplexity}}\\\\  
         \adjustbox{valign=c}{\includegraphics[height=0.75in]{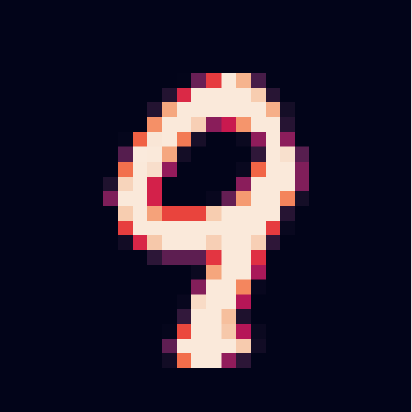}} &
         \adjustbox{valign=c}{\includegraphics[height=0.75in]{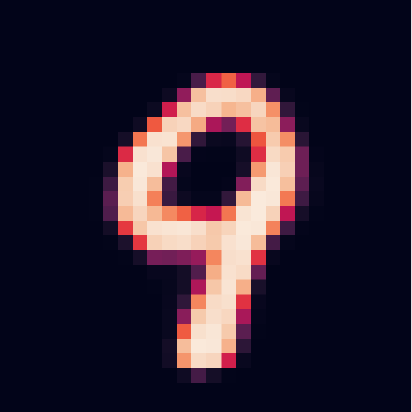}} &
         \adjustbox{valign=c}{\includegraphics[height=0.75in]{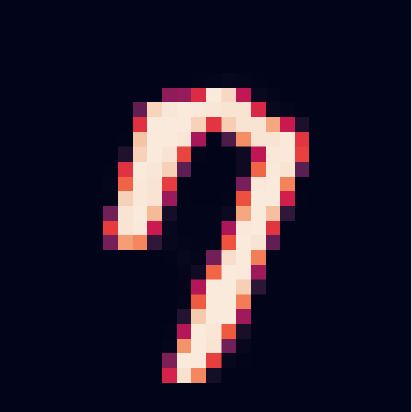}} &
         \begin{tabular}{c}
              w:0.97 \\
              \adjustbox{valign=c}{\includegraphics[height=0.75in]{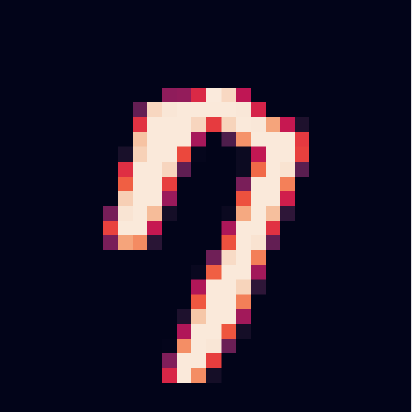}} 
         \end{tabular} &
         \begin{tabular}{c}
              w:0.02 \\
              \adjustbox{valign=c}{\includegraphics[height=0.75in]{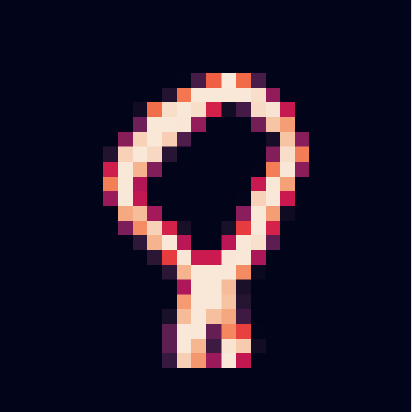}} 
         \end{tabular} & 
         \begin{tabular}{c}
              w:0.00 \\
              \adjustbox{valign=c}{\includegraphics[height=0.75in]{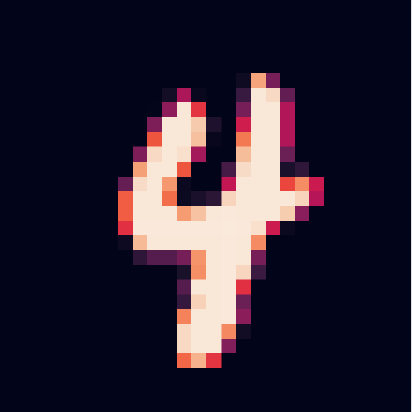}} 
         \end{tabular} &
         \multicolumn{3}{c}{\multirow[c]{12}{*}{\includegraphics[height=1.7in]{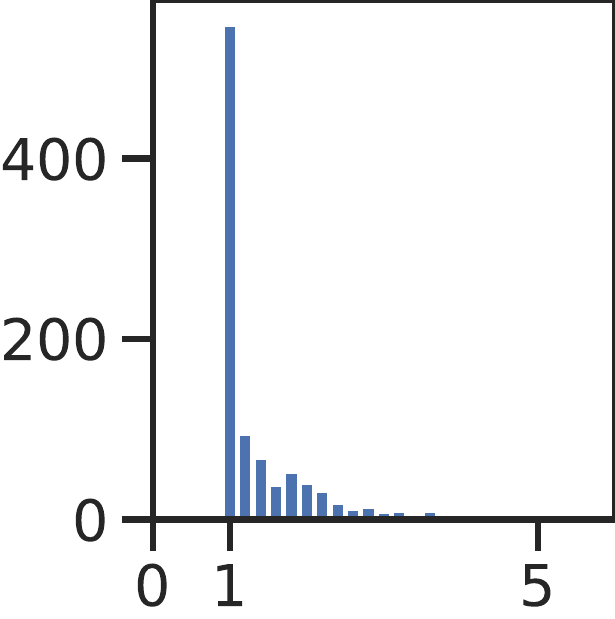}
         }} &
         \adjustbox{valign=c}{\includegraphics[height=0.75in]{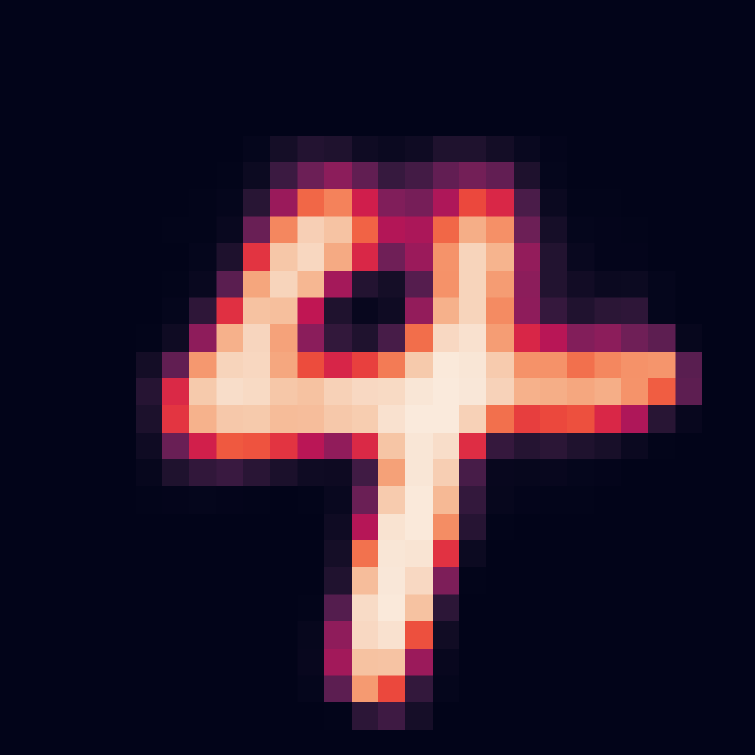}} &
         \adjustbox{valign=c}{\includegraphics[height=0.75in]{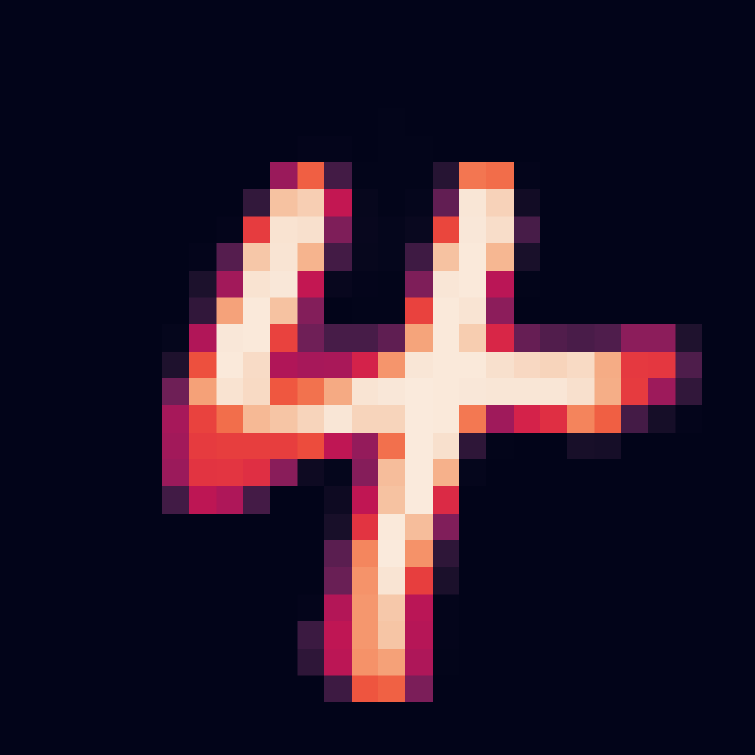}} &
         \begin{tabular}{c}
              w:0.46 \\
              \adjustbox{valign=c}{\includegraphics[height=0.75in]{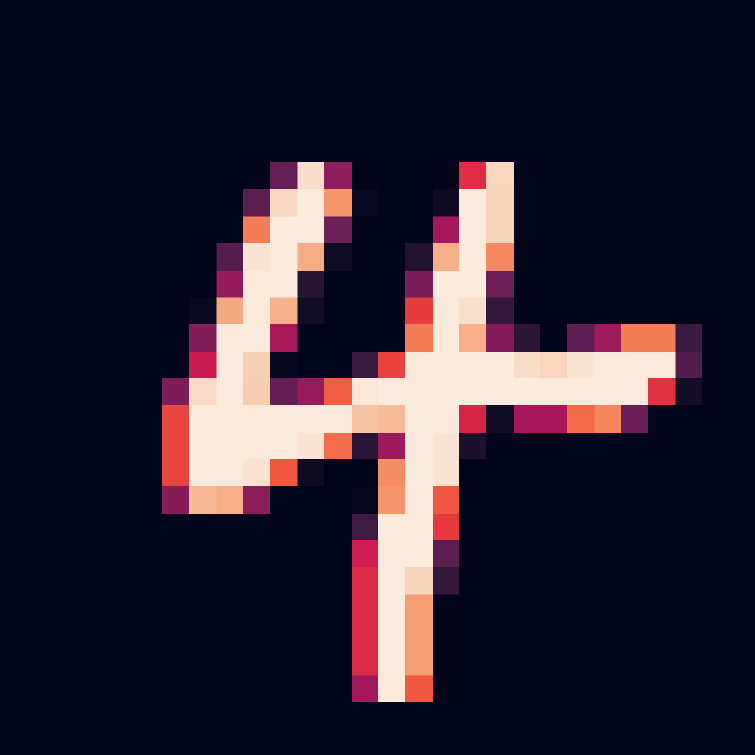}} 
         \end{tabular} &
         \begin{tabular}{c}
              w:0.37 \\
              \adjustbox{valign=c}{\includegraphics[height=0.75in]{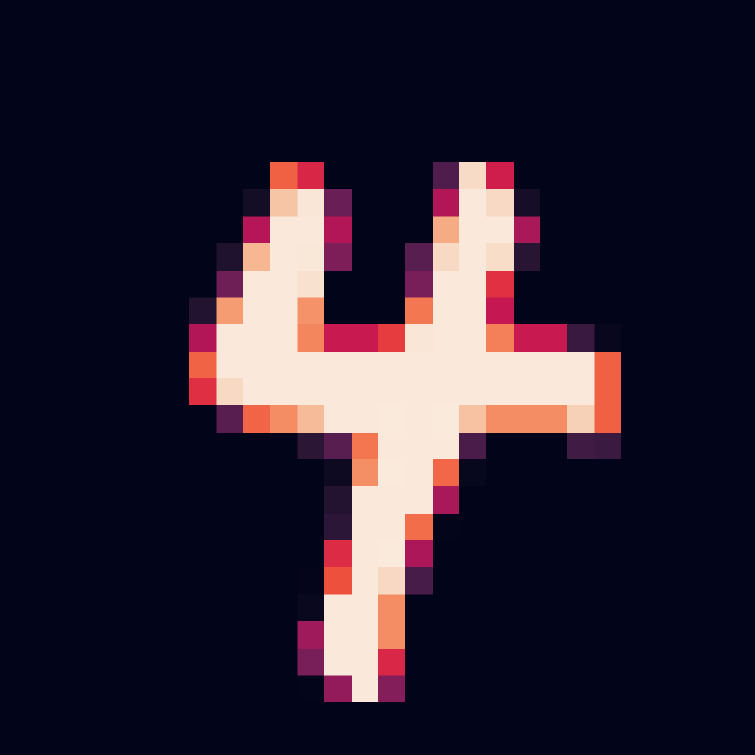}} 
         \end{tabular} & 
         \begin{tabular}{c}
              w:0.12 \\
              \adjustbox{valign=c}{\includegraphics[height=0.75in]{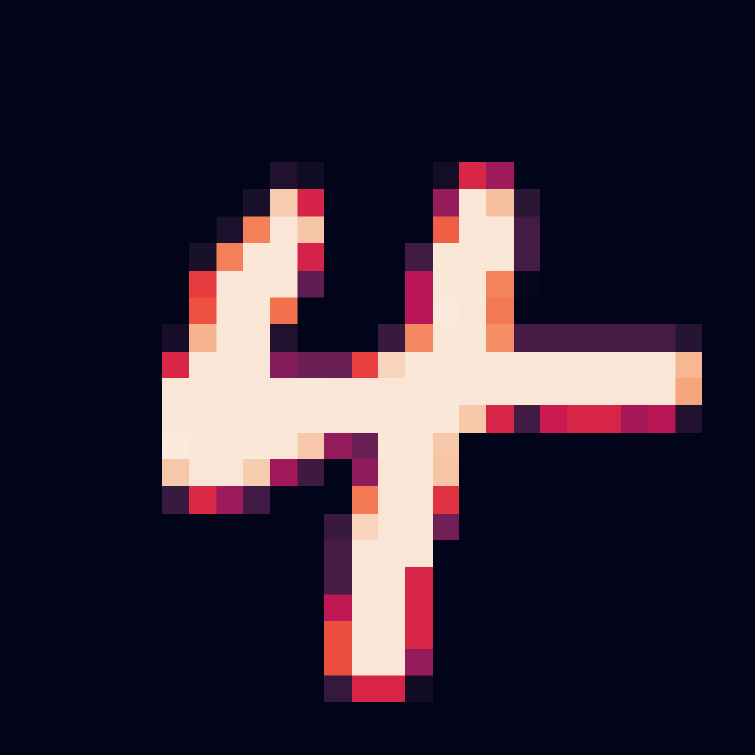}} 
         \end{tabular} & 
         \multicolumn{3}{c}{\multirow[c]{12}{*}{\adjustbox{valign=c}{\includegraphics[height=1.7in]{{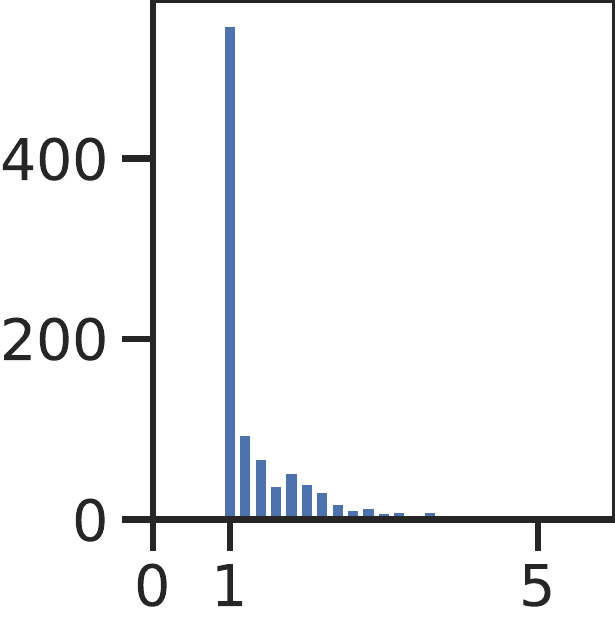}}}}} \\[0.2in]
         
         \adjustbox{valign=c}{\includegraphics[height=0.75in]{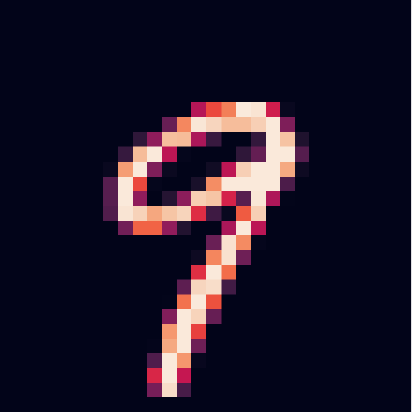}} &
         \adjustbox{valign=c}{\includegraphics[height=0.75in]{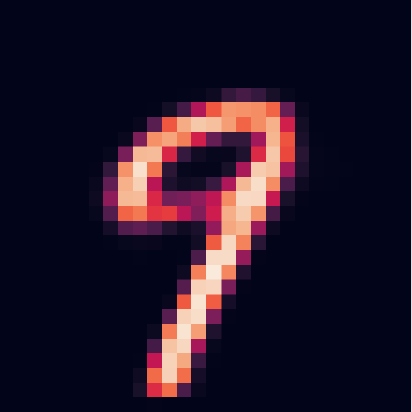}} &
         \adjustbox{valign=c}{\includegraphics[height=0.75in]{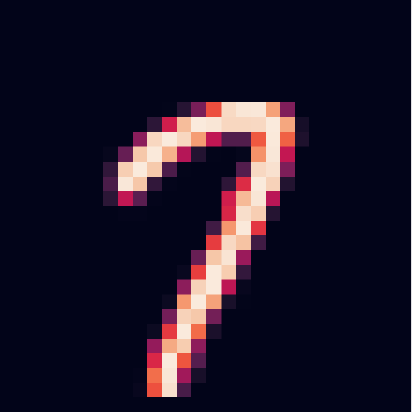}} &
         \begin{tabular}{c}
              w:0.60 \\
              \adjustbox{valign=c}{\includegraphics[height=0.75in]{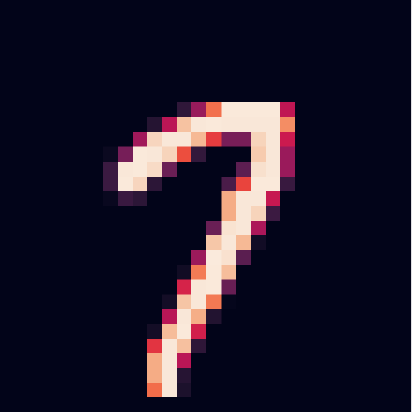}} 
         \end{tabular} &
         \begin{tabular}{c}
              w:0.36 \\
              \adjustbox{valign=c}{\includegraphics[height=0.75in]{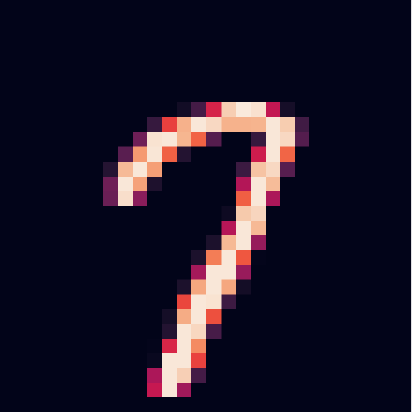}} 
         \end{tabular} & 
         \begin{tabular}{c}
              w:0.03 \\
              \adjustbox{valign=c}{\includegraphics[height=0.75in]{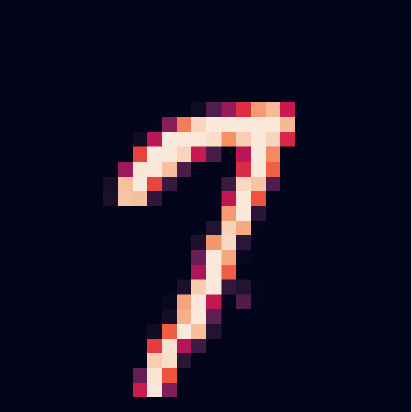}} 
         \end{tabular} &
         \multicolumn{3}{c}{} &
         \adjustbox{valign=c}{\includegraphics[height=0.75in]{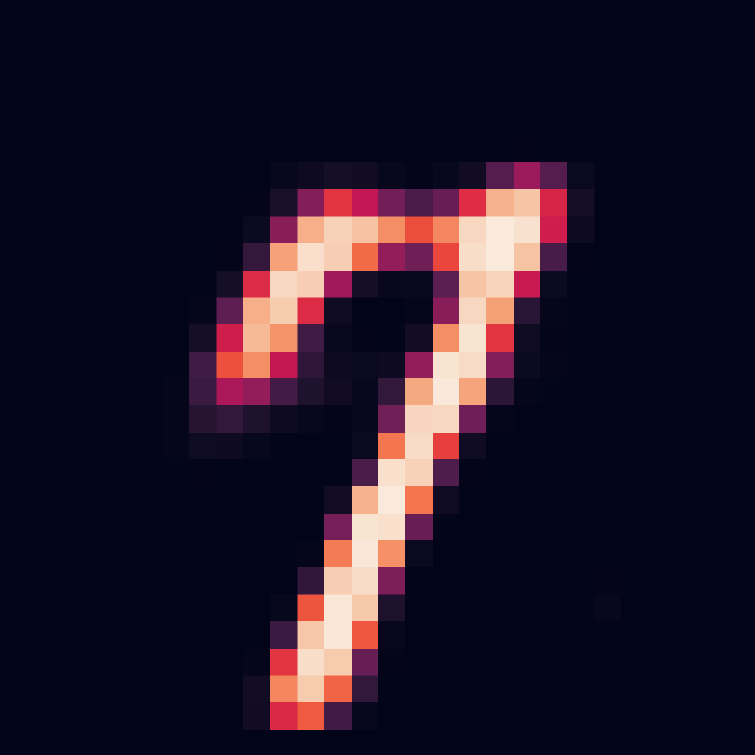}} &
         \adjustbox{valign=c}{\includegraphics[height=0.75in]{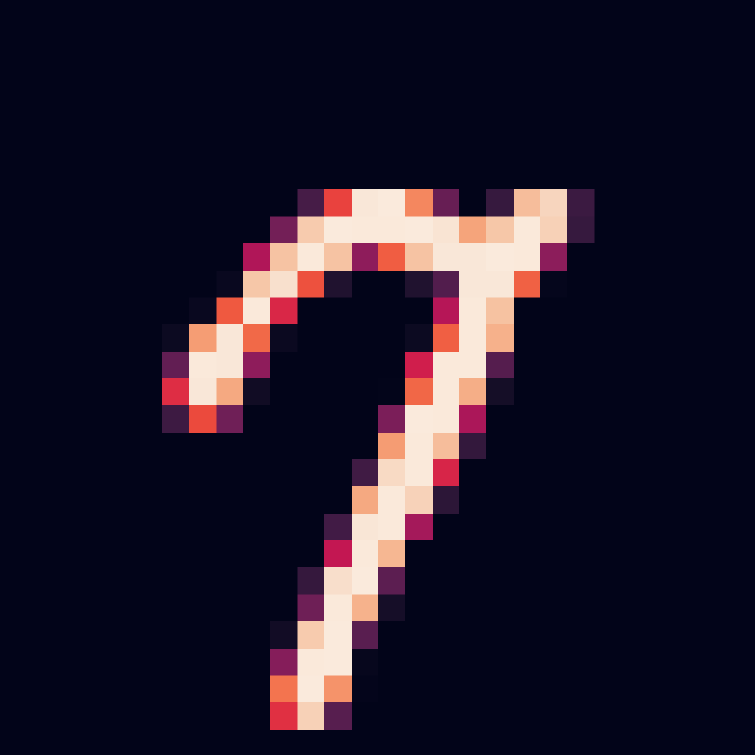}} &
         \begin{tabular}{c}
              w:1.00 \\
              \adjustbox{valign=c}{\includegraphics[height=0.75in]{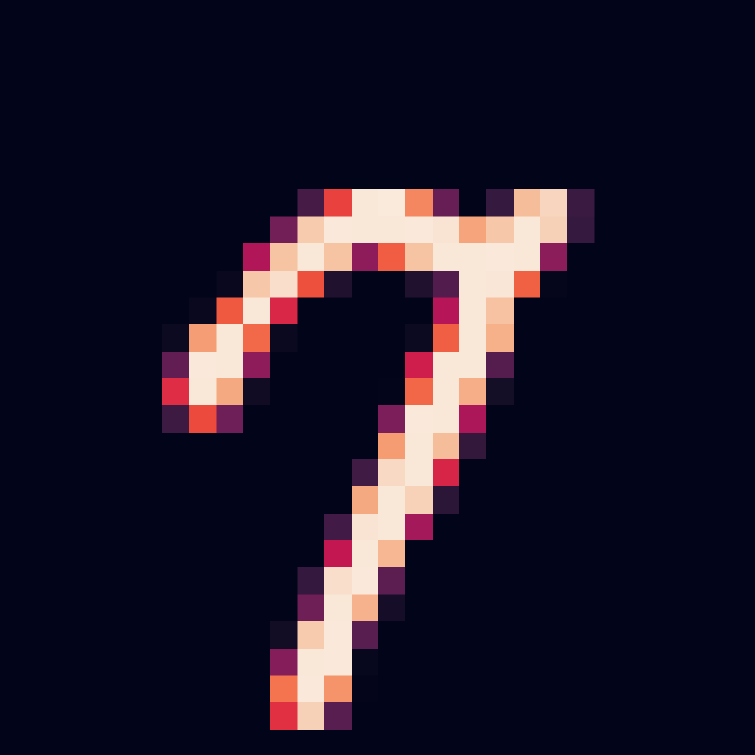}} 
         \end{tabular} &
         \begin{tabular}{c}
              w:0.00 \\
              \adjustbox{valign=c}{\includegraphics[height=0.75in]{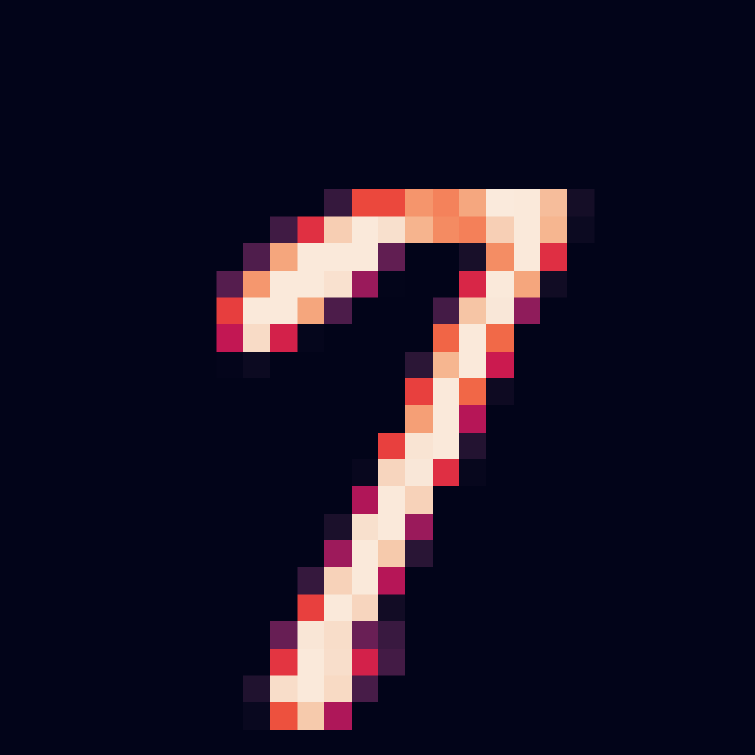}} 
         \end{tabular} & 
         \begin{tabular}{c}
              w:0.12 \\
              \adjustbox{valign=c}{\includegraphics[height=0.75in]{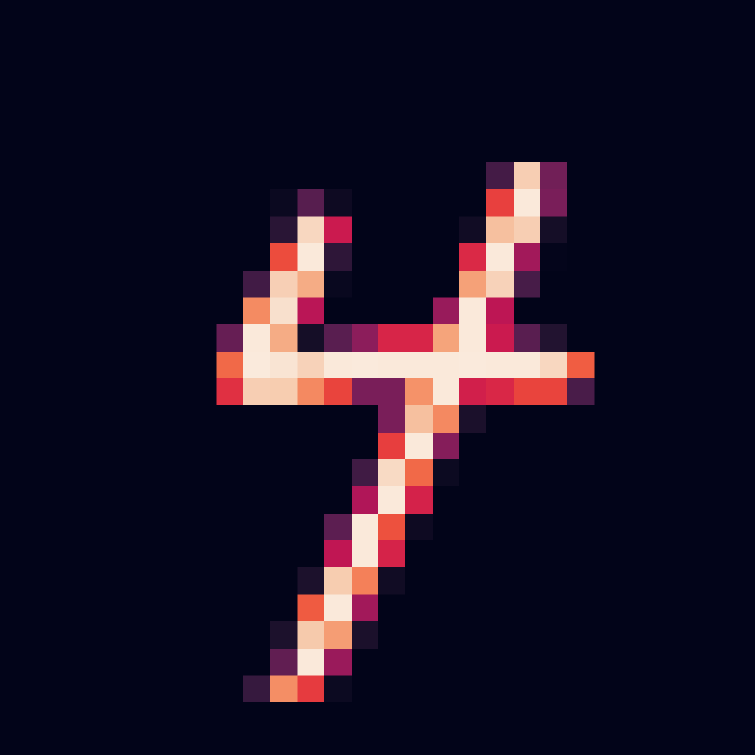}} 
         \end{tabular} & 
         \multicolumn{3}{c}{} \\[0.2in]
         
         \adjustbox{valign=c}{\includegraphics[height=0.75in]{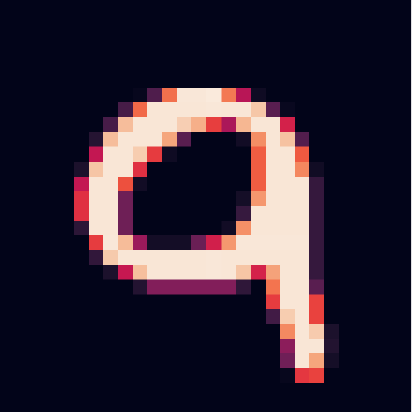}} &
         \adjustbox{valign=c}{\includegraphics[height=0.75in]{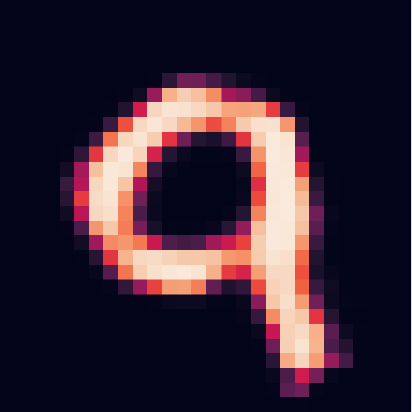}} &
         \adjustbox{valign=c}{\includegraphics[height=0.75in]{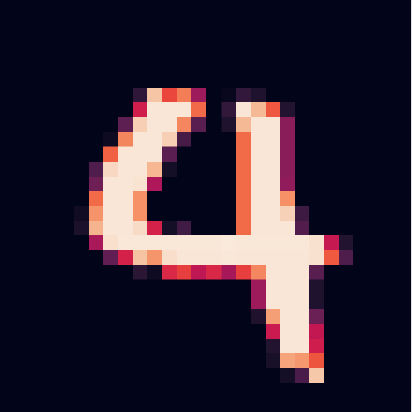}} &
         \begin{tabular}{c}
              w:1.00 \\
              \adjustbox{valign=c}{\includegraphics[height=0.75in]{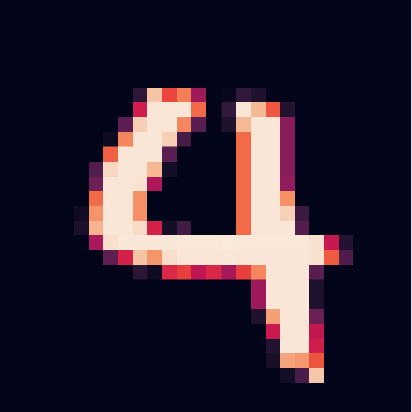}} 
         \end{tabular} &
         \begin{tabular}{c}
              w:0.00 \\
              \adjustbox{valign=c}{\includegraphics[height=0.75in]{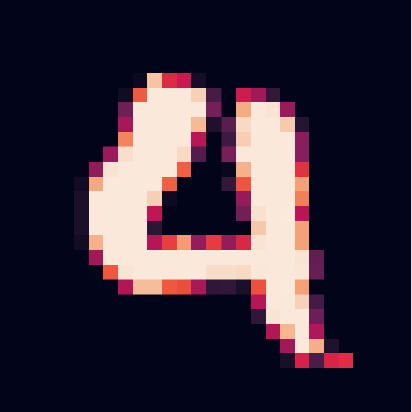}} 
         \end{tabular} & 
         \begin{tabular}{c}
              w:0.00 \\
              \adjustbox{valign=c}{\includegraphics[height=0.75in]{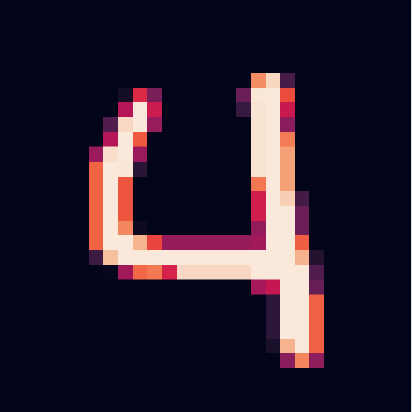}} 
         \end{tabular} &
         \multicolumn{3}{c}{} &
         \adjustbox{valign=c}{\includegraphics[height=0.75in]{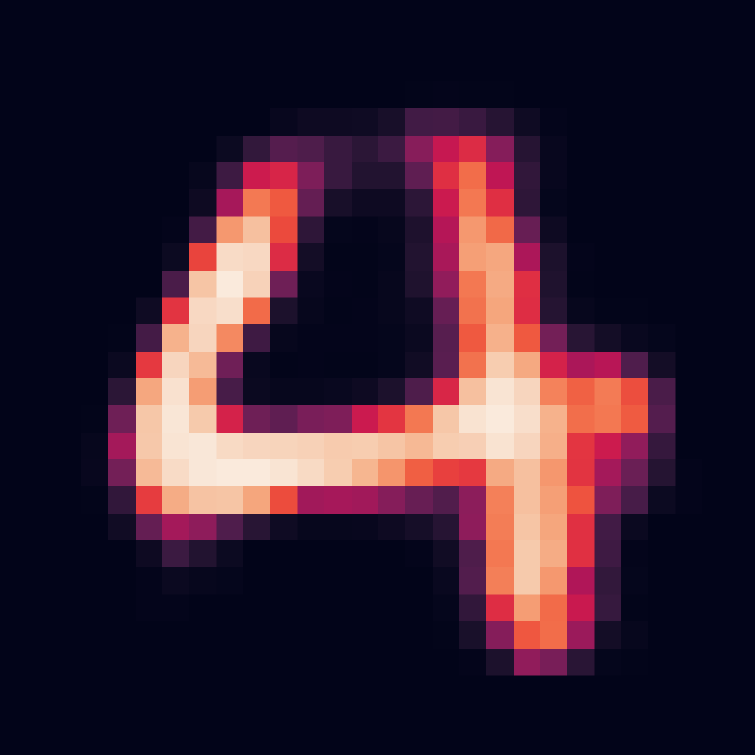}} &
         \adjustbox{valign=c}{\includegraphics[height=0.75in]{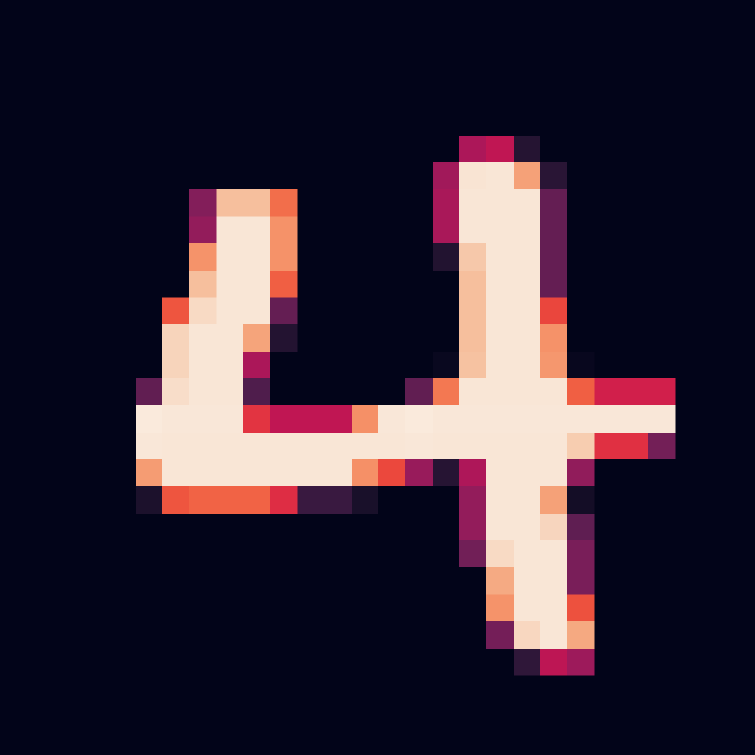}} &
         \begin{tabular}{c}
              w:1.00 \\
              \adjustbox{valign=c}{\includegraphics[height=0.75in]{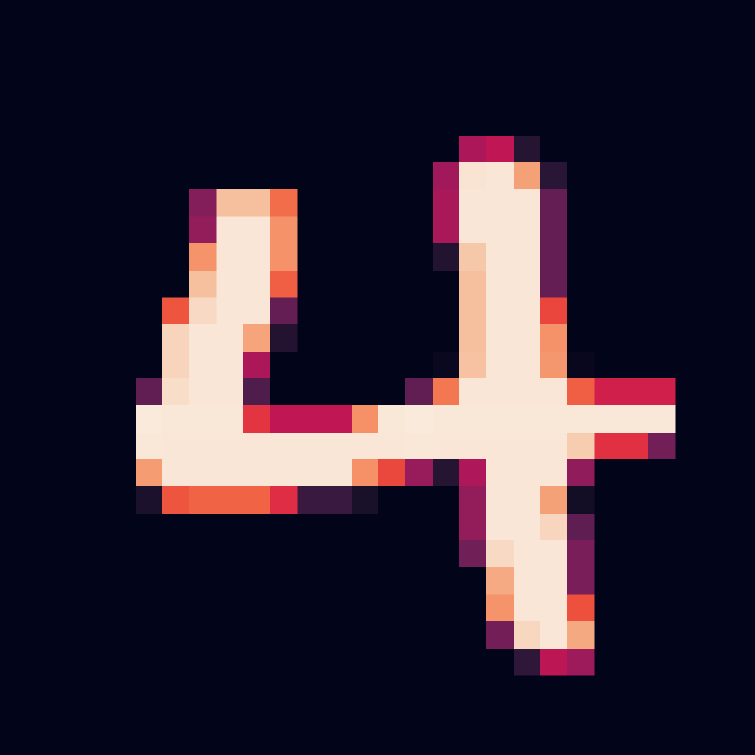}} 
         \end{tabular} &
         \begin{tabular}{c}
              w:0.00 \\
              \adjustbox{valign=c}{\includegraphics[height=0.75in]{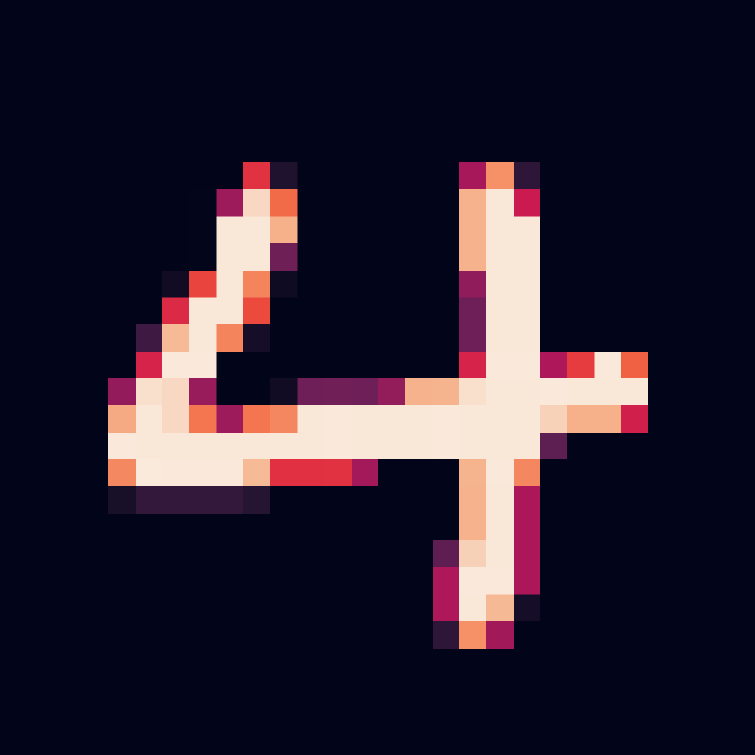}} 
         \end{tabular} & 
         \begin{tabular}{c}
              w:0.00 \\
              \adjustbox{valign=c}{\includegraphics[height=0.75in]{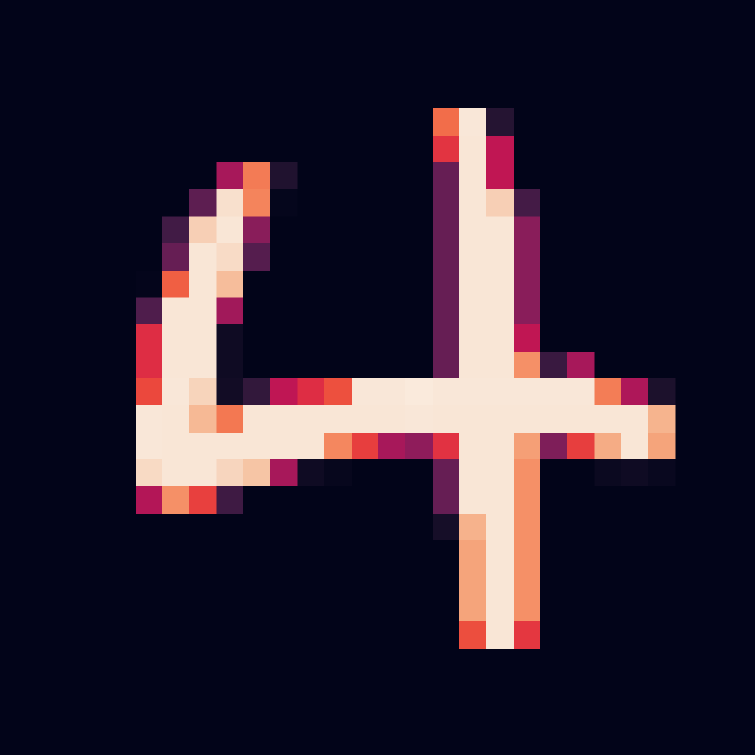}} 
         \end{tabular} &
         \multicolumn{3}{c}{} \\[0.3in]
         \bottomrule
        \end{tabular}
    }
    \caption[MNIST 9-removal experiment]{
    MNIST 9-removal experiment: Reconstruction of out-of-domain samples by 1-layer and 3-layer VAEs, alongside weighted average, and neighbours in training data with largest weights. Reconstructions in a 3-layer VAE closely match the weighted average, which in turn are often just the nearest neighbour in training data). 1-layer VAE does not exhibit this behaviour and can reconstruct samples cannot be represented by the convex combination of its training data.}
    \label{app:fig:9-removal}
    \vspace*{-2ex}
\end{figure*}
\endgroup


\begin{figure}[!h]
    \centering
    \includegraphics[width=\textwidth]{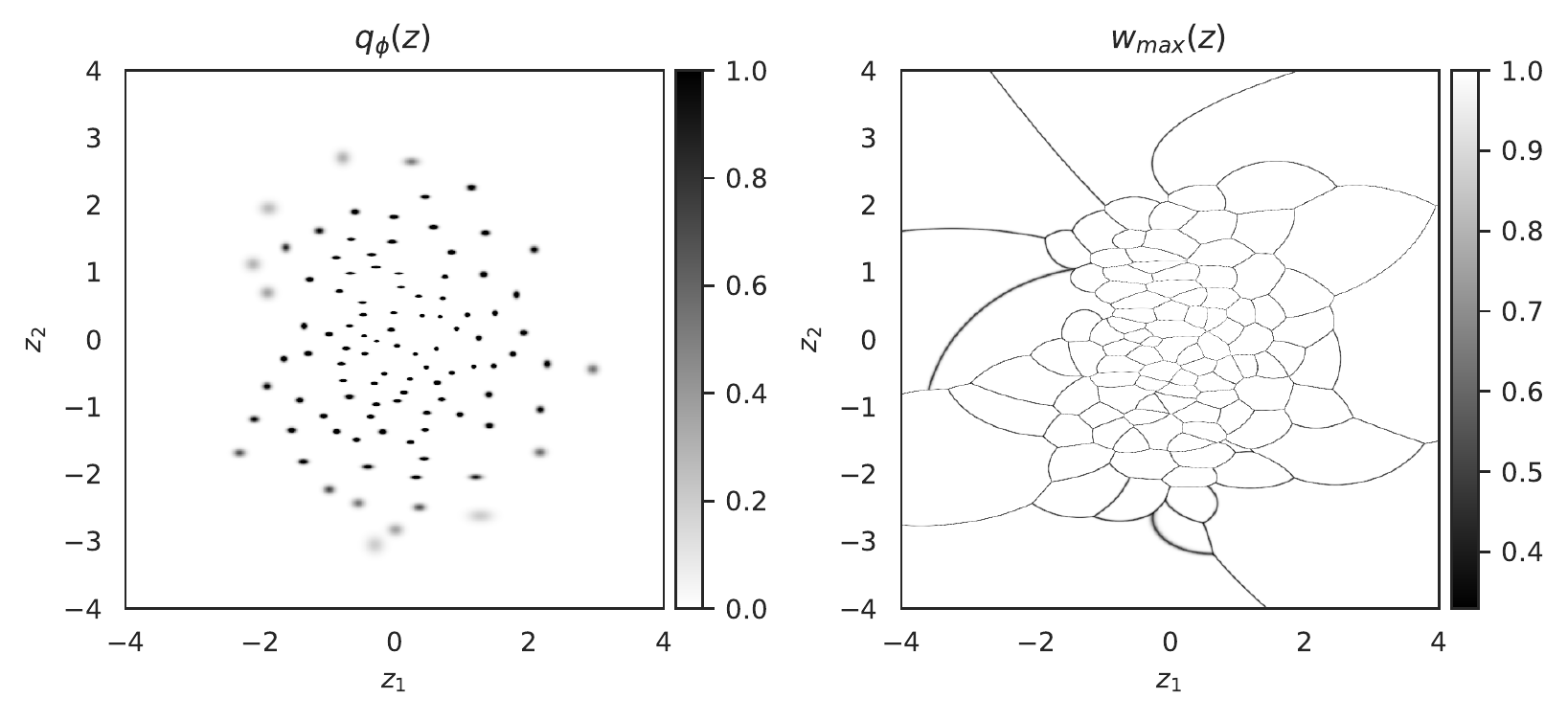}
    \vspace*{-3.5ex}
    \caption[Visualization of latent space in mini-MNIST]{Data memorization in a VAE with latent dimension trained on 100 MNIST examples. (\emph{Left}) Inference marginal $\q(\z) = \frac{1}{N} \sum_{n}\q(\z \,|\, \x_n)$.
    (\emph{Right}) Partitioning of the latent space. To close approximation, the decoder reconstructs a memorized nearest neighbor from the training data (for approximately $96\%$ of the shown region, the largest weight is $w_{max}>0.99$)}.
    \vspace*{-2.5ex}
    \label{app:fig:voronoi}
\end{figure}

\newpage

\vspace*{-1ex}
\subsection{ELBO and Log Marginal Likelihood}
\vspace*{-1ex}
\label{app:sec:elbo-lm}
As discussed in Section~\ref{sec:vae}, there are different ways of viewing the VAE objective. For the purpose understanding the effect of the rate, so far we focused on the distortion as the metric and studying the role of the rate as just a regularizer. Here, we study the effect of $\beta$ on more common metrics for evaluating generalization in VAEs, namely the ELBO and log marginal likelihood $\log \p(\x)$. 

\begin{figure}[!ht]
\centering
\includegraphics[width=\linewidth]{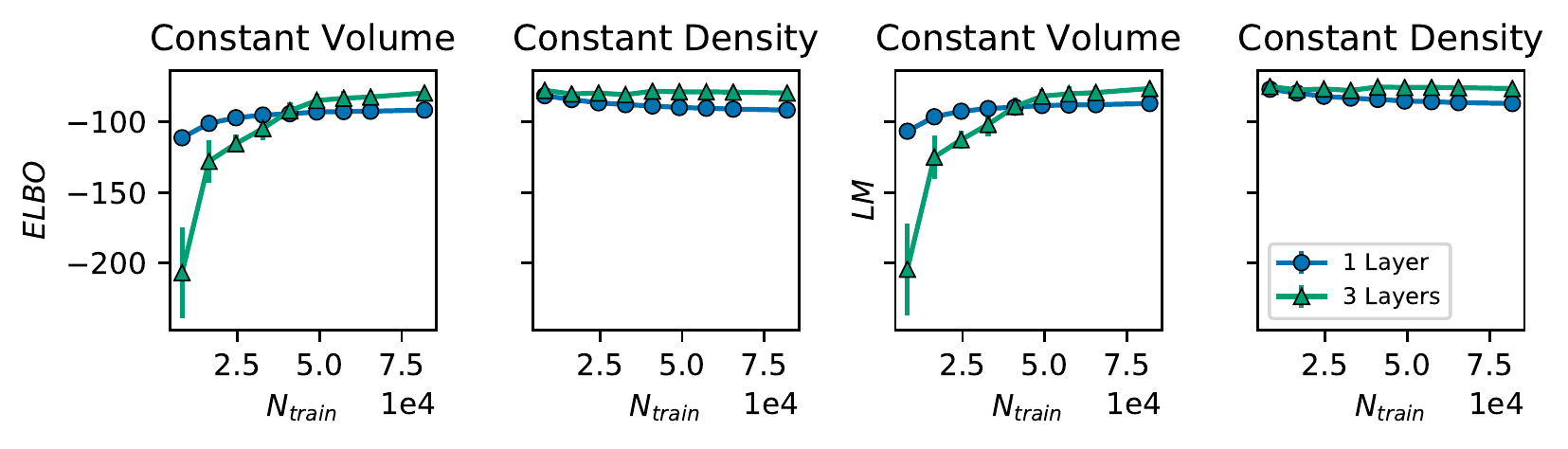}
\vspace{-4ex}
\caption[$N_\text{train}$ vs ELBO and LM for CD and CV splits with 1-layer and 3-layer VAEs]{Training set size versus test ELBO and LM, for CD and CV splits of Tetrominoes dataset with 1-layer and 3-layer VAEs ($\beta$=1), averaged over 5 independent restarts.}
\vspace{-3ex}
\label{app:fig:datasize-lm-elbo}
\end{figure}

We first look at the standard-VAE objective ($\beta=1$) in order to separate the influence of KL and the difficulty of generalization problem as well as depth. Figure~\ref{app:fig:datasize-lm-elbo} shows the ELBO and $\log \p(\x)$ as a function of the training set size at $\beta=1$. Once again, we observe two qualitatively different forms of behavior. In the CD splits, 3-layer networks almost uniformly outperform 1-layer networks. In the case of CV splits, there is a cross-over. The 1-layer model performs better for smaller (sparser) datasets, but is overtaken by the 3-layer model for larger (denser) datasets. These results suggest that overparameterizing can in fact be beneficial if the training and test sets are similar. Furthermore, it indicates it is not the size of training data but the similarly of training and test examples that is the key factor.

\vspace*{-1ex}
\subsubsection{The \texorpdfstring{$\beta \neq 1$}{Beta Not Equal to One} Case}
\vspace*{-1ex}

\begin{figure}[!ht]
\centering
\includegraphics[width=\linewidth]{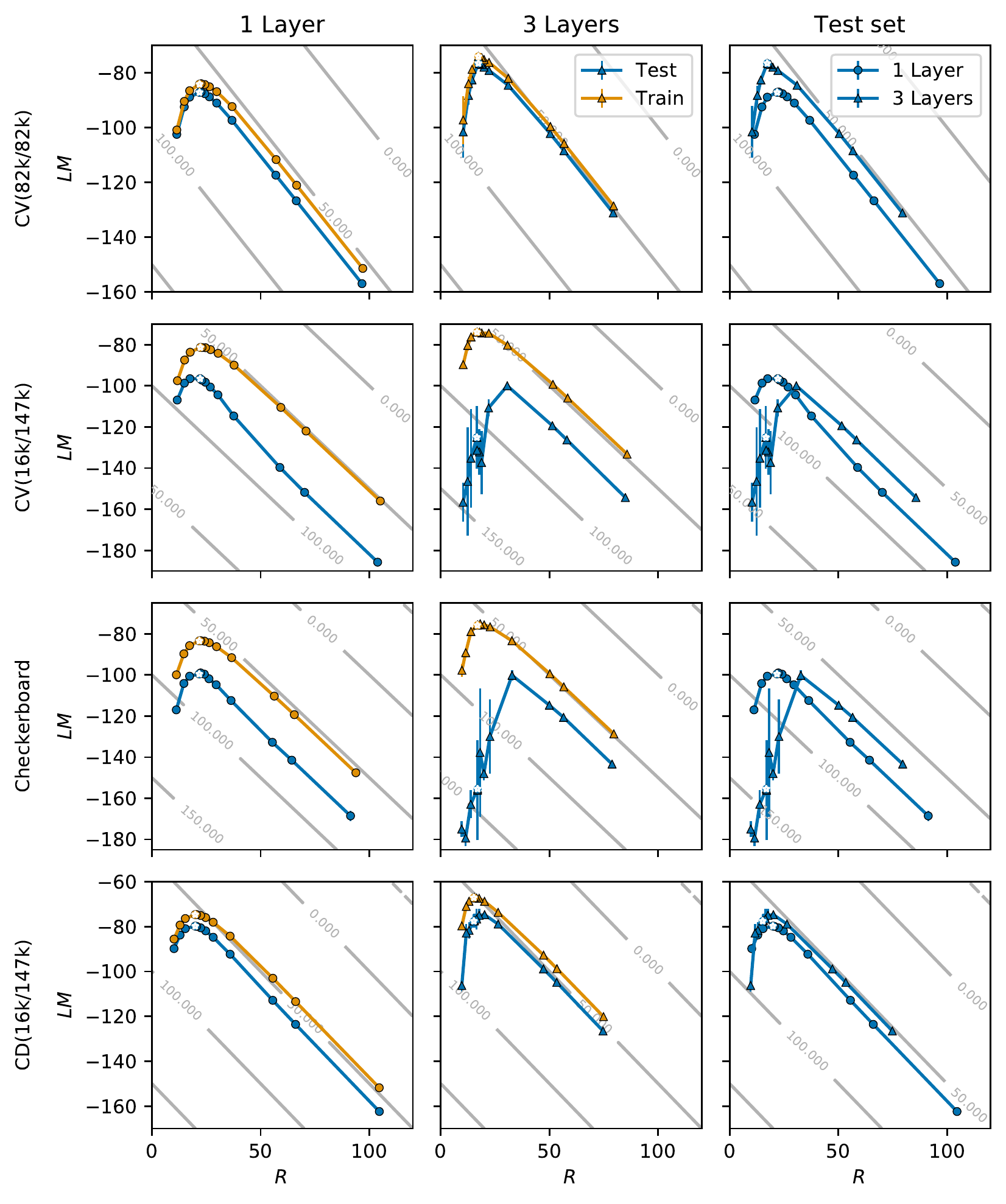}
\vspace{-4ex}
\caption[Log marginal likelihood (LM) vs rate evaluated for various splits]{Log marginal likelihood and rate evaluated on training and test sets for CV(82k/82k) (top row), CV(16k/147k) (2nd row), Checkerboard (3rd row), and CD(16k/147k) (bottom row) splits, trained with 1-layer and 3-layer models. Each dot constitutes a $\beta$ value (white stars indicate the $\beta$=1 point), averaged over 5 independent restarts.}
\vspace{-3ex}
\label{app:fig:lm-vs-kl}
\end{figure}

\begin{figure}[!ht]
\centering
\includegraphics[width=\linewidth]{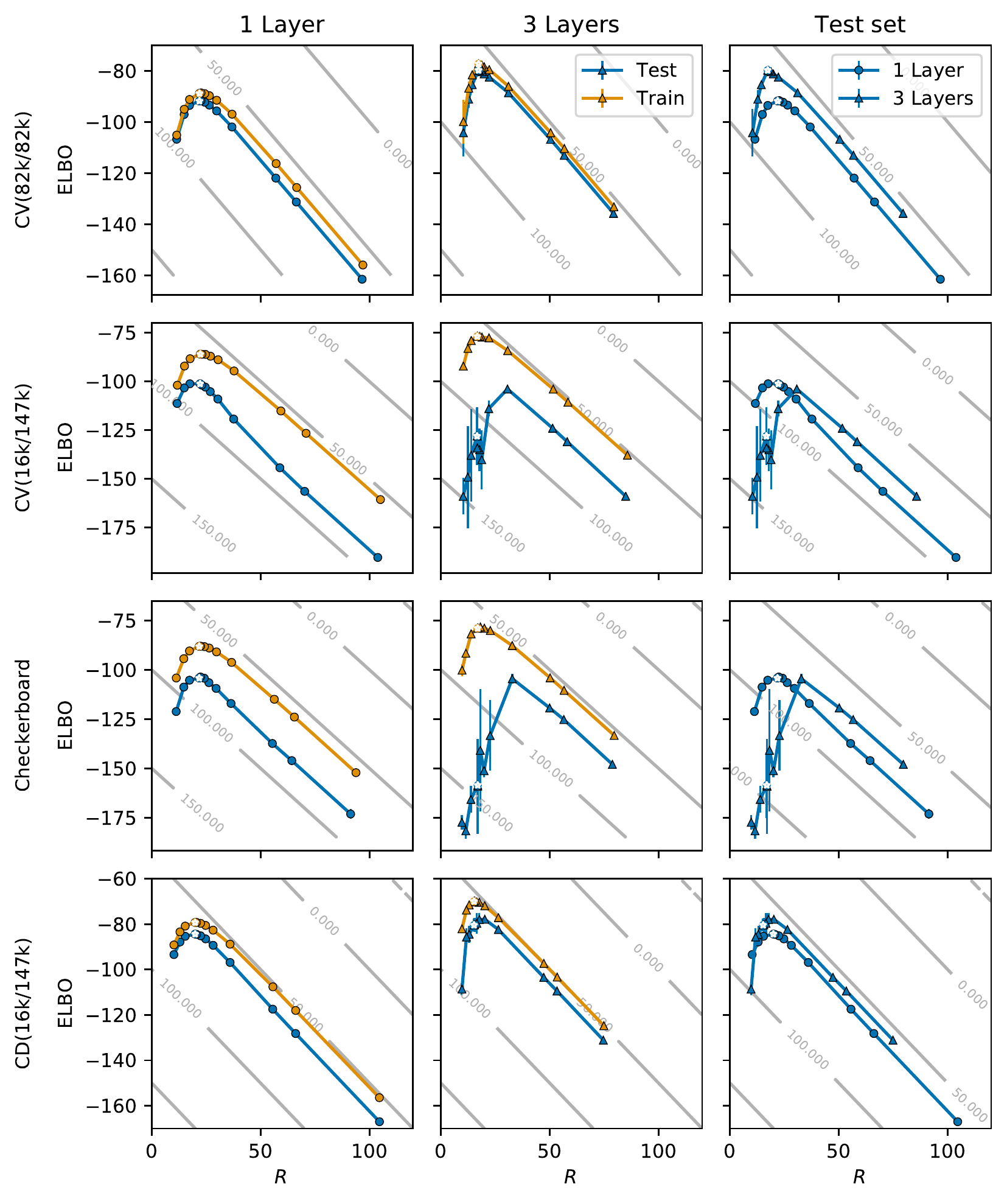}
\vspace{-4ex}
\caption[ELBO vs rate evaluated for various splits]{ELBO and rate evaluated on training and test sets for CV(82k/82k) (top row), CV(16k/147k) (2nd row), Checkerboard (3rd row), and CD(16k/147k) (bottom row) splits, trained with 1-layer and 3-layer models. Each dot constitutes a $\beta$ value (white stars indicate the $\beta$=1 point), averaged over 5 independent restarts.}
\vspace{-3ex}
\label{app:fig:elbo-vs-kl}
\vspace*{3ex}
\end{figure}

It is not considered standard to compute ELBO or Log marginal likelihood (LM) for $\beta$ values other than 1 for two main reasons. First, the $\beta$-VAE objective~\ref{eq:beta-vae} has mainly been trained for the purpose of disentanglement rather than learning a generative model. Second, for values lower than $\beta$ is no longer a lower bound. However, we observed in Figure~\ref{fig:rd-recons-curves} that it is possible to achieve a \emph{higher} test ELBO when we set $\beta < 1$. Therefore, we decided to investigate the effect of KL regularization on the test ELBO and LM by plotting the rate against LM (Figure~\ref{app:fig:lm-vs-kl}) and ELBO (Figure~\ref{app:fig:elbo-vs-kl}).

Looking at the first and last rows in Figures~\ref{app:fig:lm-vs-kl} and~\ref{app:fig:elbo-vs-kl}, we observe that $\beta=1$ (or nearby values) typically yield the highest LM and ELBO. These results are unsurprising given that $\beta=1$ results in the objective being the exact ELBO. We also observe that the 3-layer models are able to achieve a higher LM and ELBO than the 1-layer models. This confirms our intuition that when the distance between training and test examples are small, we can benefit by using models with higher capacity. By either increasing or decreasing $\beta$, we see a drop in LM and ELBO as the training objective becomes different than the metric. The results for the CV (16k/147k) and Checkerboard splits however are very different. For the 1-layer case, we still see that $\beta$ values near 1 yield the highest ELBO and LM. In the 3-layers case however, see that decreasing $\beta$ can significantly \emph{improves} generalization\footnote{In the Checkerboard split, the test set is out-of-domain, therefore one can argue that lower LM is considered to be a better result.}. In fact we see that for $\beta=0.1$, the 3-layer VAEs achieve nearly the same LM and ELBO as the 1-layer case. 

\clearpage
\subsection{Norm of the Weights for Decoder and Encoder}

\begin{figure}[!h]
\centering
\includegraphics[width=\linewidth]{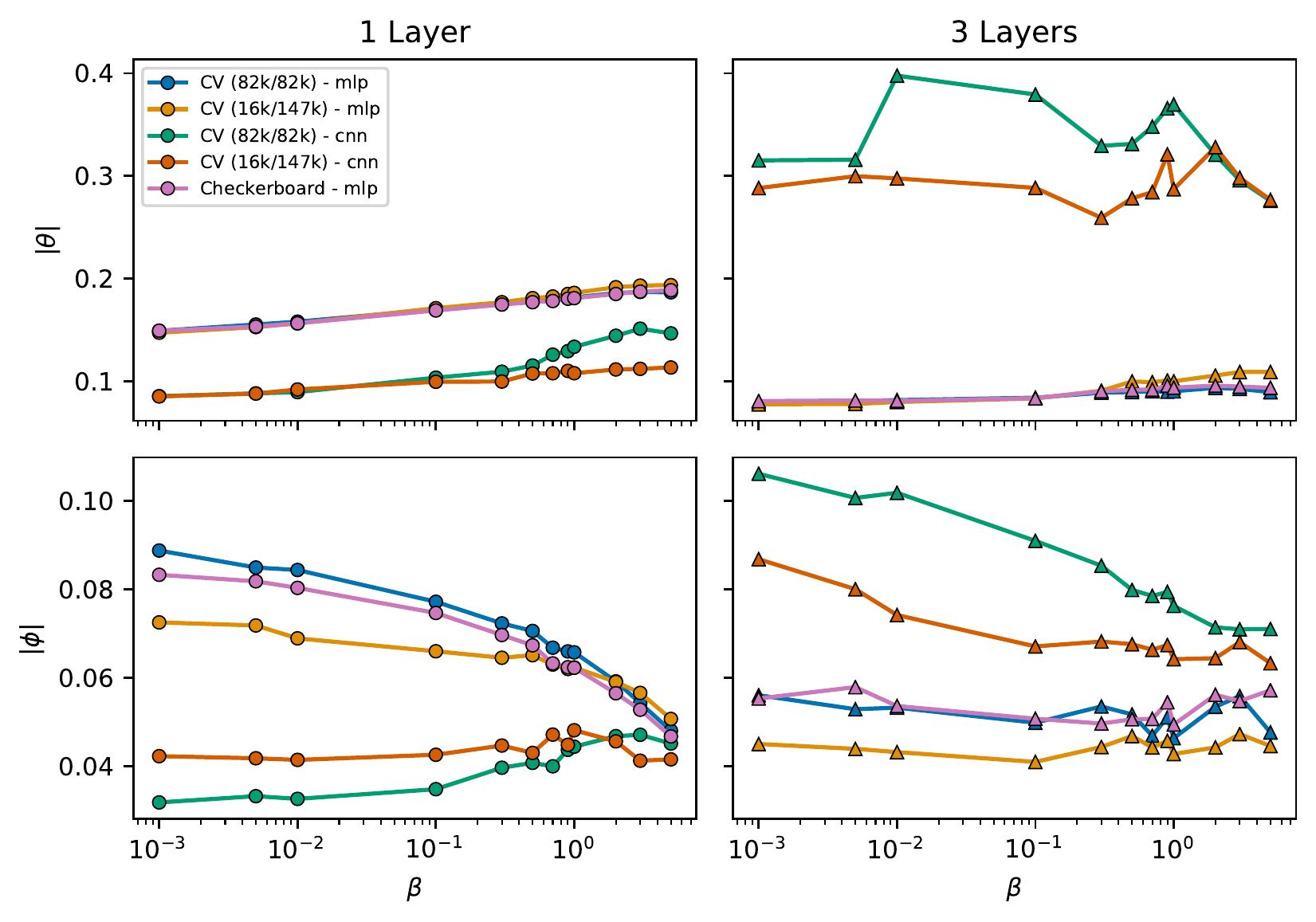}
\vspace*{-5ex}
\caption{Norm of the weights of decoder (\emph{top}), and encoder (\emph{bottom}) when trained with different values of $\beta$ on various splits trained with either MLPs or CNNs.}
\label{app:fig:theta-norms}
\end{figure}

Regularizers are typically terms that encourage the learning algorithm to choose simpler models. As a means of gaining intuition, we look at the average norms of the weights of both the encoder and the decoder (see Figure~\ref{app:fig:theta-norms}). We observe that as we increase $\beta$ (i.e.\nobreak\ rate penalty), the average norm of the decoder weights increase. In other words, having a low rate penalty results in learning not simple, but more complex decoders. This suggest that the rate not only does not act as a regularizer, but in fact it has the opposite effect.

\clearpage
\subsection{Role of Mutual Information vs. the Marginal KL}

\label{app:sec:mi_kl}
\begin{figure}[!h]
\centering
\includegraphics[width=\textwidth]{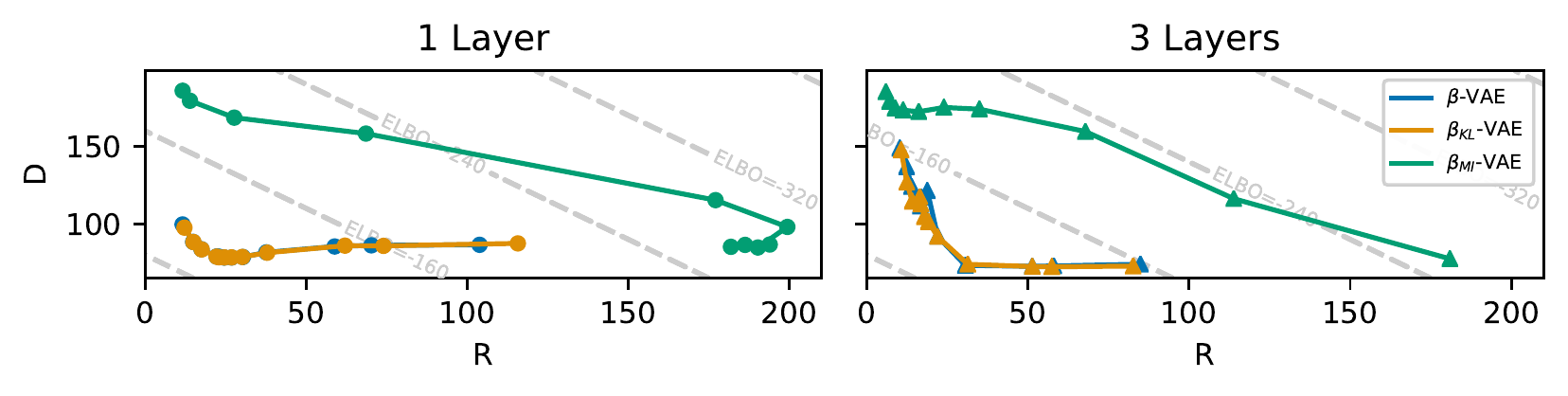}
\vspace*{-4ex}
\caption{Test $RD$ curves yielded by models trained with $\mathcal{L}_{\beta}$, $\mathcal{L}_{\text{KL}}$, $\mathcal{L}_{\text{MI}}$ objectives for 1-layer (\emph{Left}) and 3-layer (\emph{Right}) MLP architectures.}
\label{fig:gamma-vaes}
\end{figure}

In order to understand the individual impact of the MI vs. the marginal KL on the reconstruction, we report on the following analysis.
We can write a more general form of $\beta$-VAE objective where the MI and the marginal KL terms can have different coefficients: 
\begin{equation*}
    \mathcal{L}_{\beta_{\text{MI}}, \beta_{\text{KL}}}(\theta,\f) 
    = -D - \beta_{\text{MI}} \: I_{q}(\z;\x) - \beta_{\text{KL}}\:\KL{\q(\z)}{p(\z)},
\end{equation*}
and $\beta_{\text{MI}}=\beta_{\text{KL}}=\beta$ recovers the original $\beta$-VAE objective. We are interested in cases where $\beta_{MI}$=0, where effect of the MI is nullified or $\beta_{KL}$=0, where effect of the KL term is nullified. For emphasis, we refer to the objective with $\beta_{KL}$=0 as $\mathcal{L}_{\text{MI}}$, and the objective with $\beta_{MI}$=0 as $\mathcal{L}_{\text{KL}}$. 
\begin{align*}
    \mathcal{L}_{\text{MI}}(\theta, \phi) 
    &= -D - \beta_{\text{MI}} \: I_{q}(\x, \z)\\
    \mathcal{L}_{\text{KL}}(\theta, \phi) 
    &= -D - \beta_{\text{KL}} \: \KL{\q(\z)}{p(\z)}
\end{align*}
To ensure that it is the marginal KL that is causing the shape shift in the $RD$ curve, we trained VAEs with 1-Layer and 3-Layer architectures using objectives $\mathcal{L}_{\text{MI}}$\footnote{$\beta_{\text{MI}}$ $\in$ \{6., 8., 10., 11., 12., 13., 14., 15., 17., 20.\}} and $\mathcal{L}_{\text{KL}}$\footnote{$\beta_{\text{KL}}$ $\in$ \{0.001, 0.005, 0.01, 0.1, 0.3, 0.5, 0.7, 0.9, 1., 2., 3., 5.\}}, with 5 random restarts on the CV(16k/147k) split. 

Figure~\ref{fig:gamma-vaes} shows the test $RD$ curves for VAEs trained with $\mathcal{L}_{\beta}$, $\mathcal{L}_{\text{MI}}$, and $\mathcal{L}_{\text{KL}}$ objectives. The $RD$ curve for the $\beta$-VAE and the $\mathcal{L}_{\text{KL}}$ objective lie almost on top of each other, while the $RD$ curve for the $\mathcal{L}_{\text{MI}}$ is vastly different. These results confirm that it is indeed the marginal KL term that is responsible for impacting the generalization performance of VAEs.

\clearpage
\subsection{Early-Stopping}
\label{app:sec:early-stopping}

\begin{figure}[H]
\centering
\includegraphics[width=\linewidth]{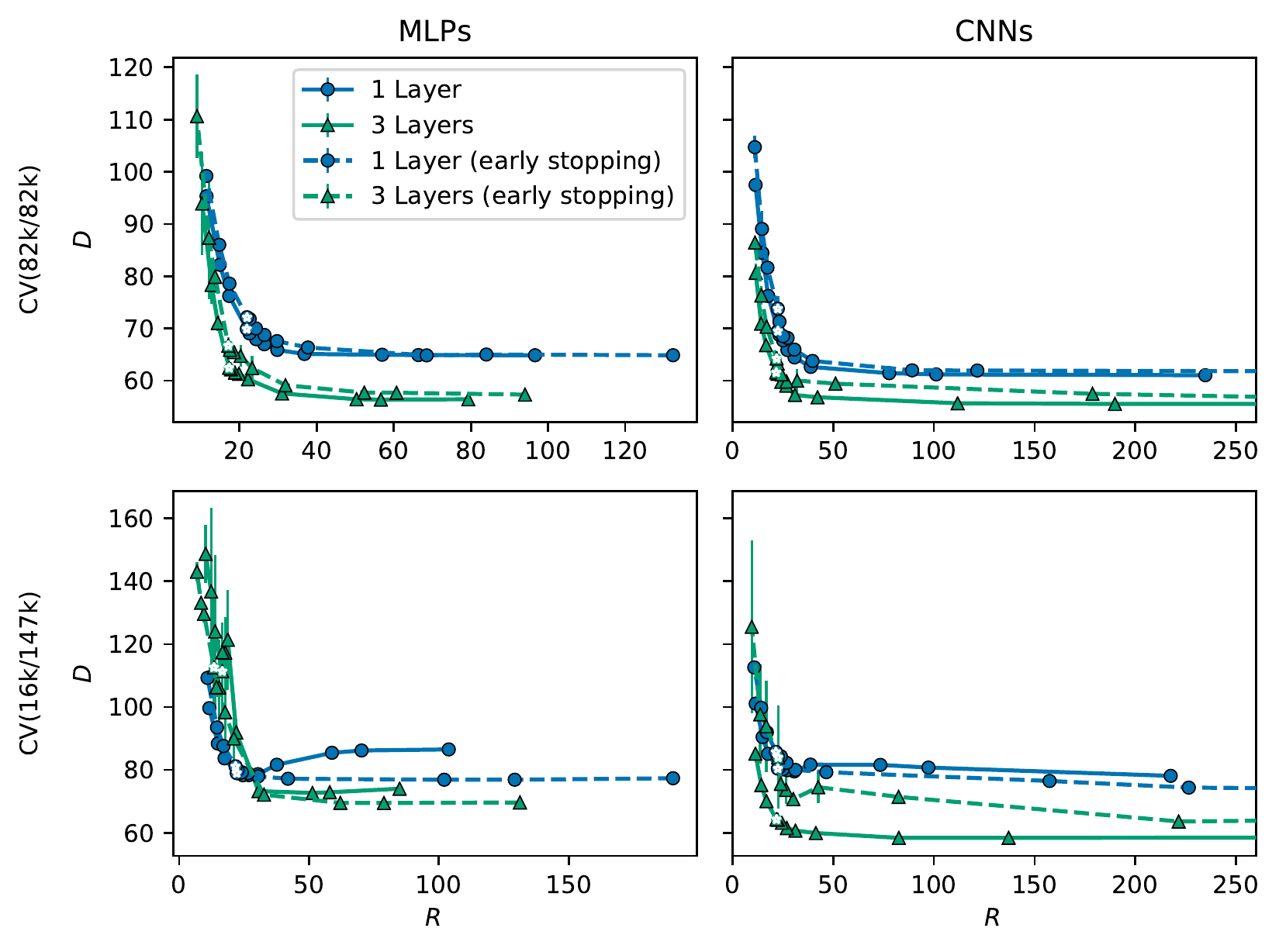}
\caption{RD curves evaluated on the test set for 1-layer and 3-layer VAEs with MLP and CNN architectures trained with and without early stopping.}
\label{app:fig:early-stoppings}
\end{figure}

In the previous experiments, we trained all models for a fixed number of 256k iterations (corresponding to 400 epochs for the CV (82k/82k) split). Here we consider applying early stopping in order to test whether the U-shaped $RD$ curve we observed in the 1-layer VAEs is due to overfitting. The results are shown in Figure~\ref{app:fig:early-stoppings}. Looking at the MLP architecture (left column), we see that early stooping does not make a strong impact in the default split. In the CV (16k/147k) split however, we see that early stopping improves generalization; shifting the curve to bottom left. Moreover, we see that U-shaped curve in the 1-layer case turns into an L-shaped when we apply early-stopping. This suggests that 1-layer VAEs in low $\beta$ regime were indeed overfitting as this can be mitigated by applying early stopping.


\clearpage
\subsection{Additional Datasets}
\label{app:sec:extra-datasets}

\begin{figure}[H]
    \centering
    \includegraphics[width=\textwidth]{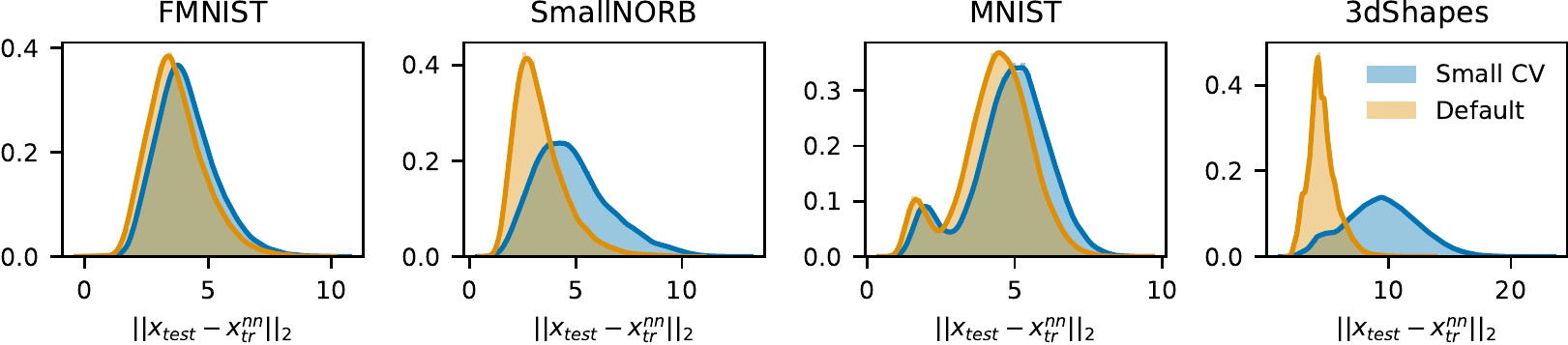}
    \caption[Histograms of $\ell_2$ distance between training and test examples in other datasets]{Normalized histograms of $\ell_{2}$ distance between test examples $\x$ and the nearest neighbour in the training set $\x^{tr}_{nn}$. for other datasets. For details of 'Default' and 'Small CV' see Table~\ref{app:tab:experiment-settings}.} 
    \label{app:fig:others-l2-hists}
\end{figure}

\begin{figure}[!h]
\centering
\includegraphics[width=0.9\linewidth]{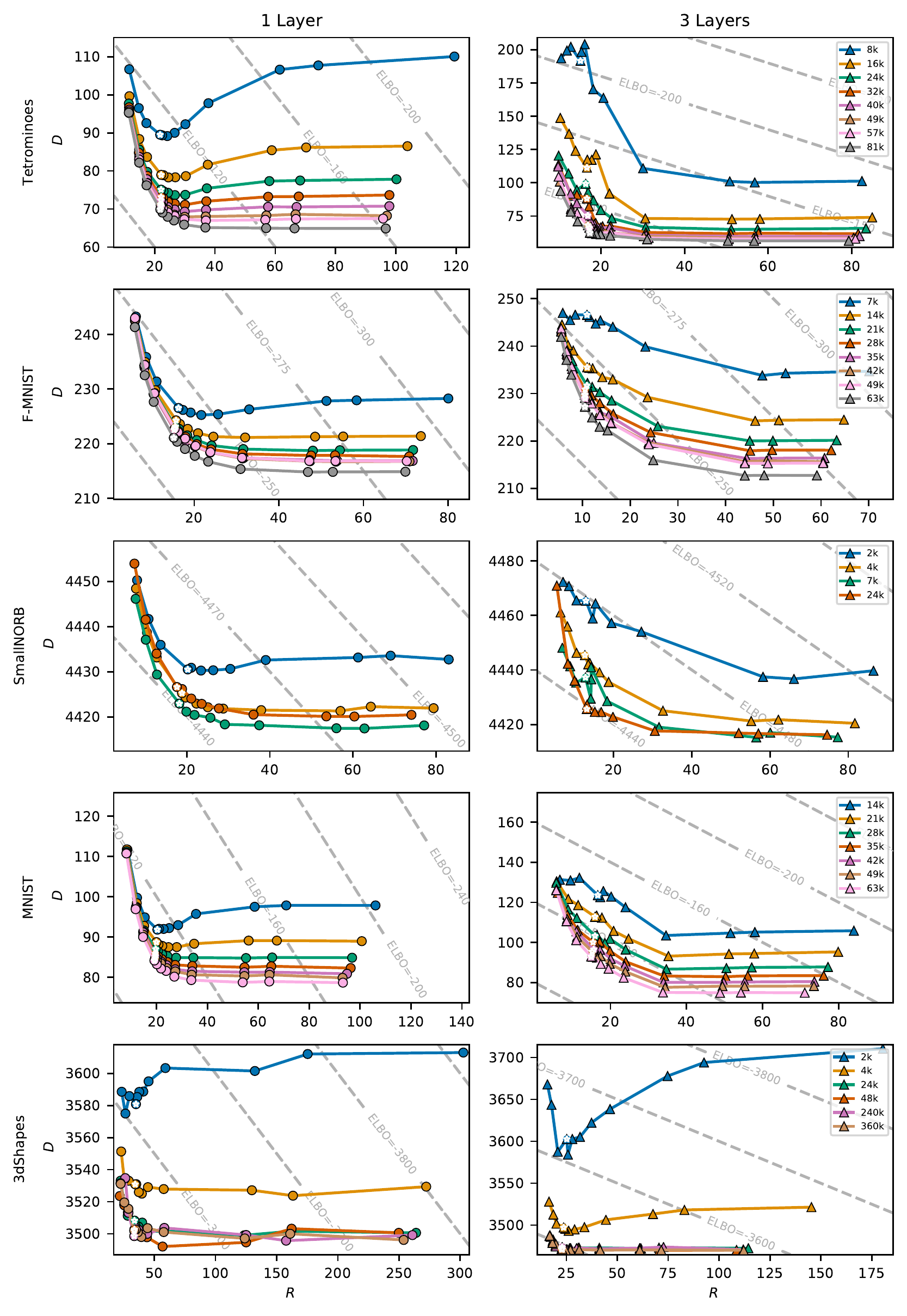}
\vspace*{-5ex}
\caption[$RD$ curves shown on various datasets with different CV splits trained with 1 and 3 layers.]{$RD$ curves shown on various datasets with different CV splits trained with 1 and 3 layers.}
\label{app:fig:other-datastes-all}
\end{figure}

Here we provide results for additional datasets. In Figure~\ref{app:fig:others-l2-hists}, we show the the histogram of $\ell_{2}$ distance between test examples and their nearest training neighbour for a large CV split (50\% train/test ratio) and a small CV split (10\% train/test ratio). In Figure~\ref{fig:other-datastes}, we only show the results for large one particular CV split. Here, we additionally show the results for other CV splits ranging from 5\% to 75\% of all the data (Figure \ref{app:fig:other-datastes-all}). For all datasets, we aimed to find train/test ratios where the 3-layer $RD$ curve would be above and below the 1-layer curve. 

In Figure \ref{app:fig:other-datastes-all}, we see that the $RD$ curve for 1-layer models resembles a U-shaped curve in most cases as the we decrease the number of training data while for 3-layer models this is not the case. We also observe the $RD$ curves for 3dShapes is very different compared to others. In particular, some of the $RD$ curves for 3-layer networks resemble a U-shaped curve. We suspect that this is because the 3dShapes datasets is a more difficult datasets compared to others both due to its size and certain factors of variations such as background color. 

\subsection{Robustness}

\begin{figure}[!h]
\centering
\includegraphics[width=\linewidth]{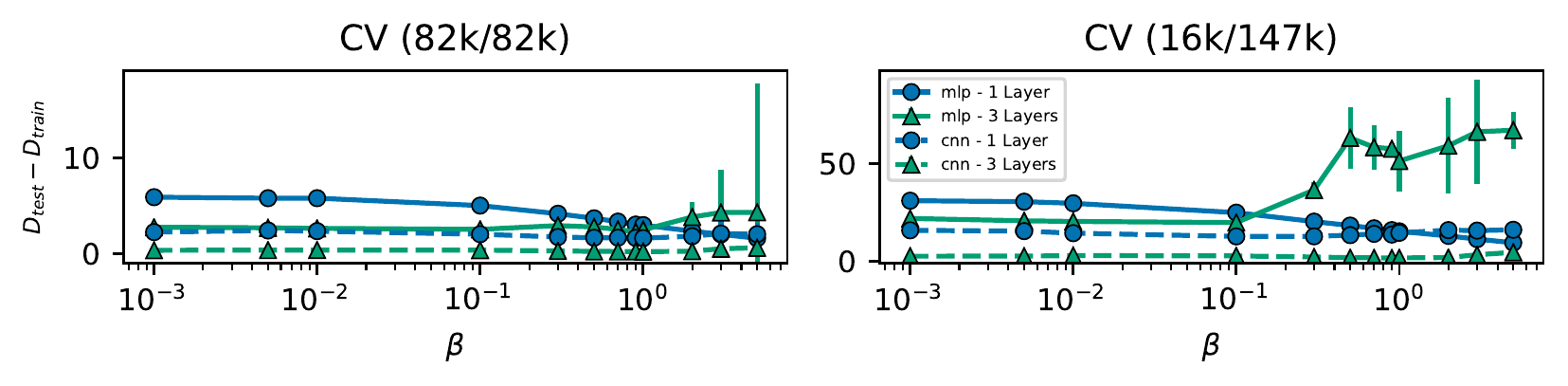}
\vspace*{-5ex}
\caption[Distortion generalization gaps shown for MLPs and CNNs]{Difference between test distortion and train distortion for VAEs trained with MLPs and CNNs on Tetrominoes dataset.}
\label{app:fig:rd-d-diffs-cnns}
\end{figure}

We also evaluate generalization based on the variance (in the context of bias-variance trade-off) by computing the difference between test set distortion and training set distortion, for CV(82k/82k) and CV(26k/147k) datasets (Figure~\ref{app:fig:rd-d-diffs-cnns}). A small difference is an indication of robustness (low variance) while a large difference is an indication of overfitting. We observe two opposite patterns, that are consistent across datasets. Increasing $\beta$ decreases the difference for 1-layer VAEs, while increasing the difference for 3-layer VAEs. In fact, when $\beta$ is low, the difference in distortion between the training and test sets is lower than the 1-layer model.

\clearpage
\subsection{Additional Figures}

\begin{figure}[!ht]
\centering
\includegraphics[
height=0.45\textheight,keepaspectratio]{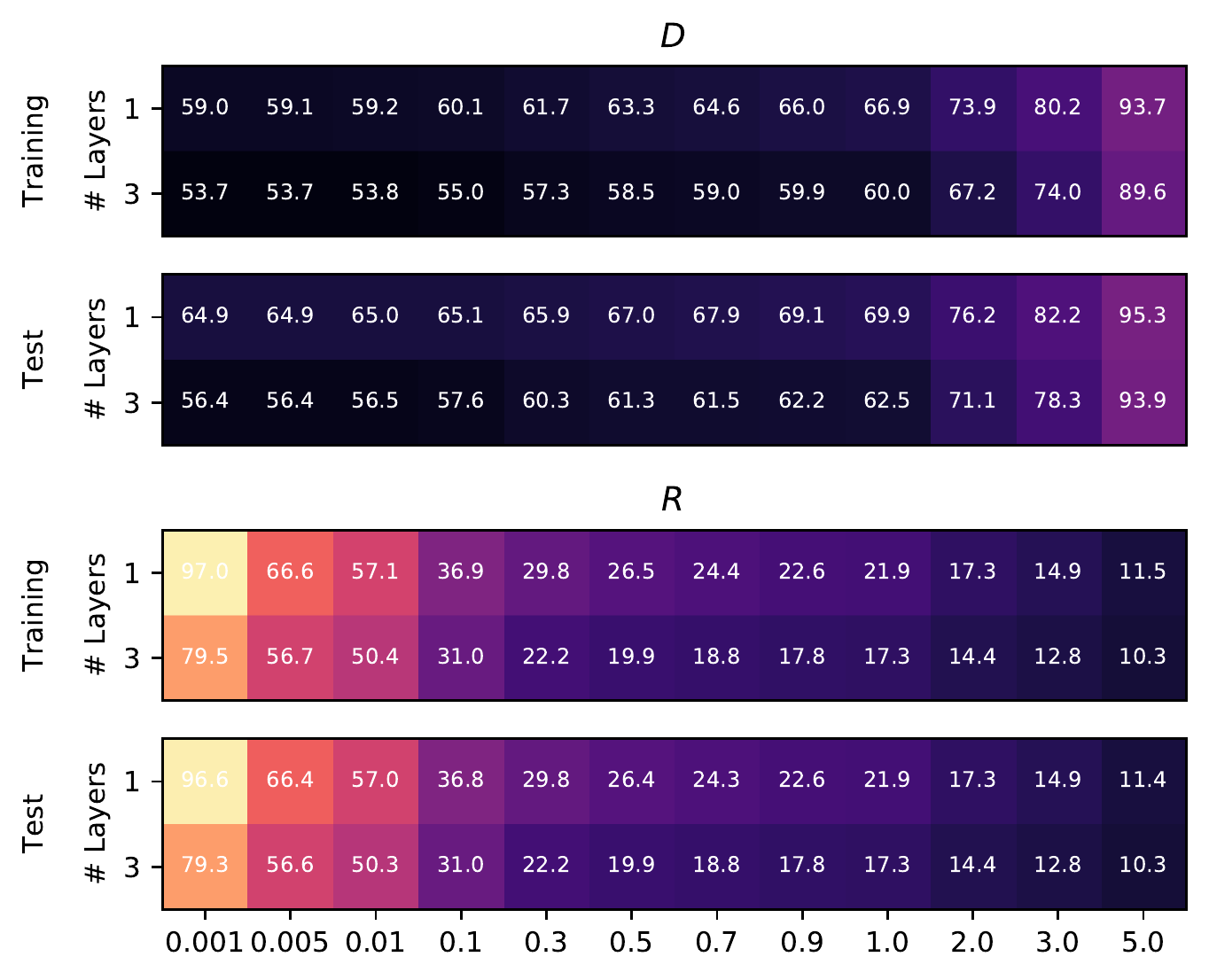}
\includegraphics[
height=0.45\textheight,keepaspectratio]{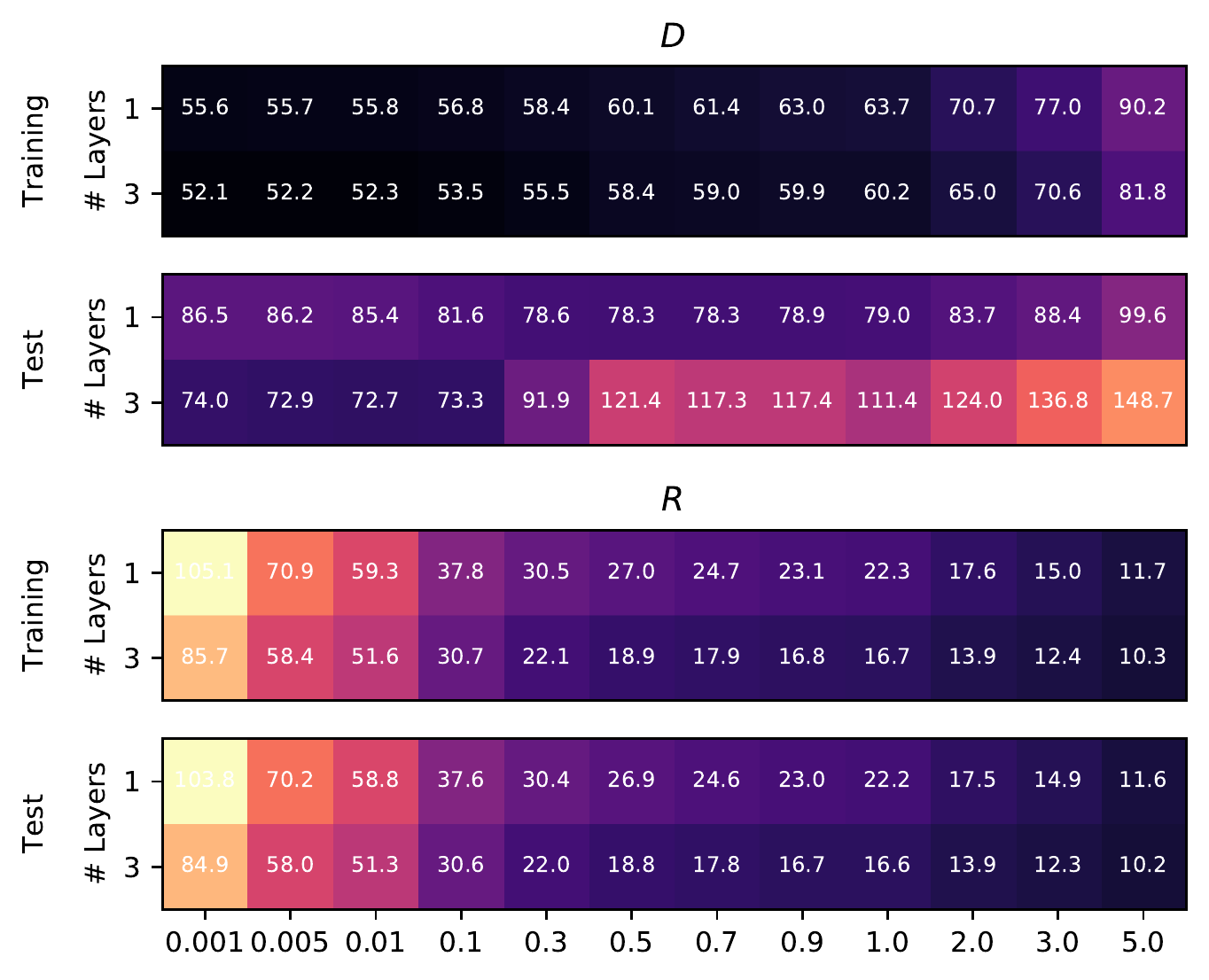}
\vspace*{-2ex}
\caption[Heatmaps of $R$ and $D$]{$RD$ values on evaluated on the test set for different values of $\beta$. The top and bottom two maps are for CV (82k/82k) the CV (16k/147) splits respectively.}
\label{app:fig:rd-heatmaps}
\end{figure}

\begin{figure}[!h]
    \centering
    \includegraphics[width=\linewidth]{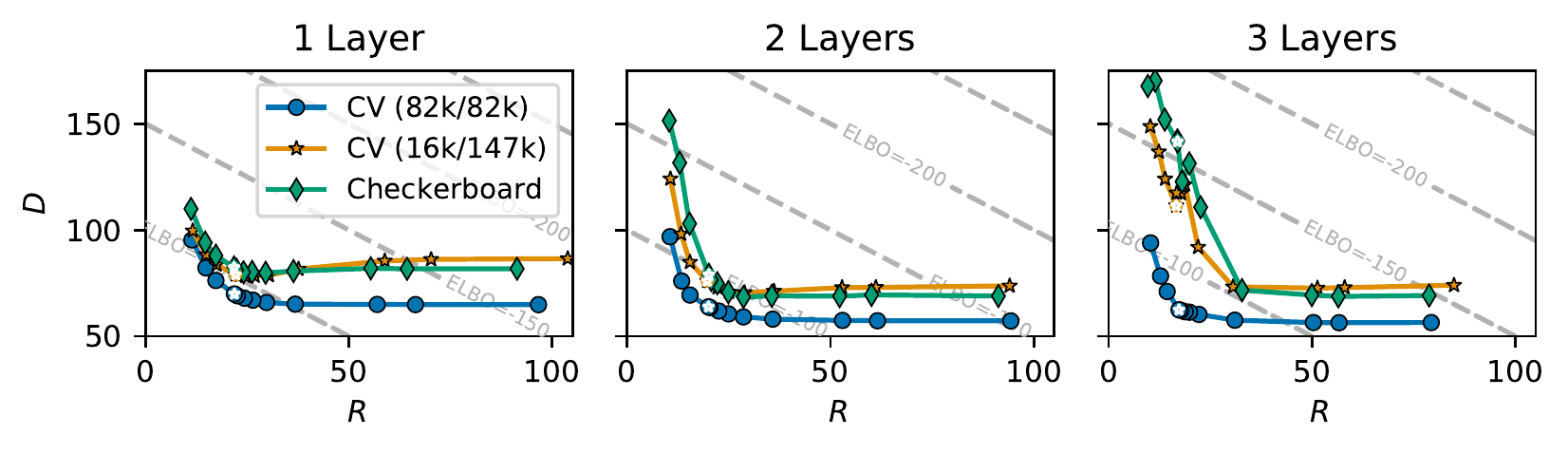}
    \includegraphics[width=\linewidth]{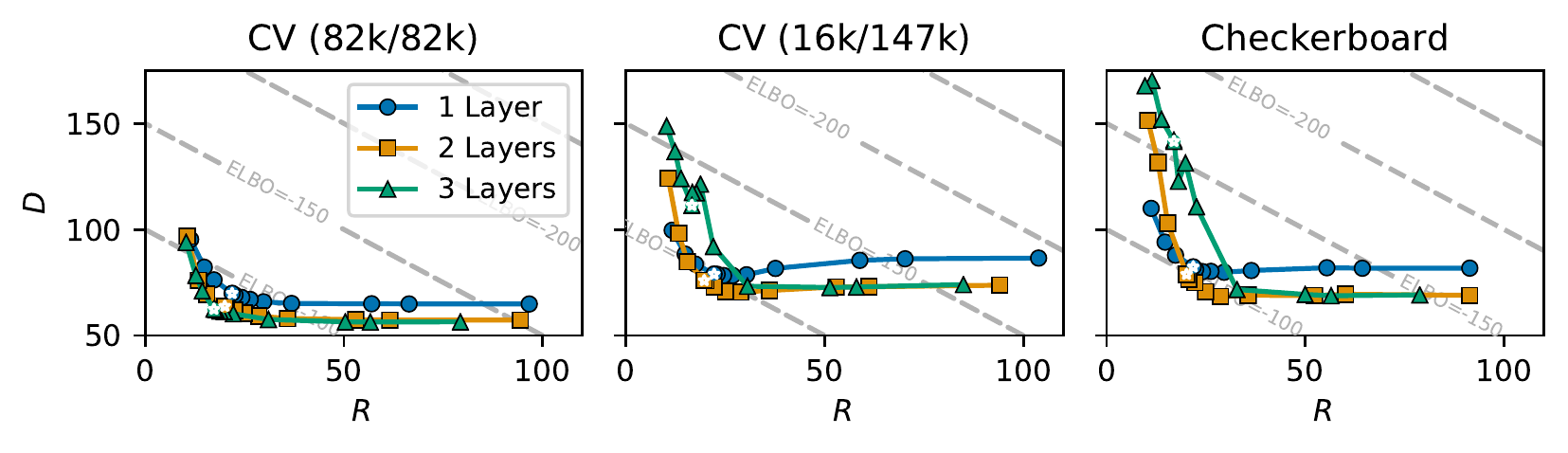}
    \vspace*{-5ex}
    \caption[$RD$ curves shown in alternative views]{$RD$ curves shown in two alternative views. The first view compares the effect of network depth for datasets with different levels of difficulty (\emph{Top}). The second view is the effect of making the generalization problem more difficult in models with different capacity (\emph{bottom}).}
    \label{app:fig:rd-alternative-views}
\end{figure}

\begin{figure}[!h]
\centering
\subfigure[Training set]{\includegraphics[width=\linewidth]{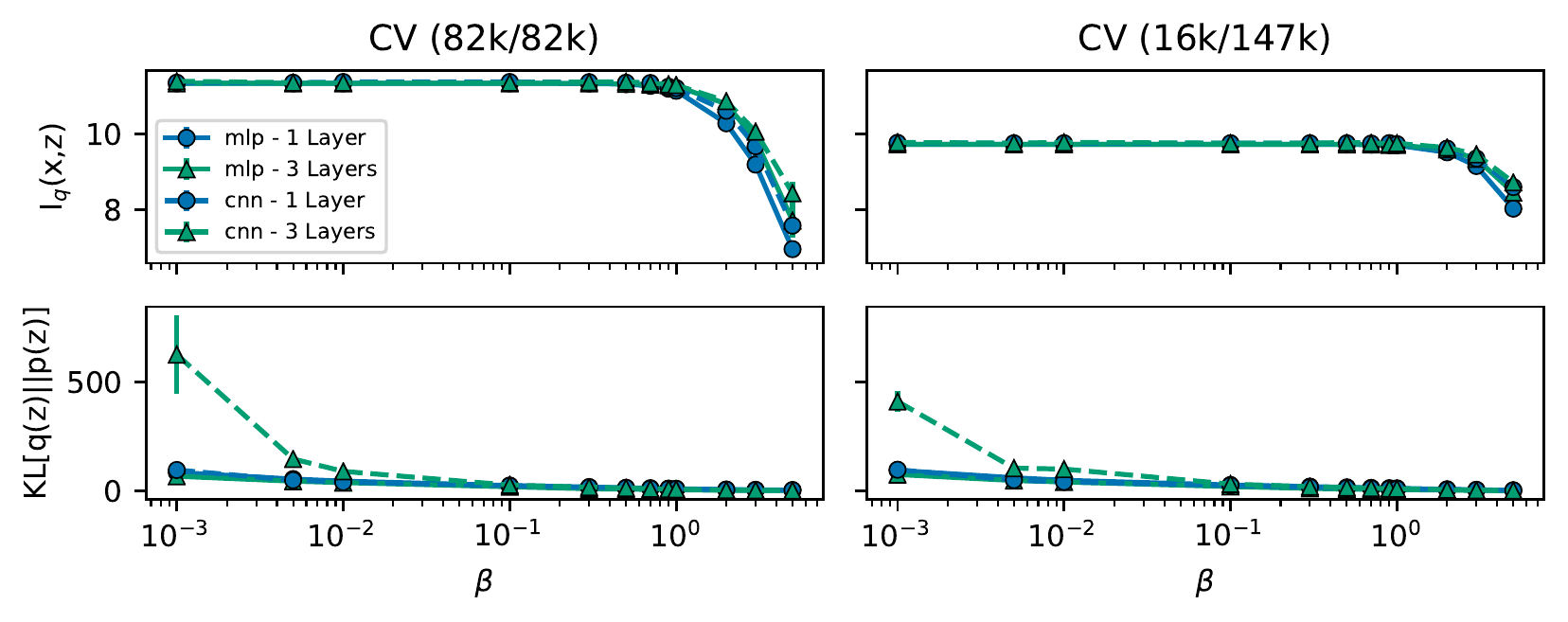}\vspace*{-5ex}}\vspace*{-2.5ex}\quad
\subfigure[Test set]{\includegraphics[width=\linewidth]{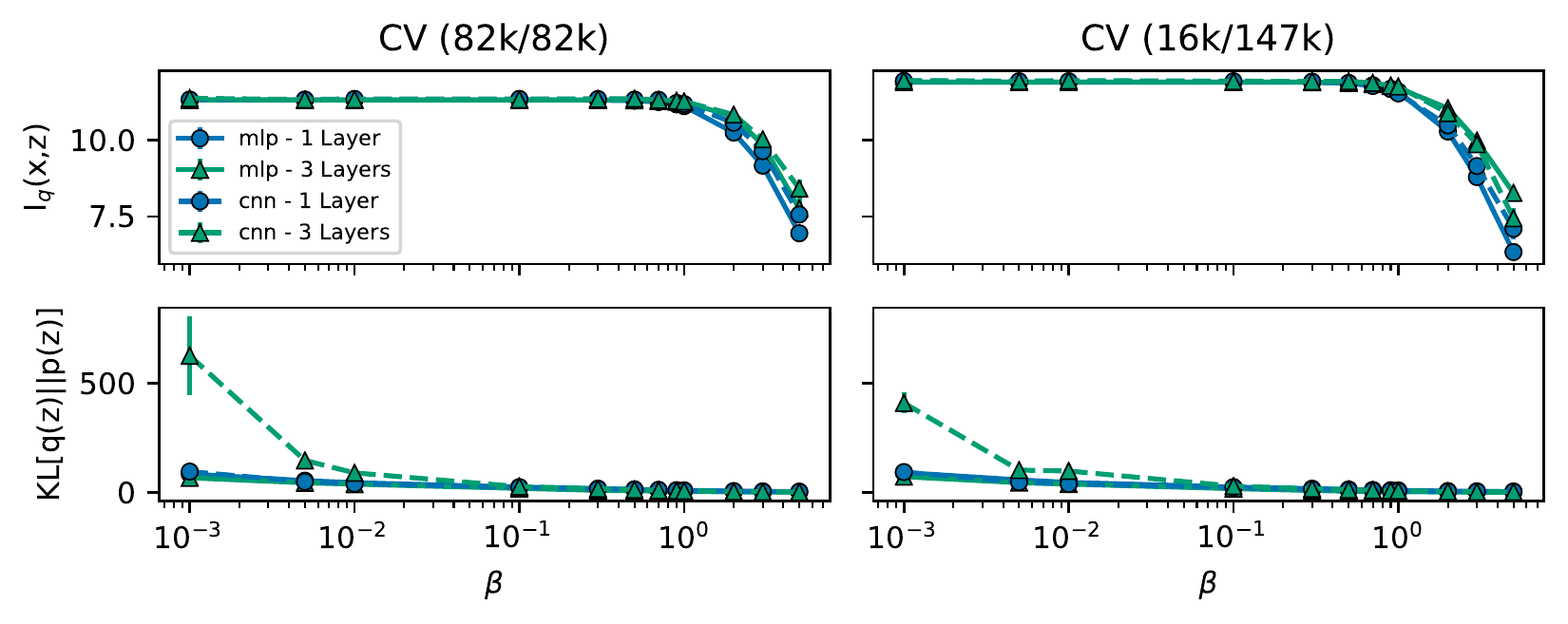}\vspace*{-5ex}}
\caption[MI and marginal KL for MLPs and CNNs]{$I_{q}(\x, \z)$ and $\KL{\q(\z)}{p(\z)}$ for $\beta$-VAE with MLP and CNN architectures, trained on CV(82k/82k) and CV(16k/147k) with different $\beta$-values.}
\label{app:fig:beta-Ixz-kl}
\end{figure}

\begin{figure}[!h]
    \centering
    \includegraphics[width=\linewidth]{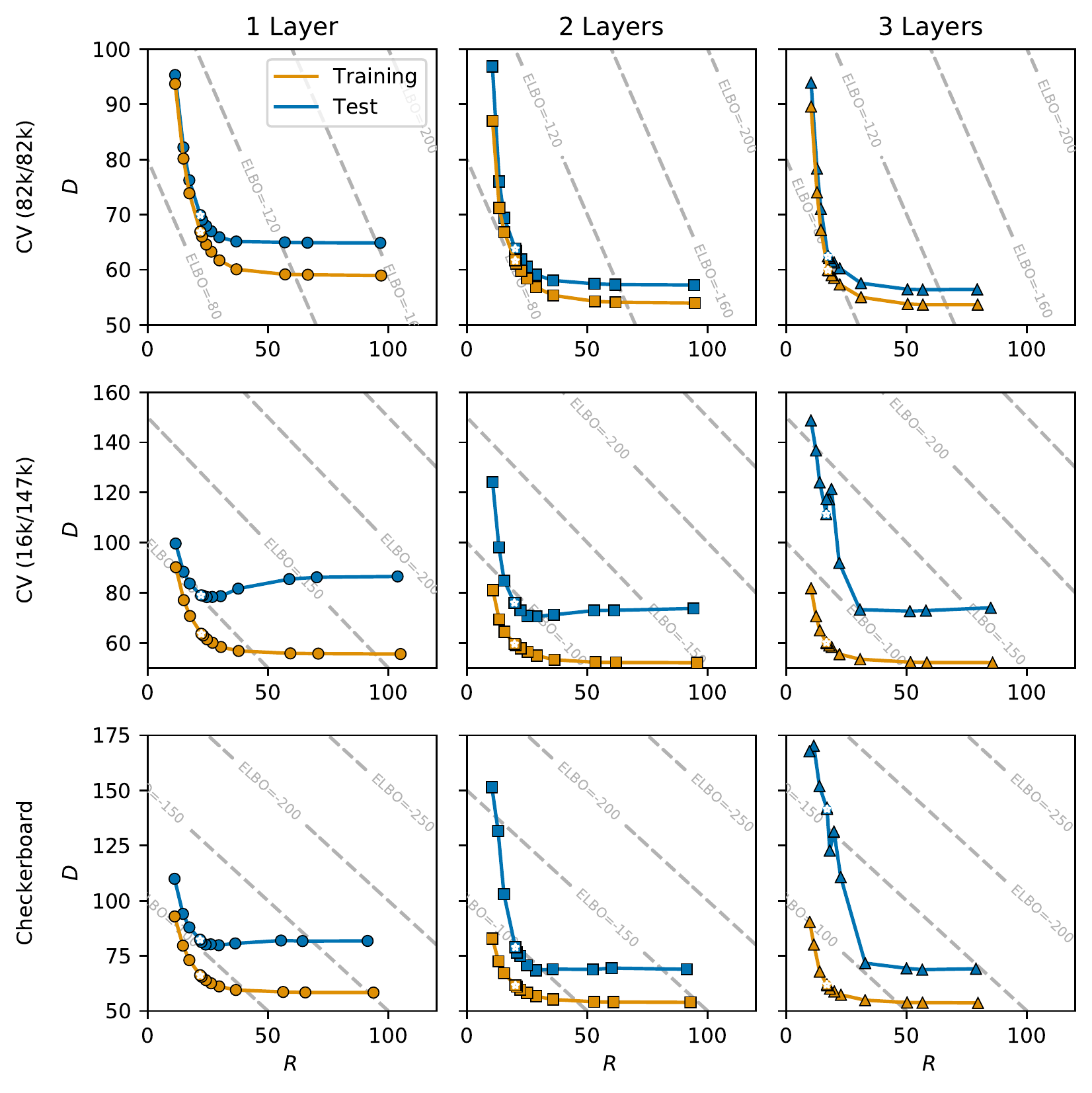}
    \vspace*{-5ex}
    \caption[$RD$ curves for training and test set]{$RD$ curves on training and test set with the Default, CV (16k/147k), and Checkerboard splits in models with 1, 2, and 3 layers.}
    \label{app:fig:rd-train-test}
\end{figure}

\begin{figure}[!h]
\centering
\includegraphics[width=\linewidth]{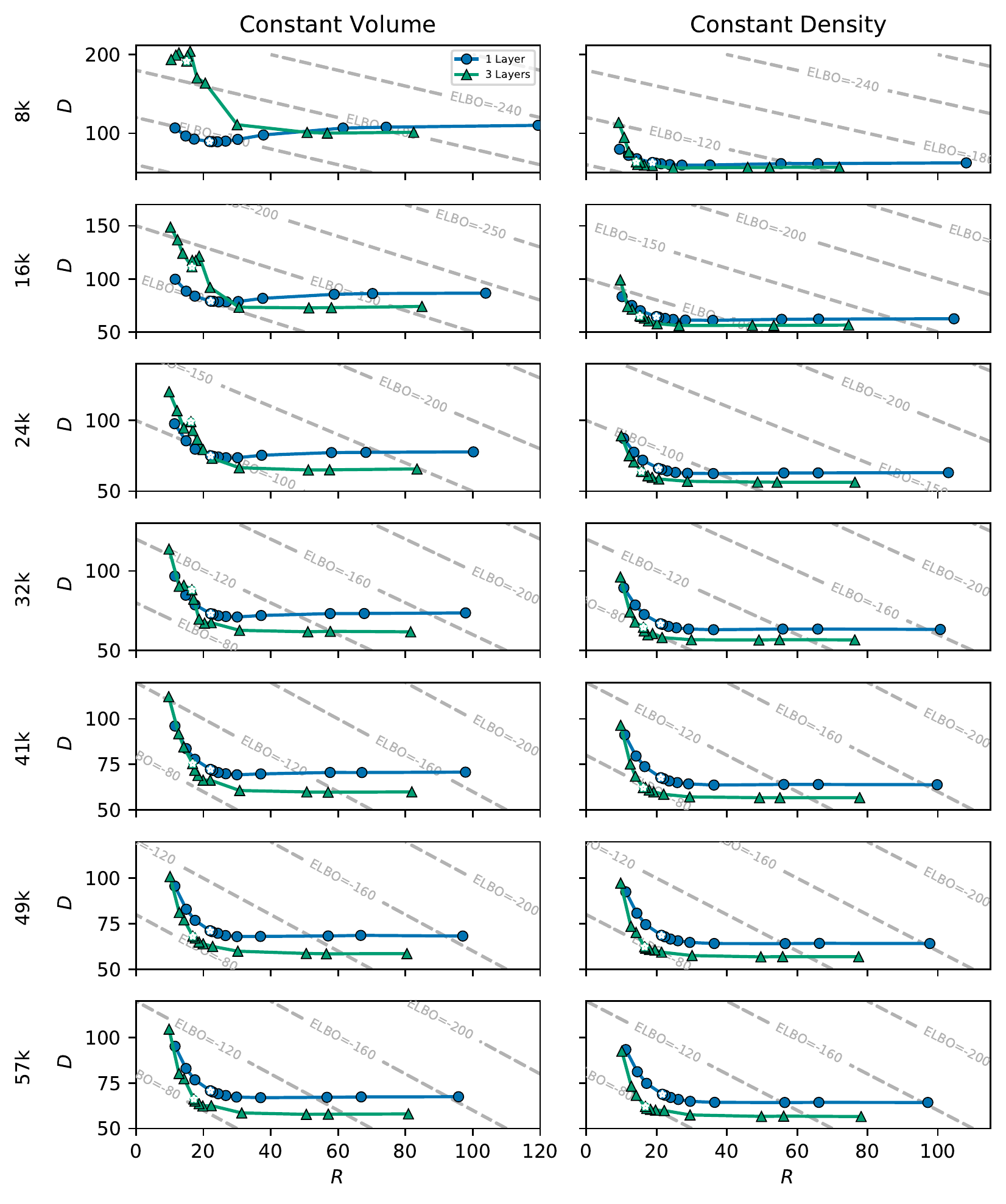}
\vspace*{-5ex}
\caption[$RD$ curves shown for different \ntrain{}]{Estimated $RD$ curves for different number of training data for when the volume is kept constant constant and reduce to keep the density the same. The uppermost panel is ($\ntrain{} = 8k$) and the lowest is for ($\ntrain{} = 57k$)}
\label{app:fig:rd-datasize-full}
\end{figure}

\begin{figure}[!h]
\centering
\includegraphics[width=\linewidth]{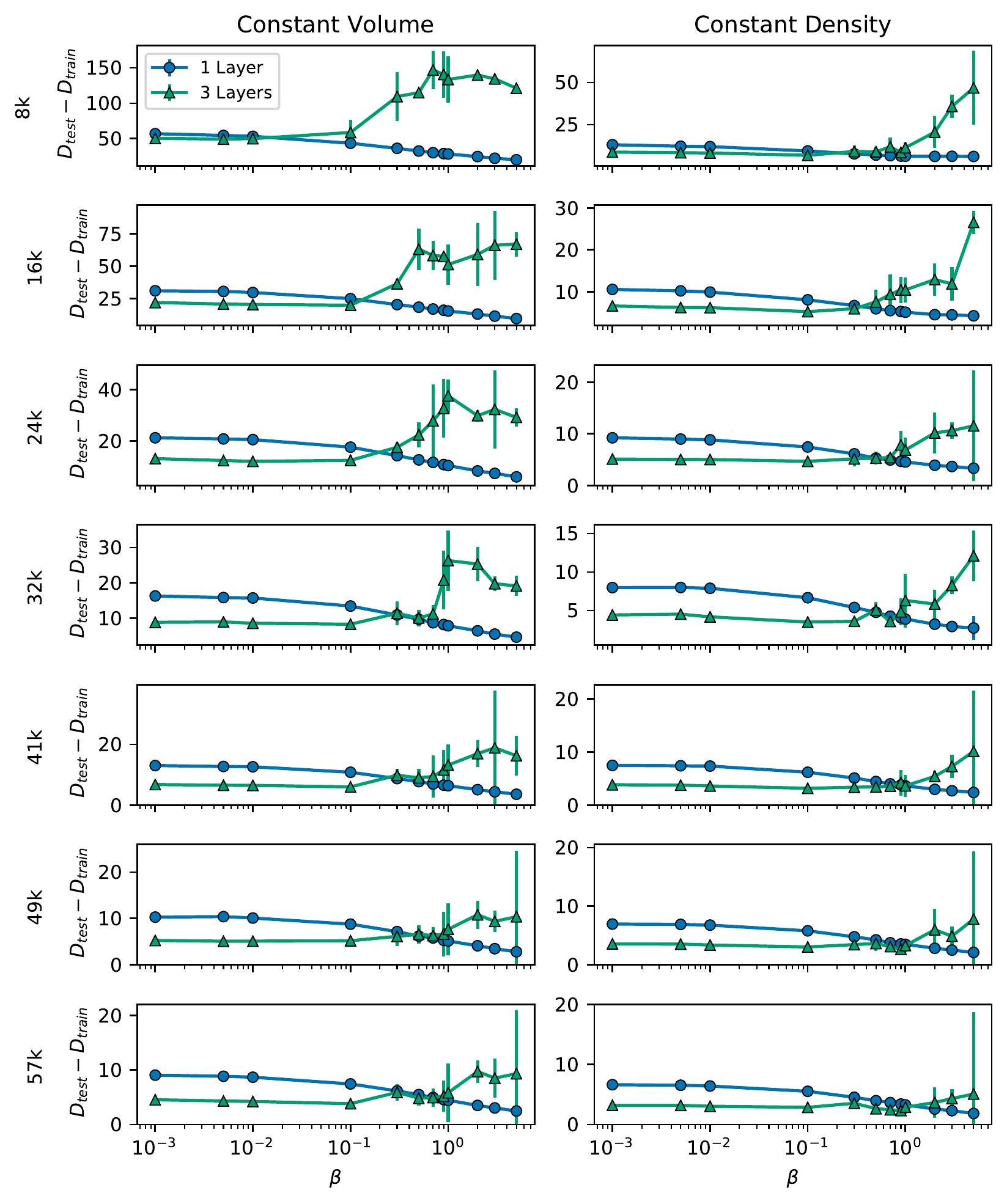}
\vspace*{-5ex}
\caption[Distortion generalization gaps shown for different \ntrain{}]{Difference between test distortion and train distortion for CV and CD splits of Tetrominoes dataset with different \ntrain{}.}
\label{app:fig:rd-d-diffs-all}
\end{figure}

\begin{figure}[!h]
\centering
\includegraphics[width=\linewidth]{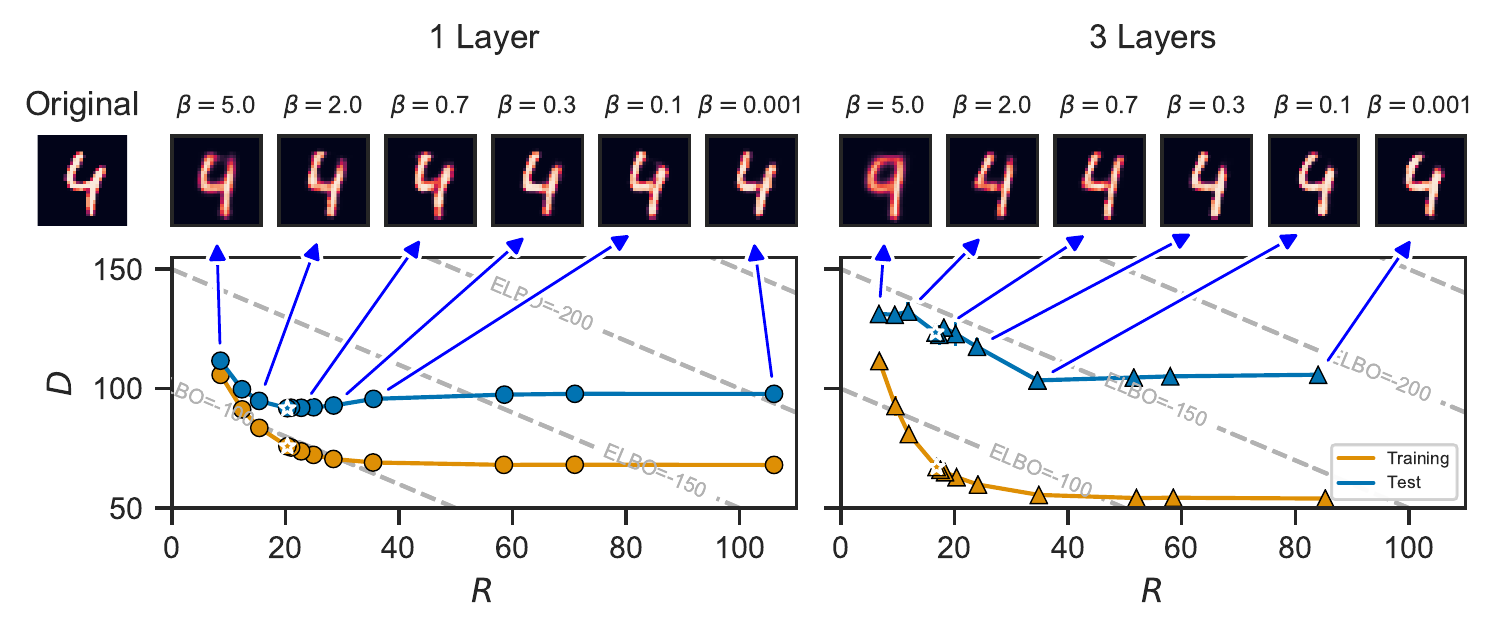}
\includegraphics[width=\linewidth]{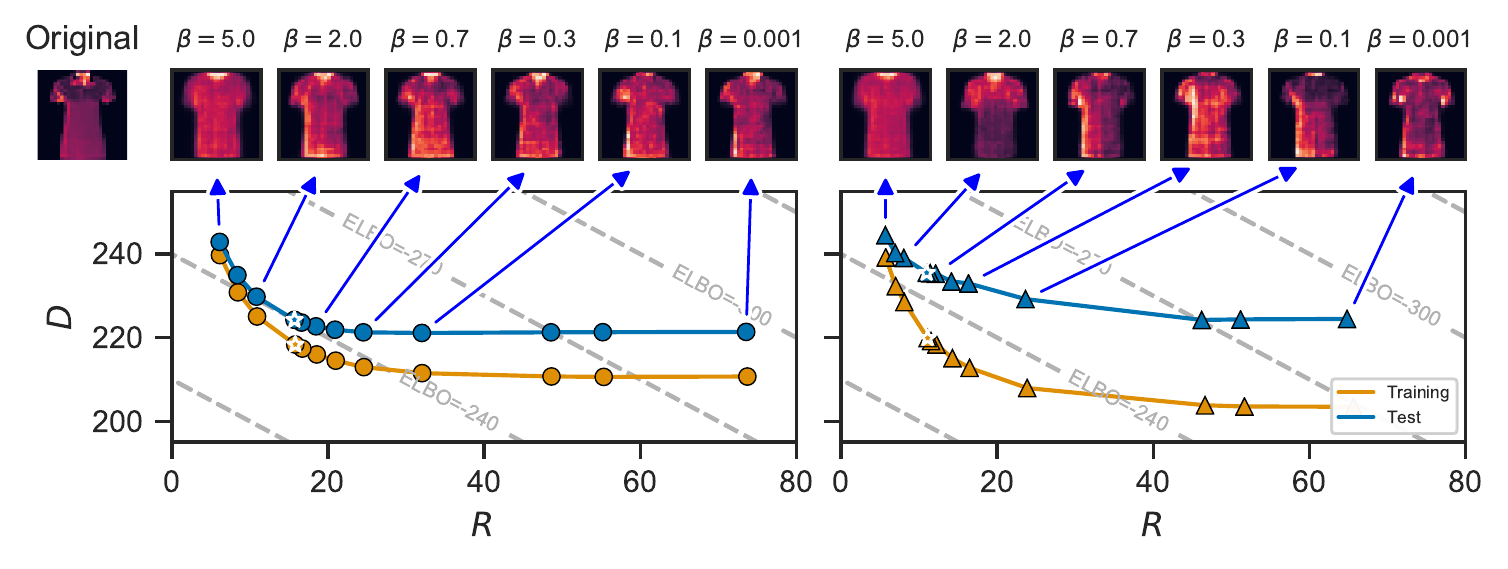}
\includegraphics[width=\linewidth]{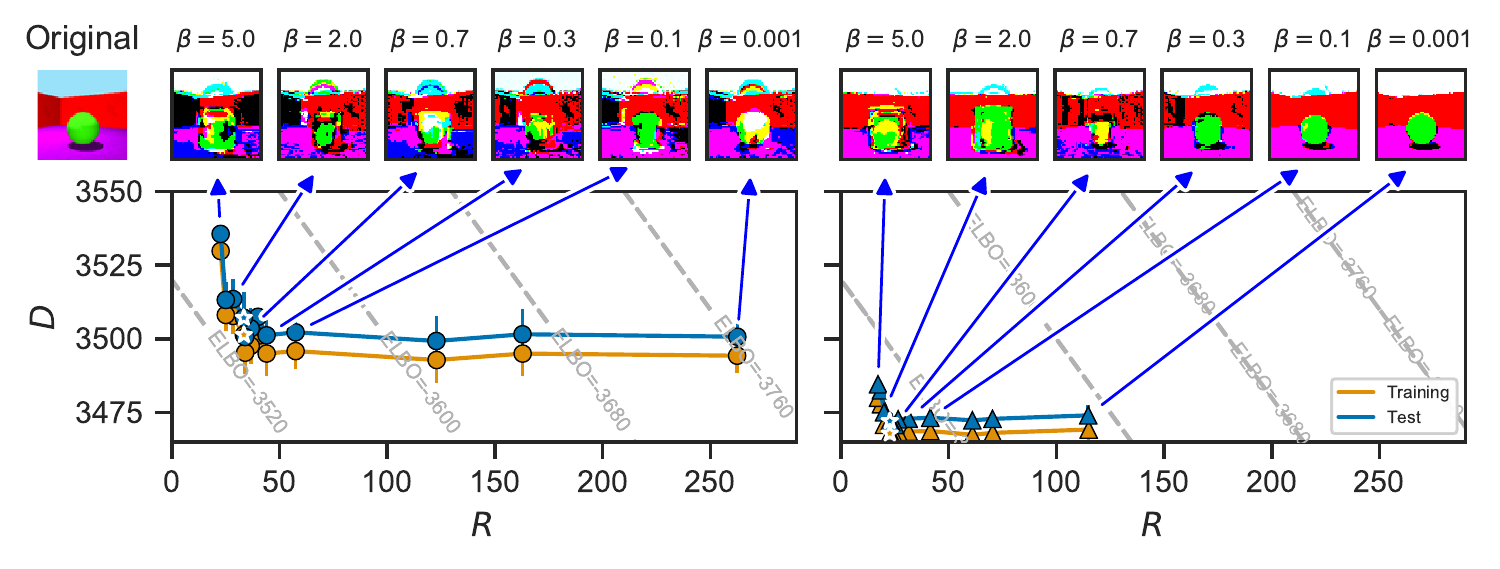}
\caption{Training and test $RD$ curves for a 1-layer (left) and a 3-layer (right) architecture for different datasets. Each dot constitutes a $\beta$ value (white stars indicate the $\beta$=1), averaged over 5 restarts. Images show reconstructions of a test example. The splits are CV(14k/56k), CV(7k/63k), and CV(48k/432k) for MNIST, Fashion-MNIST, and 3dShapes respectively.}
\label{app:fig:rd-recons-curves-other-datasets}
\end{figure}

\begin{figure}[!ht]
\centering
\includegraphics[width=\linewidth]{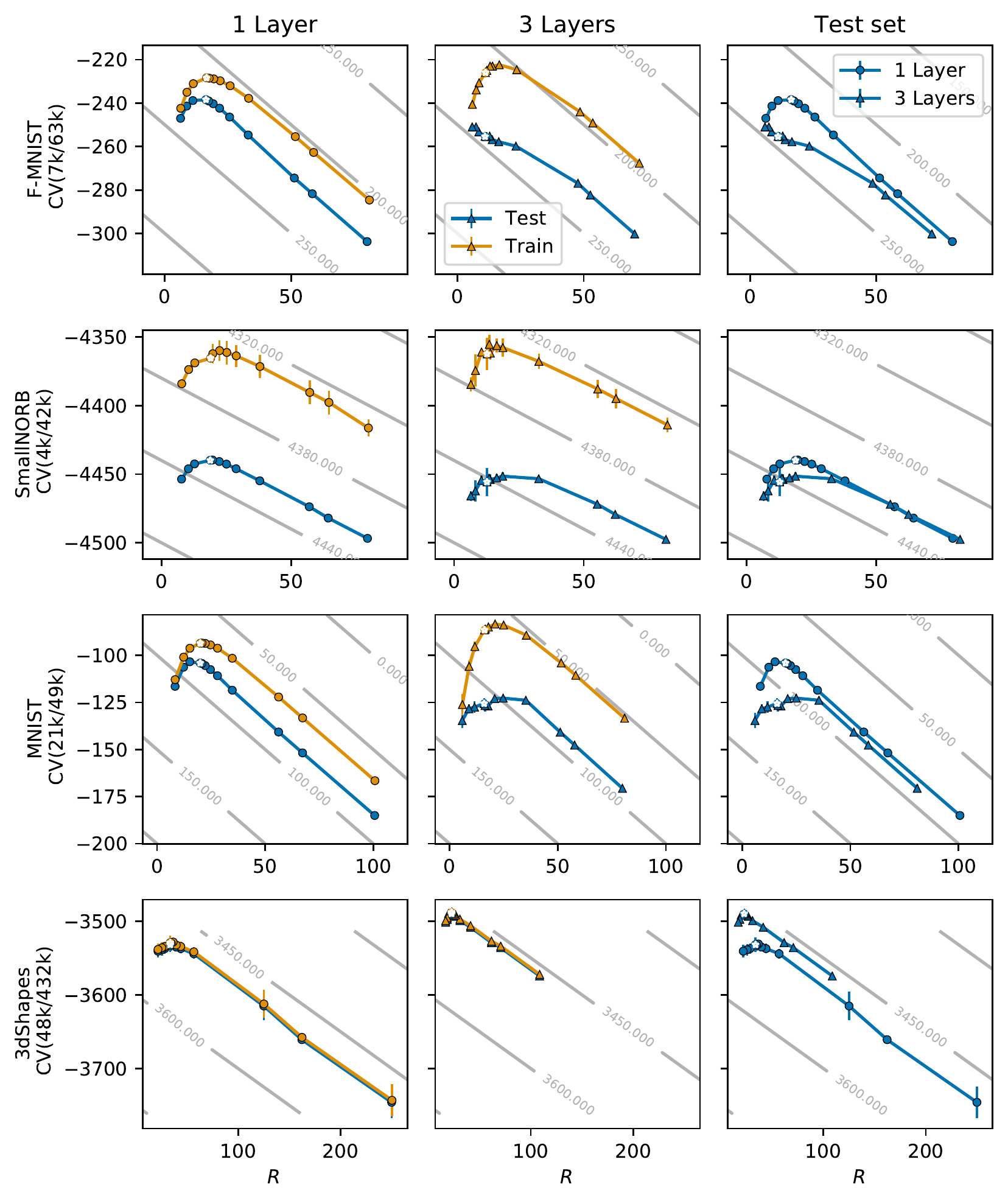}
\vspace{-4ex}
\caption[LM vs rate evaluated for various datasets]{Log marginal likelihood and rate evaluated on training and test sets for different datasets, trained with 1-layer and 3-layer models. Each dot constitutes a $\beta$ value (white stars indicate the $\beta$=1 point), averaged over 5 independent restarts.}
\vspace{-3ex}
\label{app:fig:logp-vs-kl-other-datasets}
\vspace*{3ex}
\end{figure}

\begin{figure}[!ht]
\centering
\includegraphics[width=\linewidth]{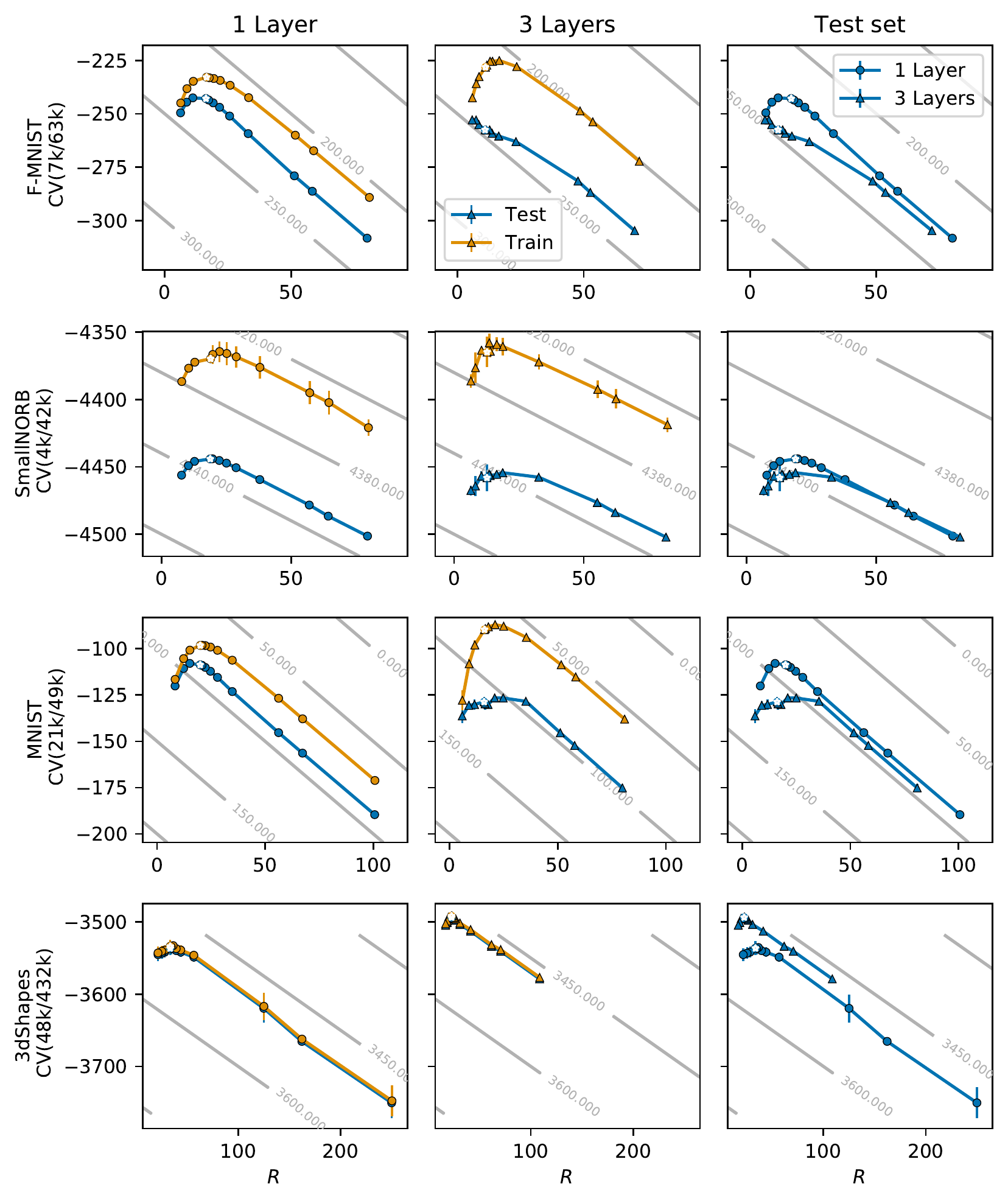}
\vspace{-4ex}
\caption[ELBO vs rate evaluated for various datasets]{ELBO and rate evaluated on training and test sets for different datasets, trained with 1-layer and 3-layer models. Each dot constitutes a $\beta$ value (white stars indicate the $\beta$=1 point), averaged over 5 independent restarts.}
\vspace{-3ex}
\label{app:fig:elbo-vs-kl-other-datasets}
\vspace*{3ex}
\end{figure}

\begin{figure}[!h]
\centering
\includegraphics[width=\linewidth]{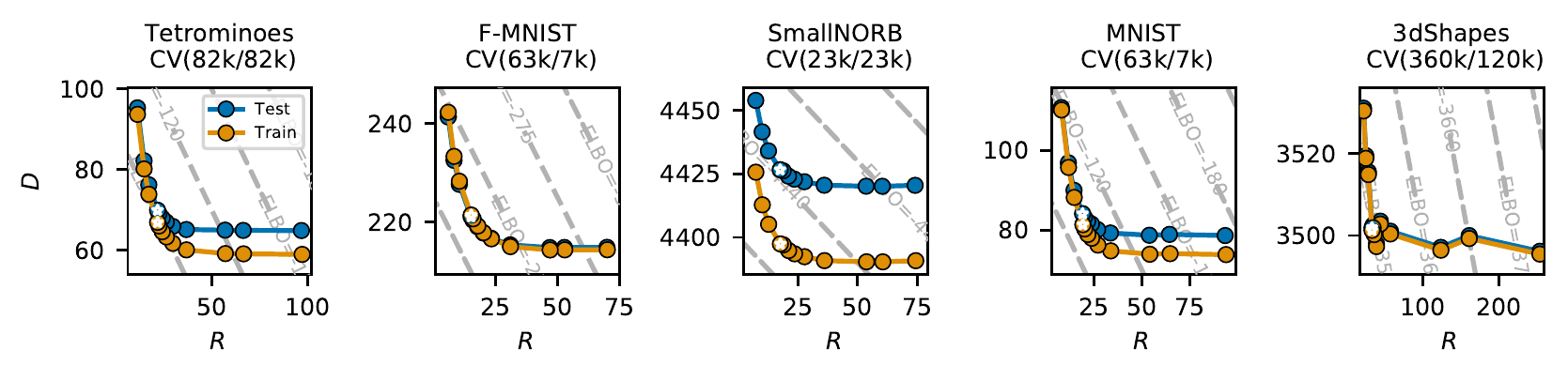}
\includegraphics[width=\linewidth]{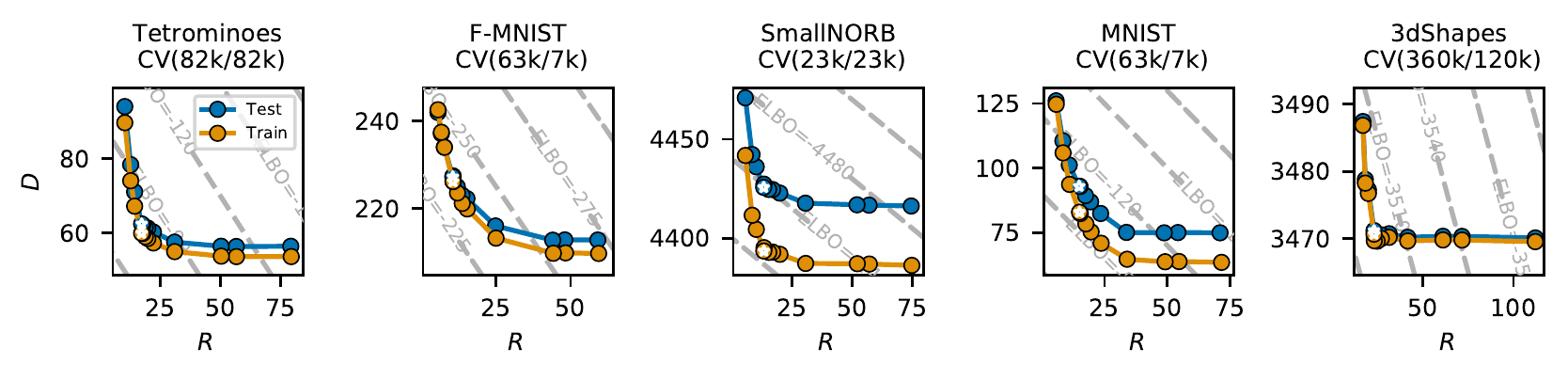}
\includegraphics[width=\linewidth]{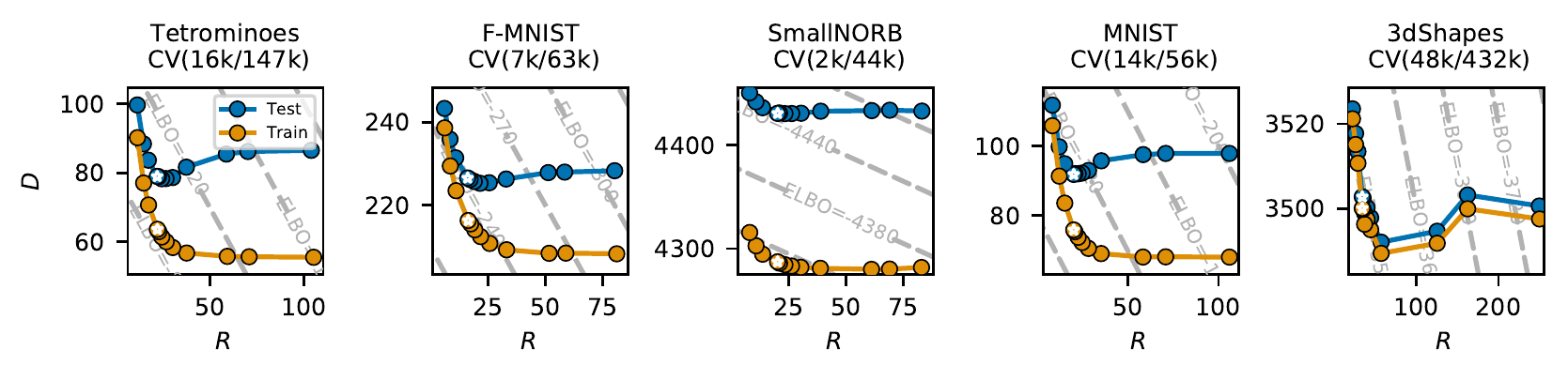}
\includegraphics[width=\linewidth]{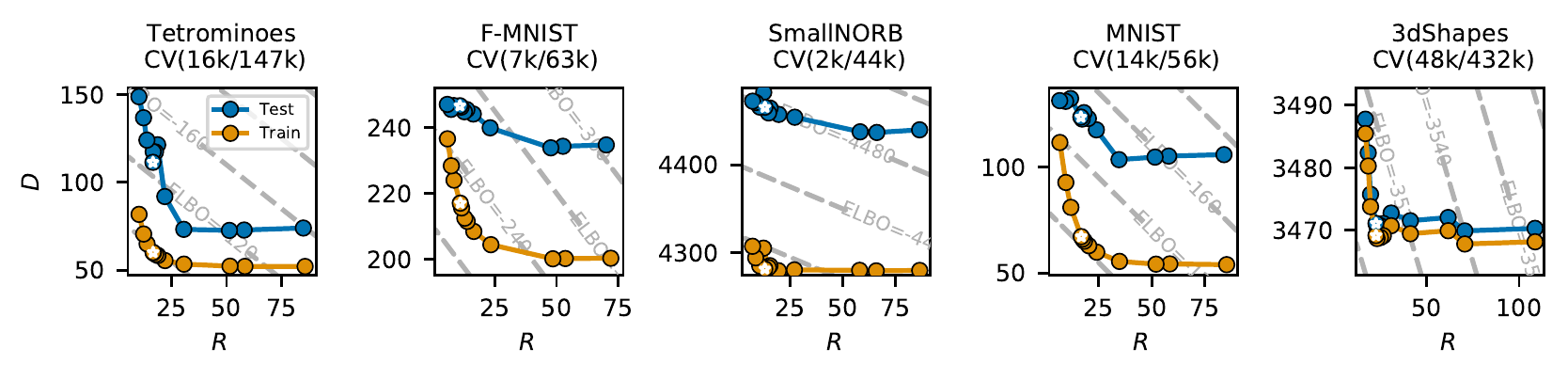}
\caption{$RD$ curves evaluated on training and test set for different datasets for large CV split trained with 1-layer (1\textsuperscript{st} from top), large CV split trained with 3-layer (2\textsuperscript{nd} from top), small CV split trained with 1-layer (2\textsuperscript{nd} from bottom), small CV split trained with 3-layer (1\textsuperscript{st} from bottom).}
\label{app:fig:other-datasets-rds-train}
\end{figure}
\end{document}